\documentclass[authoryear,sort&compress,3p]{elsarticle} %

\usepackage{multirow, rotating, array}
\usepackage{amsfonts, amssymb, amsmath}
\usepackage{enumitem}
\usepackage{pdflscape}
\usepackage{onimage}
\usepackage{tikz}
\usepackage{floatrow}
\usepackage{flafter}
\usepackage{setspace}
\usepackage[percent]{overpic}

\usepackage[hidelinks]{hyperref}
\hypersetup{
  colorlinks   = true, %
  urlcolor     = blue, %
  linkcolor    = red, %
  citecolor    = red   %
}
\usepackage{color, graphicx, float}
\usepackage{subcaption}
\usepackage{afterpage}
\usepackage{flafter}
\usepackage[section]{placeins}
\usepackage{algorithmic}
\usepackage[algoruled]{algorithm2e}

\usepackage{wrapfig}
\usepackage{makeidx}
\usepackage{makecell}
\usepackage{booktabs}

\usepackage[T1]{fontenc}
\usepackage[utf8]{inputenc}

\graphicspath{{images/}}
\DeclareGraphicsExtensions{.png,.jpg}

\newcommand{\norm}[1]{\left\lVert#1\right\rVert}
\newcommand{\abs}[1]{\lvert#1\rvert}
\makeatletter
\g@addto@macro\@floatboxreset\centering
\makeatother

\makeatletter
\AtBeginDocument{%
  \expandafter\renewcommand\expandafter\subsection\expandafter{%
    \expandafter\@fb@secFB\subsection
  }%
}
\makeatother

\def\yes{$\color{green!50!black} \checkmark$}
\def\no{\color{red} X}

\newcommand{\review}[1]{\textcolor{magenta}{#1}}
\renewcommand{\review}[1]{#1}
\usepackage[normalem]{ulem}
\renewcommand{\sout}[1]{}

\journal{Medical Image Analysis}

\begin{document}

\begin{frontmatter}

    \title{Non Local Spatial and Angular Matching : \\
           Enabling higher spatial resolution diffusion MRI datasets through adaptive denoising}

    \author{Samuel St-Jean\corref{sherbrooke}\fnref{sherbrooke}}
    \ead{samuel.st-jean@usherbrooke.ca}
    \author{Pierrick Coupé\fnref{bordeaux}}
    \author{Maxime Descoteaux\fnref{sherbrooke}}
    \address{This paper was published as Samuel St-Jean, Pierrick Coupé, Maxime Descoteaux, Non Local Spatial and Angular Matching: Enabling higher spatial resolution diffusion MRI datasets through adaptive denoising, Medical Image Analysis, Volume 32, August 2016, Pages 115-130, ISSN 1361-8415. doi:10.1016/j.media.2016.02.010}
    \fntext[sherbrooke]{Sherbrooke Connectivity Imaging Laboratory,
      Computer Science Department, Université de Sherbrooke}
    \fntext[bordeaux]{Unité Mixte de Recherche CNRS (UMR 5800), Laboratoire
       Bordelais de Recherche en Informatique, Bordeaux, France}

\begin{abstract}

Diffusion magnetic resonance imaging (MRI) datasets suffer from low Signal-to-Noise Ratio
(SNR), especially at high b-values. Acquiring data at high b-values contains
relevant information and is now of great interest for microstructural and
connectomics studies.
High noise levels bias the measurements due to the \review{non-Gaussian} nature of
the noise, which in turn can lead to a false and biased estimation of the
diffusion parameters.
Additionally, the usage of in-plane acceleration
techniques during the acquisition leads to a spatially
varying
noise distribution, \review{which depends on the parallel acceleration method
implemented on the scanner.}
\sout{with increased noise farther away from the receiver coils.}
This paper proposes a novel diffusion MRI denoising
technique that can be used on all existing data, without adding to the
scanning time.
We first apply a statistical framework to convert both stationary and non stationary
Rician and non central Chi
distributed noise to
Gaussian distributed noise, effectively removing the bias. We then introduce
a spatially and angular adaptive denoising technique, the
Non Local Spatial and Angular Matching (NLSAM) algorithm.
Each volume is first
decomposed in small 4D overlapping patches, thus capturing the spatial and
angular structure of the diffusion data, and a dictionary of atoms is learned on those patches.
A local sparse decomposition is then found by bounding the reconstruction error
with the local noise variance.
We compare against three other state-of-the-art denoising
methods and show quantitative local and connectivity results on a synthetic phantom and on an
\textit{in-vivo} high resolution dataset. Overall, our method restores perceptual
information, removes the noise bias in common diffusion metrics, restores
the extracted peaks coherence and improves
reproducibility of tractography on the synthetic dataset.
On the 1.2 mm high resolution \textit{in-vivo} dataset, our denoising improves
the visual quality of the data and reduces the number of spurious tracts when
compared to the noisy acquisition.
Our work paves the way for higher spatial
resolution acquisition of diffusion MRI datasets,
which could in turn reveal new
anatomical details that are not discernible at the spatial resolution
currently used by the diffusion MRI community.

\end{abstract}

\begin{keyword}

Diffusion MRI\sep
Denoising\sep
Block Matching\sep
Noise bias\sep
Dictionary learning%

\end{keyword}

\end{frontmatter}

\section{Introduction}

Diffusion magnetic resonance imaging (MRI) is an imaging technique that
allows probing microstructural features of the white matter architecture of
the brain. Due to the imaging
sequence used, the acquired images have an inherently low signal-to-noise
ratio (SNR), especially at high b-values. Acquiring data at high b-values
contains relevant information and is now of great interest for
connectomics~\citep{VanEssen2013} and microstructure~\citep{Alexander2010} studies.
High noise levels bias the measurements because
of the \review{non-Gaussian} nature of the noise, which in turn prohibits high resolution
acquisition if no further processing is done. This can also lead to a false and
biased estimation of the diffusion parameters, which impacts on the scalar
metrics (e.g. fractional anisotropy (FA)),
or in the fitting of various diffusion models (e.g. diffusion tensor
imaging (DTI) and high angular resolution diffusion imaging (HARDI)
models). This can further impact subsequent tractography and connectivity analysis
if the spatially variable noise bias is not taken into account.
Therefore, high SNR
diffusion weighted images (DWIs) are crucial in order to draw meaningful
conclusions in subsequent data or group analyses~\citep{Jones2012}.

This paper focuses on denoising techniques since they can be used on all
existing data, without adding to the scanning time. They also can be readily
applied to any already acquired dataset just like motion
and eddy current corrections that are commonly applied on acquired datasets.
One possible way to acquire higher quality data is to use better hardware, but
this is costly and not realistic in a clinical setting. One can also use a
bigger voxel size in order to keep the relative SNR at the same level, but at
the expense of a lower spatial resolution or acquiring fewer directions to keep
an acceptable acquisition time~\citep{Descoteaux2013}.
Averaging multiple acquisitions also
increases the SNR, but this should be done either using Gaussian distributed
noisy data~\citep{Eichner2015} or
in the complex domain to avoid the increased noise bias~\citep{Jones2012}.

With the advance of parallel imaging and acceleration techniques such as the
generalized autocalibrating partially parallel acquisitions (GRAPPA) or the
sensitivity encoding for fast MRI~(SENSE), taking into account the modified
noise distribution is the next step~\citep{Dietrich2008,Aja-Fernandez2014}. The
noise is usually modeled with a Rician distribution when SENSE is used
and a non central Chi (nc-$\chi$) distribution with 2N
degrees of freedom (with N the number of receiver coils)
\review{when a Sum of Squares (SoS) reconstruction
is used. If GRAPPA acceleration is also used with a SoS reconstruction, the
degrees of freedom of the nc-$\chi$ distribution will vary between 1 and 2N~\citep{Aja-Fernandez2014}.}
Some techniques have been specifically adapted by the medical imaging
community to take into account the Rician nature of the noise such as
non local means algorithms~\citep{Tristan-Vega2010,Coupe2008,Manjon2010a},
Linear Minimum Mean Square Error estimator~\citep{Aja-Fernandez2008a,Brion2013b},
generalized total variation~\citep{Liu2014},
a majorize-minimize framework with total variation denoising~\citep{Varadarajan2015},
maximum likelihood~\citep{Rajan2012}
or block matching~\citep{Maggioni2013a}.
Some methods
~\citep{Tristan-Vega2010,Brion2013b,Manjon2013,St-Jean2014,Bao2013a,Becker2014,Gramfort2013b,Lam2013a} %
have also been
specifically designed to take advantage of the properties of the diffusion MRI
signal such as symmetry, positivity or angular redundancy.
Since the data acquired in diffusion MRI depicts the same structural
information, but under different sensitizing gradients and noise realization,
these ideas take advantage of the information contained in the multiple
acquired diffusion MRI datasets.

We thus propose to exploit the structural redundancy across DWIs
through a common sparse representation using dictionary learning and
sparse coding to
reduce the noise level and achieve a higher SNR.
Our method can be thought of a Non Local Spatial and Angular
Matching (NLSAM) with dictionary learning. %
To the best of our knowledge, most recent state-of-the-art denoising algorithms
either concentrate on modeling the nc-$\chi$
noise bias or the spatially varying nature of the noise in a Rician setting.
Our method thus fills the gap by being robust to both of these aspects at the same
time, as seen in Table~\ref{tbl:denoising}.
We will compare our method against one structural MRI method and two other
publicly available algorithms :
the Adaptive Optimized Non Local Means (AONLM)~\citep{Manjon2010a}, which is designed
for 3D structural MRI,
the Local Principal Component Aanalysis (LPCA)~\citep{Manjon2013}
and the multi-shell Position-Orientation Adaptive Smoothing (msPOAS)
algorithm~\citep{Becker2014}, both designed for processing diffusion MRI datasets.
More information on each method
features and parameters will be detailed later. %

\begin{table}[bt]
\centering
\begin{tabular}{@{}lccccc@{}}
\toprule
\multicolumn{2}{c}{Noise type}                 & AONLM & LPCA & msPOAS & NLSAM \\ \midrule
\multirow{2}{*}{Stationary}     & Rician       & \yes  & \yes & \yes   & \yes  \\
                                & nc-$\chi$    & \no   & \no  & \yes   & \yes  \\
\addlinespace
\multirow{2}{*}{Variable}       & Rician       & \yes  & \yes & \no    & \yes  \\
                                & nc-$\chi$    & \no   & \no  & \no    & \yes  \\
\midrule
\multicolumn{2}{@{}l}{Use 4D angular information} & \no   & \yes & \yes   & \yes  \\
\bottomrule
\end{tabular}
\caption{Features of the compared denoising algorithm, see Section~\ref{sec:compared_algo}
for an in-depth review of each method. The NLSAM algorithm is the only technique robust to
both the spatially varying nature of the noise and the nc-$\chi$ bias at the same time.}
\label{tbl:denoising}
\end{table}

The contributions of our work are :

\begin{enumerate}[label=\textbf{\roman*})]
\itemsep0em
    \item Developing a novel denoising technique specifically tailored for diffusion MRI,
        which takes into account spatially varying Rician and nc-$\chi$ noise.
    \item Quantitatively comparing all methods on common diffusion MRI metrics. %
    \item Quantifying the impact of denoising on local reconstruction models.
    \item Analyzing the impact of denoising on tractography with a synthetic
    phantom and a high spatial resolution dataset.
\end{enumerate}

\FloatBarrier

\section{Theory}
\label{sec:theory}

\begin{figure}[hp!]
\small
\centering

            \begin{subfigure}[b]{0.75\textwidth}
                    \includegraphics[width=\textwidth]{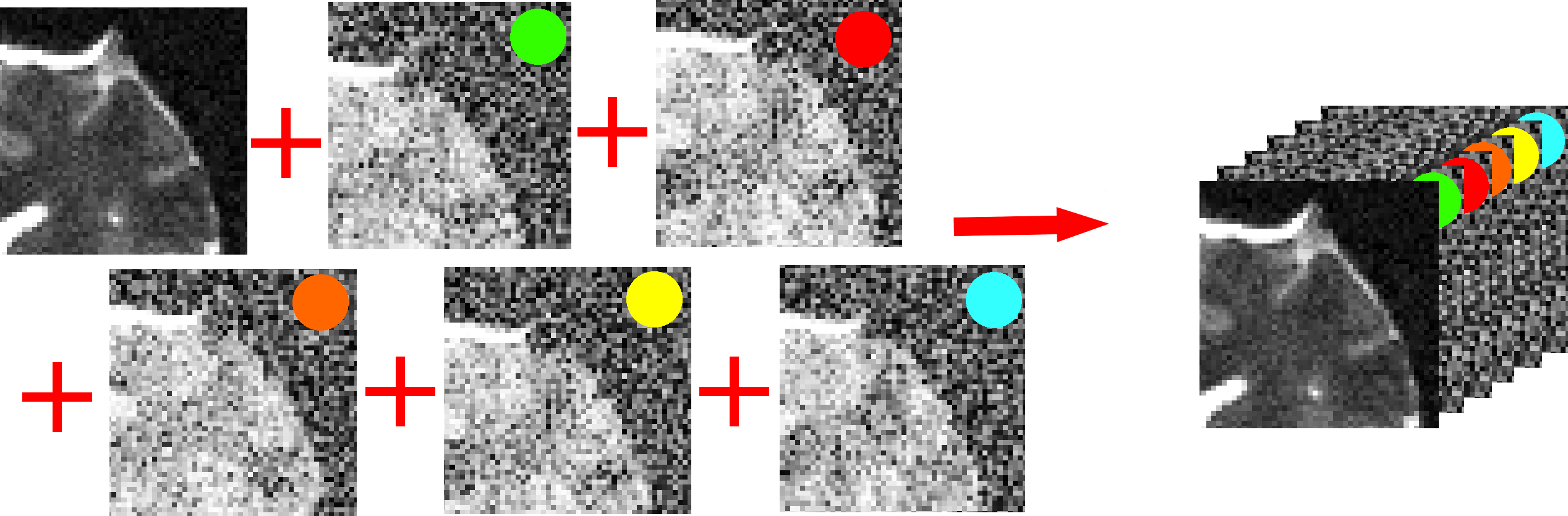}
                    \caption{A block is made of the b0 and some angular neighbors}
            \end{subfigure}
            \begin{subfigure}[b]{0.24\textwidth}
                    \includegraphics[width=\textwidth]{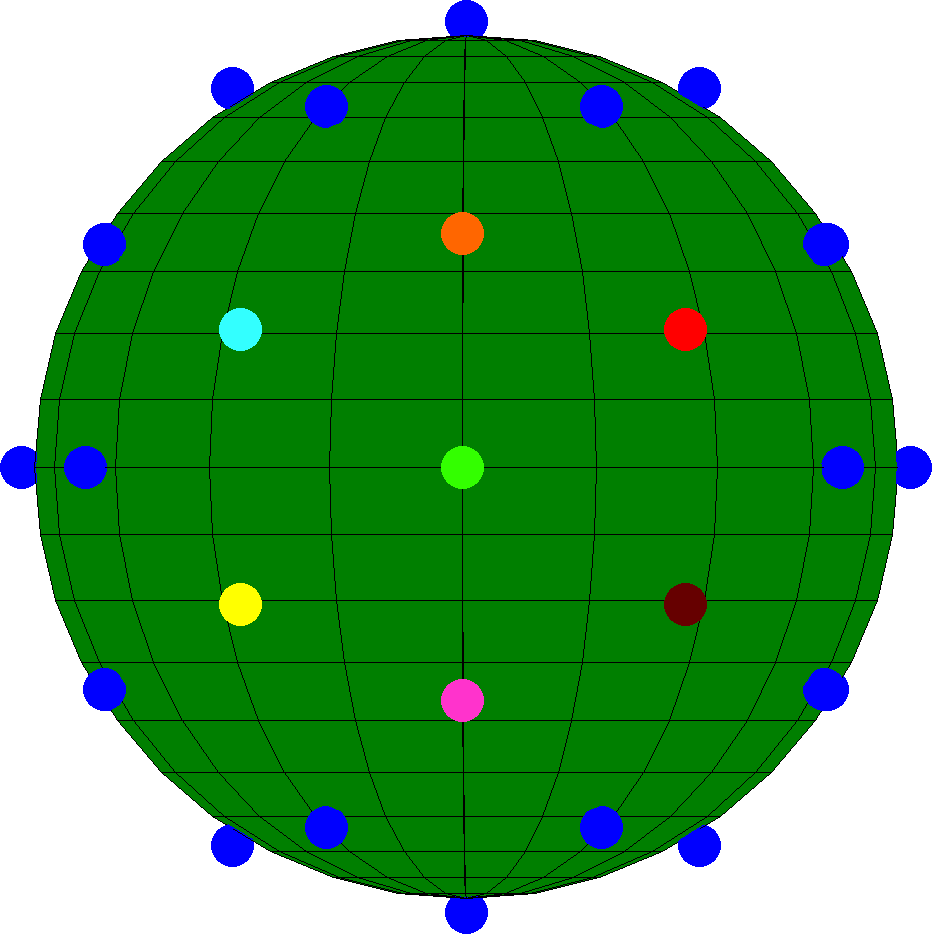}
                    \caption{Angular neighbors position on the sphere.}
            \end{subfigure}

    \caption{\textbf{a)} A 3D block is made by stacking along the 4th dimension
    the b0, a DWI and its angular neighbors, which share similar structure, but under a slightly
    different noise realization. \textbf{b)} Disposition of equidistant
    angular neighbors on the sphere.}

    \label{fig:stack}

\end{figure}

We now define two important terms used throughout the present work. Firstly, a
patch is defined as a 3D region of neighboring spatial voxels, i.e. a
small local region of a single 3D DWI. Secondly, a block is defined as
a collection of patches taken at the \emph{same} spatial position, but
in different DWIs, i.e. a block is a 4D stack of patches which are similar in
the angular domain. The reader is referred to Fig.~\ref{fig:stack} for a
visual representation of the process.

\paragraph{The Block Matching Algorithm}
Reusing the key ideas from the non local means, the
block matching algorithm~\citep{Dabov2007} further exploits image self-similarity.
Similar 2D patches found inside a local neighborhood are stacked into
a 3D transform domain and jointly filtered via wavelet hard-thresholding and
Wiener filtering. Combining these filtered estimates using a weighted average based
on their sparsity leads to superior denoising performance than the
non local means filter.
The idea has been extended in 3D for MRI image denoising in~\citep{Maggioni2013a} and an adaptive patch size version
for cardiac diffusion MRI image denoising was successfully employed by~\citep{Bao2013a}.
\paragraph{The Dictionary Learning Algorithm}

Dictionary learning has been used in the machine learning community to find
data driven sparse representations~\citep{Elad2006a,Mairal2009a}.
Typically, a set of atoms (called the dictionary) is learned over the data,
providing a way to represent it with a basis tailored to the signal
at hand~\citep{Olshausen1996}.
This is analogous to using an off-the-shelf basis like the discrete cosine
transform or wavelets, but in a
data-driven manner which gives better results than using a fixed, general purpose basis.
The main difference is that this is not necessarily a basis
in the sense it can also be overcomplete, i.e. it can have more atoms than
coefficients.
Given a set of input data $\mathbf{X} = \lbrack x_1, \dots, x_n \rbrack \in \mathbb{R}^{m \times n}$
organized as column vectors, the process is expressed as
\begin{equation}
    \label{eq:dl}
   \min_{\mathbf{D},\boldsymbol{\alpha}} \frac{1}{n} \sum_{i=1}^n \left(\frac{1}{2}
        \norm{x_i - \mathbf{D} \alpha_i}_2^2
        + \lambda \norm{\alpha_i}_1 \right) \enspace \text{s.t.} \enspace \norm{\mathbf{D}}_2^2 = 1 \enspace,
\end{equation}
\noindent
where $\mathbf{D} \in \mathbb{R}^{m \times p}$ is the learned dictionary, $\lambda$ is a trade-off
parameter between the data fidelity term and the penalization on the
coefficients $\boldsymbol{\alpha} = \lbrack \alpha_1, \dots, \alpha_n \rbrack \in \mathbb{R}^{p \times n}$.
A higher value of $\lambda$
promotes sparsity at the expense of the similarity with the original data.
The columns of $\mathbf{D}$ are also constrained to be of unit $\ell_2$ norm in
order to avoid degenerated solutions~\citep{Mairal2009a,Elad2006a,Gramfort2013b}.
The key is to devise a sparse representation
to reconstruct structural information and discard noise, since the latter does
not typically allow a sparse representation in any basis.
Using a penalization on the $\ell_1$ norm of the coefficients
promotes sparsity,
hence providing denoising through the regularized reconstruction.
This idea has led to inpainting and denoising applications from the
machine learning community~\citep{Mairal2009a,Elad2006a} or even to accelerated
acquisition process in the diffusion MRI community for
diffusion spectrum imaging (DSI)~\citep{Gramfort2013b}.

\paragraph{Adjusting for various noise types}

\begin{figure}[ht]
\small
\centering
    \begin{subfigure}{\textwidth}
            \includegraphics[width=\textwidth]{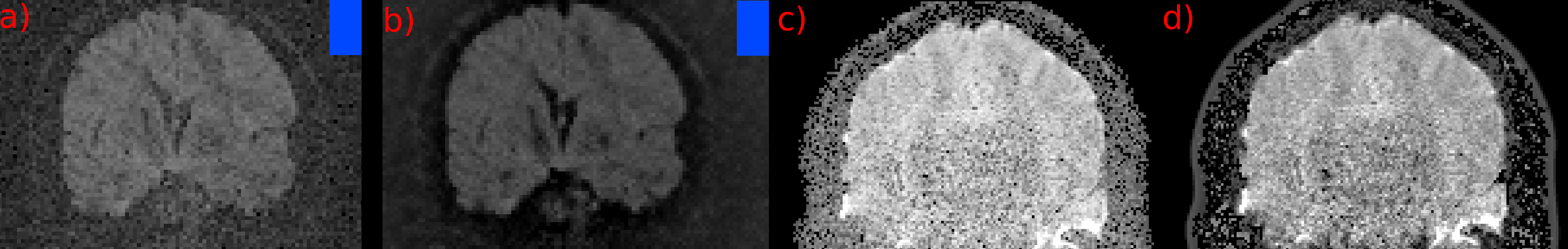}
    \end{subfigure}
    \begin{subfigure}[b]{\textwidth}
            \includegraphics[width=0.48\textwidth]{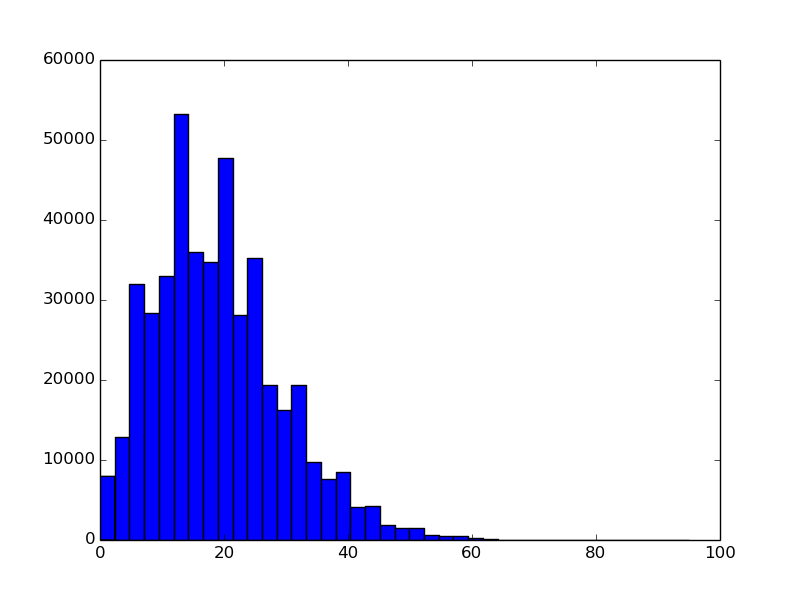}
            \includegraphics[width=0.48\textwidth]{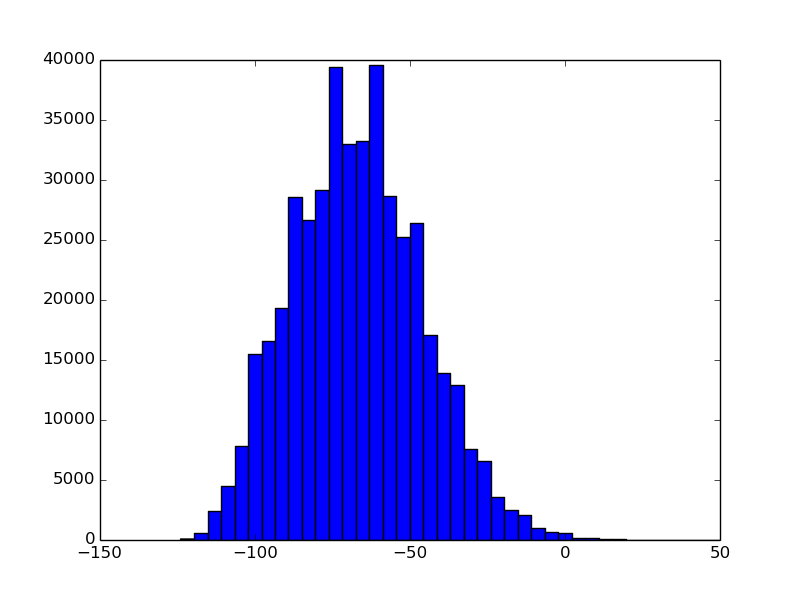}
    \end{subfigure}

     \caption{\review{\textbf{Top} : \textbf{a)} A noisy acquisition with slowly varying nc-$\chi$ noise
                and \textbf{b)} the resulting stabilized, Gaussian distributed noisy DWI.
           \textbf{c)} A noisy acquisition with fast varying Rician noise where the
                background was masked by the scanner with \textbf{d)} its stabilized counterpart.
        \textbf{Bottom} : Histogram of the nc-$\chi$ noise distribution in the selected background region of
        \textbf{a)} before stabilization and \textbf{b)} after stabilization.
        Note the non-Gaussianity of the noise in \textbf{a)} versus \textbf{b)}.}}
    \label{fig:stabilization}
\end{figure}

Although the original formulations of Eq.~(\ref{eq:dl}) and Eq.~(\ref{eq:dl_constrained}) assume
stationary, white additive Gaussian noise, this is usually not true in diffusion MRI data,
especially at high b-values and low SNR.
The noise is usually modeled as
following a Rician distribution or a nc-$\chi$ distribution
when used with parallel imaging depending on the reconstruction algorithm
and the number of coils N used during the acquisition~\citep{Dietrich2008,Aja-Fernandez2014}.
This introduces a bias which depends on the intensity of the signal that must be
taken into account to recover the expected value
of the original signal as shown in Fig.~\ref{fig:stabilization}.
\review{Note, though, that other common preprocessing corrections,
such as motion correction or eddy current correction, require interpolation
and could thus change the theoretical noise distribution~\citep{Veraart2013b}.}

\FloatBarrier

\begin{figure}[bht]
\small
\centering
    \begin{subfigure}[b]{0.32\textwidth}
            \includegraphics[width=5.2cm]{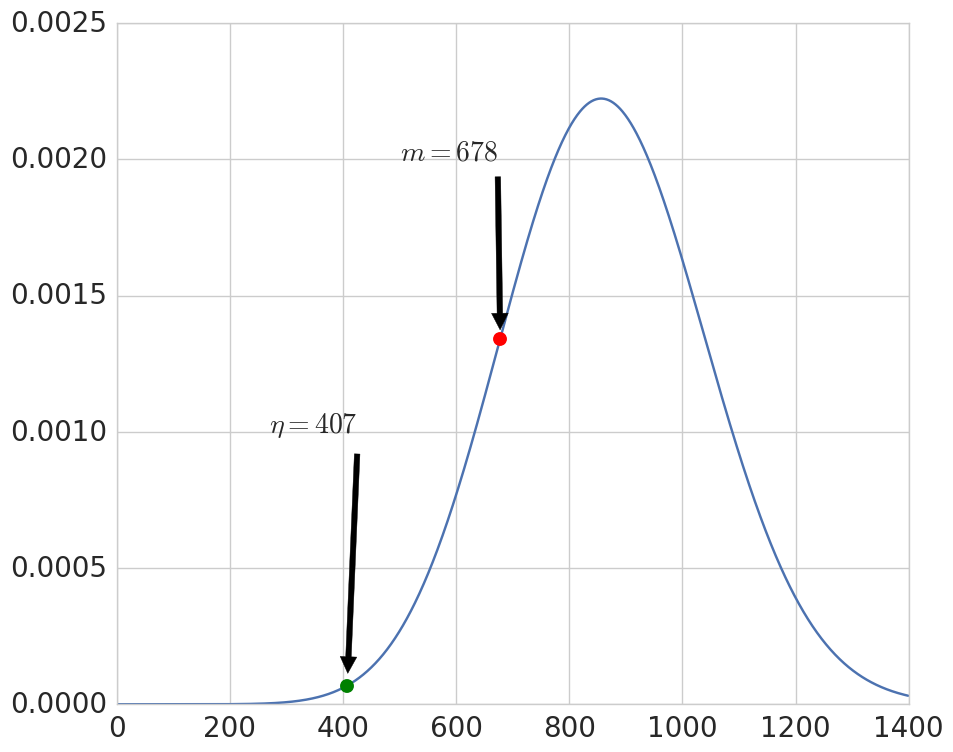}
            \caption{pdf of nc-$\chi$ distribution.}
    \end{subfigure}
    \begin{subfigure}[b]{0.32\textwidth}
            \includegraphics[width=5.2cm]{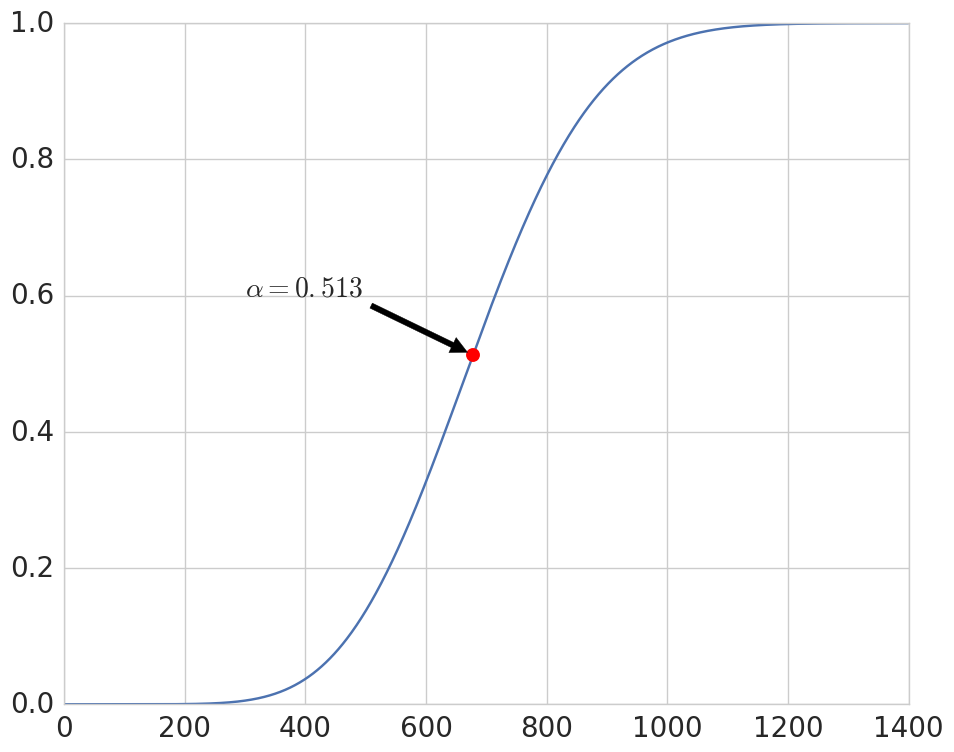}
            \caption{cdf of nc-$\chi$ distribution.}
    \end{subfigure}
    \begin{subfigure}[b]{0.32\textwidth}
            \includegraphics[width=5.2cm]{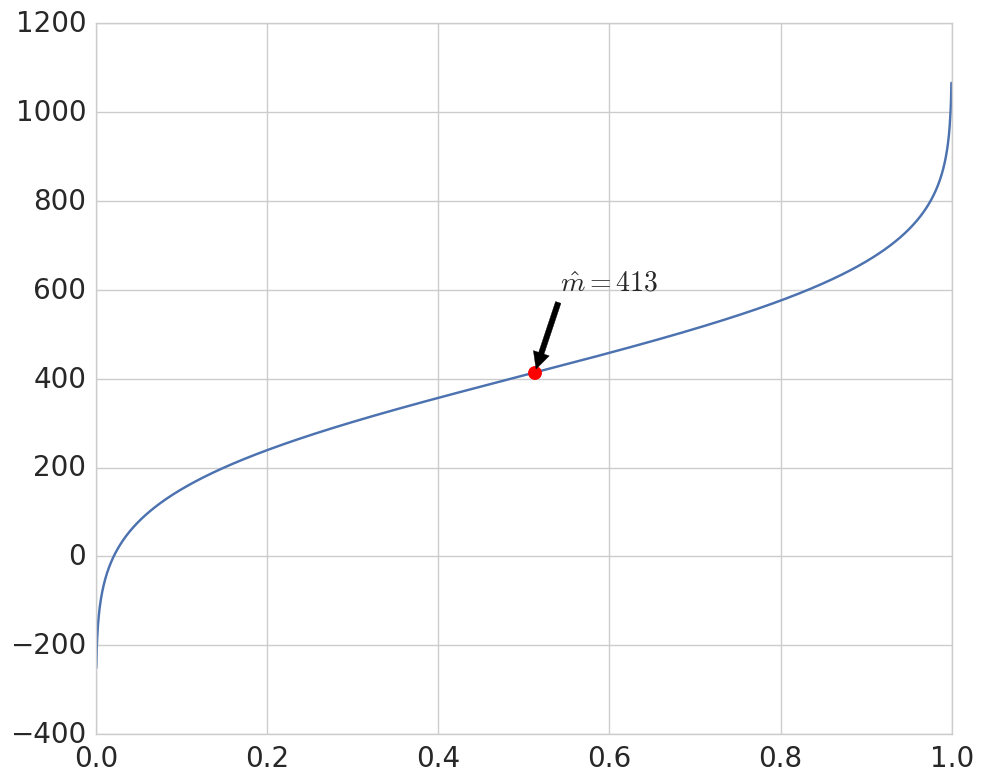}
            \caption{icdf of Gaussian distribution.}
    \end{subfigure}
    \caption{\review{A synthetic example of the stabilization algorithm.}
    \textbf{a)} Given a noisy value $m = 678$ observed in a nc-$\chi$ distribution
    with $N=4$ and \review{$\sigma_G=200$}, the underlying value is estimated as $\eta=407$.
    \textbf{b)} The associated probability in the nc-$\chi$ cdf with $\eta$ is $\alpha=0.513$,
    \textbf{c)} thus giving \review{from the inverse cdf of a Gaussian distribution}
    \review{with} mean $\review{\mu} = 407$ and standard deviation \review{$\sigma_G=200$}
    a new noisy value $\hat{m}=413$.}
    \label{fig:graph_stabilization}
\end{figure}

The key idea lies in the fact that the nc-$\chi$ distribution is actually
made from a sum of Gaussians, from which the Rician distribution is a special
case with N = 1. By making the hypothesis that each of the 2N Gaussian
distributions share the same standard deviation $\sigma_G$~\citep{Koay2009a},
one can map a value $m$ from a
nc-$\chi$ distribution to an equivalent value $\hat{m}$ from a Gaussian
distribution. We first compute estimates for $\sigma_G$ and $\eta$ (which is
an estimate of the signal value in a Gaussian setting).
If $\eta$ is below the noise floor due to a low local SNR,
that is when $\eta < \sigma_G \sqrt{\pi / 2}$,
we set $\eta = 0$ instead of being negative as suggested by~\citep{Bai2014}.
The next step uses the cumulative distribution function (cdf)
of a nc-$\chi$ distribution and the inverse cumulative distribution function
(icdf) of a Gaussian distribution to find the equivalent value $\hat{m}$ between
the two distributions. This effectively maps a \emph{noisy} nc-$\chi$ distributed signal
$m$ to an equivalent \emph{noisy} Gaussian distributed signal $\hat{m}$. See
Fig.~\ref{fig:graph_stabilization} for \review{a synthetic example}
with a visual depiction of the
process for mapping nc-$\chi$ signals to Gaussian distributed signals
and~\citep{Koay2009a} for the original in-depth details.

Using a variance stabilization means considering the noise as additive white Gaussian noise,
which allows any already designed technique for Gaussian noise to be
used without any modification. The author of~\citep{Foi2011} has shown
that techniques with a Rician noise adaptation performed
equally well as their Gaussian noise version through the use of a noise
stabilization approach. The same idea has been directly applied with block matching
~\citep{Dabov2007} for structural MRI in~\citep{Maggioni2013a}.
The classical solution to remove the noise bias is to include the noise model
into the denoising algorithm itself, as for example done in
~\citep{Aja-Fernandez2008a,Becker2014,Lam2013a,Manjon2010a}. %
The drawback with this solution is that each method has to be rethought
to account for any other noise type not considered in its original formulation.

\FloatBarrier

\section{Method}
\label{sec:method}

\paragraph{Adjusting for various noise types}

In this paper, we will deal with both the Rician and nc-$\chi$ noise
model on a voxelwise basis through the noise stabilization technique of~\citep{Koay2009a}.
This indeed makes our algorithm easily adaptable for any noise type
by simply changing the pre-applied transformation as needed.
We will use the Probabilistic Identification and Estimation of
Noise (PIESNO)~\citep{Koay2009b} to estimate the stationary noise standard deviation.
PIESNO works on a slice by slice basis, assuming the background noise as stationary
along the selected slice, and is designed to find the underlying standard
deviation of the Gaussian noise given its Rician or nc-$\chi$ nature.
Voxels that are considered as pure background noise are found automatically by the method,
using the fact that the squared mean of those voxels follows a Gamma distribution.
Once automatically identified, the standard deviation $\sigma_G$
of those voxels can be computed and a new estimation of the Gamma distribution
is made with the updated $\sigma_G$ until convergence.
In the case of spatially varying noise, we will use a method similar
to~\citep{Manjon2010a}, where the noise is estimated locally as
\begin{equation}
    \review{\sigma^2_i} = \min ||u_i - u_j||_2^2, \forall i \ne j,
\end{equation}
with $u_i$ a noisy patch computed
by subtracting a patch to a low-pass filtered version of itself and applying
the local Rician correction factor of~\citep{Koay2006}.
If the background was masked automatically by the scanner
or is unreliable due to the scanner preprocessing for statistical estimation,
we use a similar idea by
computing the local standard deviation of the noise field as
\begin{equation}
    \review{\sigma_i = \text{std}(u_i - \text{low\rule{1ex}{.4pt}pass}(u_i))}
    \label{eq:noise_field}
\end{equation}
If a noise map was acquired during the scanning session,
it can be sampled directly
to estimate the parameters of the noise distribution.
In the event that such a map is unavailable, a synthetic one can be constructed by subtracting
the image from its low-pass filtered counterpart (see Eq.~\ref{eq:noise_field}).
Since the noise is assumed
as independent and identically distributed across DWIs, we apply a median filter
on the 4D dataset to get a 3D noise field. Finally, a Gaussian filter
with a full-width at half maximum of 10 mm is applied to regularize the noise field,
which is then corrected for the more general nc-$\chi$ bias with
the correction factor of~\citep{Koay2006}.
A similar approach based on
extracting the noise field with a principal component analysis was used by~\citep{Manjon2013}.

\paragraph{Locally Adapting the Dictionary Learning}
\phantomsection
\label{sec:local_adapt}

In order to locally adapt the method to spatially varying noise,
we add some more constraints to the classical formulation of Eq.~(\ref{eq:dl}).
Firstly, since the measured signal in diffusion MRI is always positive,
we use this assumption to constrain the positivity of the global dictionary
$\mathbf{D}$ and the coefficients $\boldsymbol{\alpha}$,
i.e. $\mathbf{D} \ge 0, \boldsymbol{\alpha} \ge 0$ as done in~\citep{Gramfort2013b}.
We fixed the regularization parameter $\lambda$ for Eq.~(\ref{eq:dl})
in the same fashion as~\citep{Mairal2009a}, that is
$\lambda = 1.2 / \sqrt{m}$, with $m = ps^3 \times an$, $ps$ is the patch size and $an$
the number of angular neighbors.
Secondly, once $\mathbf{D}$ is known, we use Eq.~(\ref{eq:dl_constrained})
\review{(see the next paragraph)}
iteratively until convergence with the constraint
$\boldsymbol{\alpha} \ge 0$ and $\lambda_i = \sigma_i^2 (m + 3 \sqrt{2m})$,
where \review{$\sigma^2_i$} is the local noise variance found either with PIESNO or
Eq.~(\ref{eq:noise_field}).
\review{In accordance with \citep{Candes2008}, $\lambda_i$ is an upper bound on the $\ell_2$
norm of the noise.}
We set the convergence as reached \review{for $\alpha_i$ at iteration $j$}
when $\max |\review{\alpha_{i, j} - \alpha_{i, j-1}}| < 10^{-5}$
or until a maximum of 40 iterations is realized.
\paragraph{Adaptive and Iterative $\ell_1$ Minimization}

While Eq.~(\ref{eq:dl}) will both construct the dictionary $\mathbf{D}$ and find
the coefficients $\boldsymbol{\alpha}$, there are specialized iterative algorithms for
solving $\ell_1$ problems in order to yield sparser
solutions~\citep{Daubechies2010,Candes2008}.
An equivalent constrained formulation for solving each column $i$ of $\boldsymbol{\alpha}$ is
\begin{equation}
    \label{eq:dl_constrained}
   \min_{\alpha_i} \norm{w_{j} \alpha_i}_1  \enspace \text{s.t.} \enspace \frac{1}{2}
         \norm{x_i - \mathbf{D}\alpha_i}_2^2 \le \lambda_i,
\end{equation}
\noindent
where $w_{j}$ is a weighting \review{vector} penalizing the coefficients of $\alpha_i$
at iteration $j$. Eq.~(\ref{eq:dl_constrained}) can thus be iterated to further
identify non zero coefficients in $\alpha_i$ by setting
$w_{j+1} = \frac{1}{\review{\abs{\alpha_{i}}} + \epsilon}$
for the next iteration. The algorithm is then started with $w_0 = 1$ and
$\epsilon = \max\abs{\mathbf{D}^T \xi}$.
As suggested by~\citep{Candes2008},
$\xi \sim \mathcal{N}(0,  \sigma^2)$ is set as a random Gaussian vector, which gives a
baseline where significant signal components might be recovered.
While similar in spirit to Eq.~(\ref{eq:dl}), Eq.~(\ref{eq:dl_constrained}) provides a way to
find the sparser representation for $\alpha_i$ while
bounding the $\ell_2$ reconstruction error.

\sout{We will use this formulation
to find the best reconstruction bounded by the local variance of the noise by setting
$\lambda_i = \sigma_i^2 (m + \gamma \sqrt{2m})$ as in Candes2008
for finding $\alpha_i$ once $\mathbf{D}$ is known,
where $\sigma_i^2$ is the local noise variance and $\gamma$ a constant.}

To the best of our knowledge, \review{our paper is also the first
to use the noise variance} as an explicit bound on the $\ell_2$ reconstruction
error. This yields a sparse representation while controlling at the same time
the fidelity with respect to the original data, while the classical
way is to use the variance as a regularized penalization factor.

\subsection{The proposed algorithm}

Our new NLSAM algorithm combines ideas from block matching and sparse coding.
We will use the same kind of framework, but by replacing
the thresholding part in the block matching with a step of dictionary
denoising instead, allowing the penalization on the sparsity of the signal to
regularize the noisy blocks. We also take explicit advantage of the fact that
diffusion MRI data is composed of multiples volumes of the \emph{same}
structure, albeit with different noise realizations and contrasts across DWIs.
This allows sparser estimates to be found, further enhancing the separation of
the data from the noise~\citep{Olshausen1996}.
Our method is thus composed of three steps :

\begin{enumerate}

    \item Correct the noise bias \review{if needed}.

    \item Find angular neighbors on the sphere for each DWI.

    \item Apply iterative local dictionary denoising on each subset of neighbors.

\end{enumerate}

\textbf{Step 1.}
\review{In case the noise is not Gaussian distributed,}
we first correct for the noise bias by finding the Gaussian
noise standard deviation with PIESNO~\citep{Koay2009b}. If the background is masked,
we instead use Eq.~(\ref{eq:noise_field}).
We then transform the DWIs into Gaussian distributed,
noisy signals using the correction scheme of~\citep{Koay2009a}.

\textbf{Step 2.}
We find the angular neighbors for each of the DWIs.
In this step, the local angular information is encoded in a 4D
block of similar angular data, as seen in Fig.~\ref{fig:stack}.
\review{The gradients are symmetrized to account for opposite polarity DWIs, which share
similar structure to their symmetrized counterpart. The search is also made along
all the shell at the same time, since structural information
(such as sharp edges) is encoded along the axial part of the data.}
This encodes the similar angular structure of the data along the 4th dimension
in a single vector.

\textbf{Step 3.}
The dictionary $\textbf{D}$ is constructed with Eq.~(\ref{eq:dl})
and the blocks are then denoised with Eq.~(\ref{eq:dl_constrained}).
This step can be thought of finding a linear combination with the smallest
number of atoms to represent a block.
In order to adapt to spatially varying noise, each
block is penalized differently based on the local variance of the noise.
This enables the regularization to adapt to the amount of noise
in the block, which is usually stronger as the acquired signal
is farther from the receiver coils.
Since each overlapping block is extracted, each voxel is represented many times
and they are recombined using
a weighted average based on their sparsity as in~\citep{Manjon2013,Maggioni2013a}.
For each voxel \review{$i$ with intensity $v_i$}
contained in multiple overlapping blocks $V_j$ in neighborhood $V$,
we set the final value of $v_i$ as
\begin{equation}
    v_i =  \frac{\sum\limits_{j \in V} v_j (1+||V_j||_0)}{\sum\limits_{j \in V} 1 + ||V_j||_0},
    \label{eq:w_avg}
\end{equation}
where $V$ is the same spatial position for voxel $i$ across multiple blocks $V_j$.
This assumes that more coefficients in block $V_i$ also means more noise in the reconstruction.
The $\ell_0$ norm thus penalizes reconstructions with more coefficients and assigns
a lower weight in that case for the overlapping weighted average.

This third step is then repeated for all the DWIs. Since each DWI will
be processed more than once with a different set of neighbors each time
(see Fig.~\ref{fig:stack} for the block formation process), we obtain multiple
denoised volumes of \emph{exactly} the same data, but denoised in a different
angular context. Once all the DWIs have been processed, we average the multiple denoised
versions obtained previously in order to further reduce any residual noise.
See~\ref{sec:appendix} for an outline of the NLSAM algorithm as pseudocode.
The result will be a denoised version of the input, through
both dictionary learning and spatial and angular block matching.

\FloatBarrier
\subsection{Datasets and acquisition parameters}
\label{sec:experiments}

\begin{figure}[tbh]
\small
\centering

    \raisebox{1.5cm}{\rotatebox[origin=t]{90}{b0 image}}
    \begin{subfigure}[b]{0.19\textwidth}
            \includegraphics[width=\textwidth]{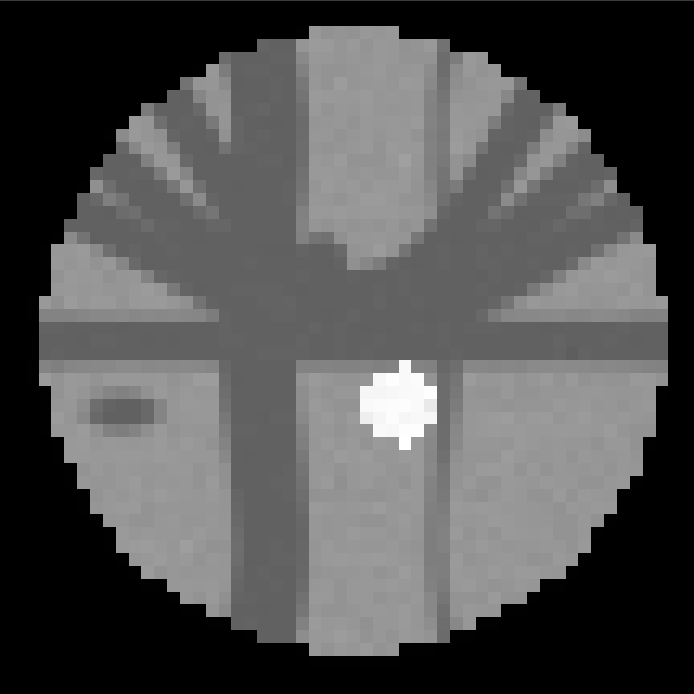}
    \end{subfigure}
    \begin{subfigure}[b]{0.19\textwidth}
            \includegraphics[width=\textwidth]{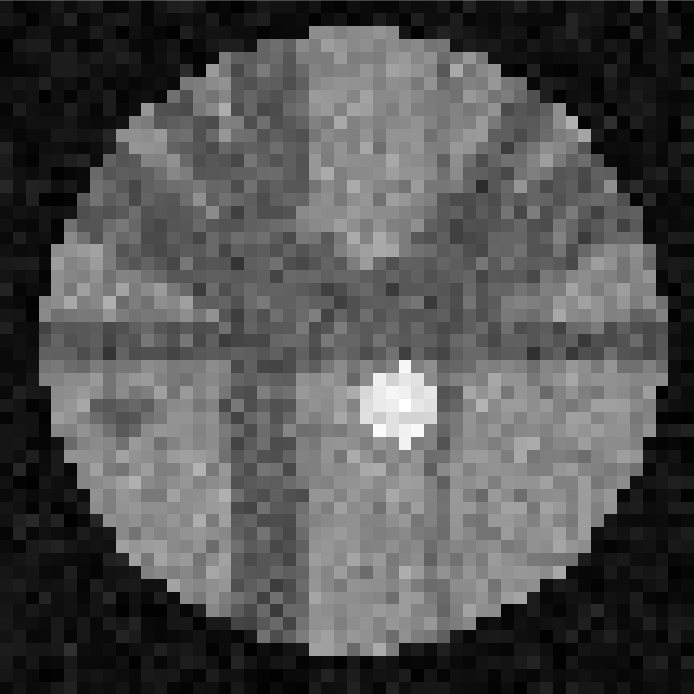}
    \end{subfigure}
    \begin{subfigure}[b]{0.19\textwidth}
            \includegraphics[width=\textwidth]{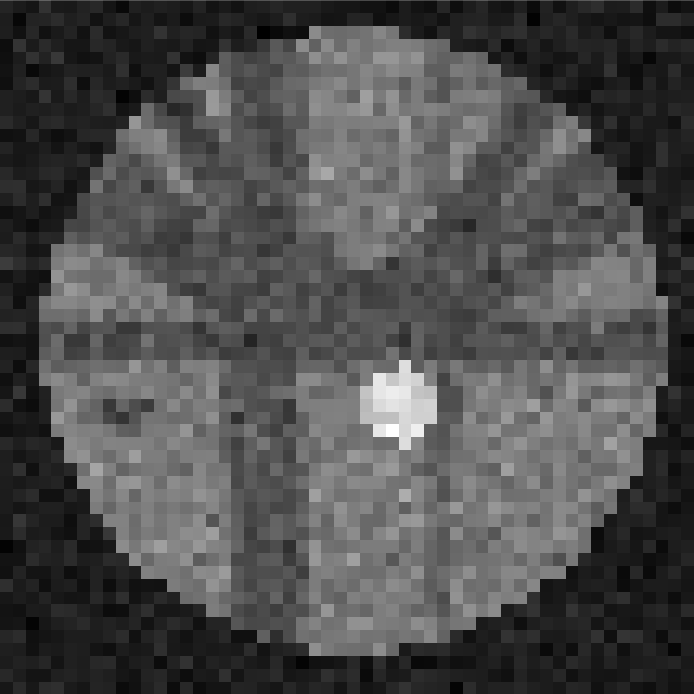}
    \end{subfigure}
    \begin{subfigure}[b]{0.19\textwidth}
            \includegraphics[width=\textwidth]{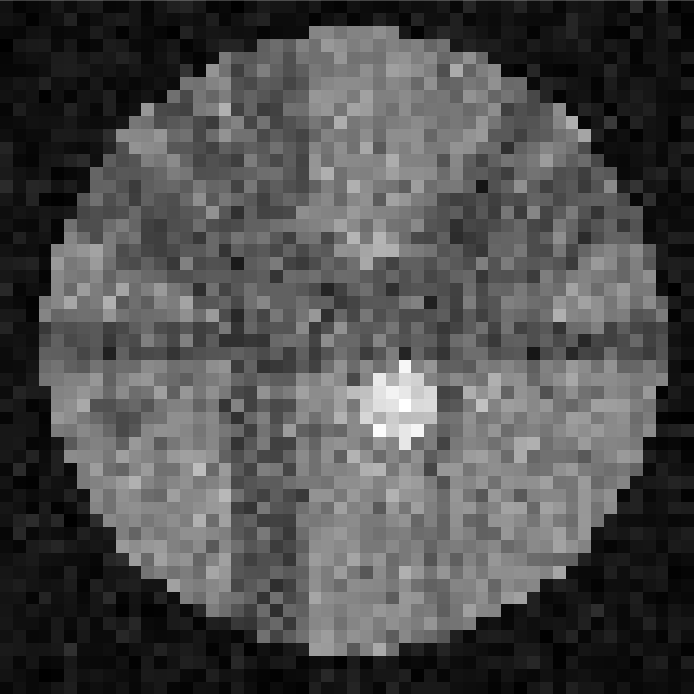}
    \end{subfigure}
    \begin{subfigure}[b]{0.19\textwidth}
            \includegraphics[width=\textwidth]{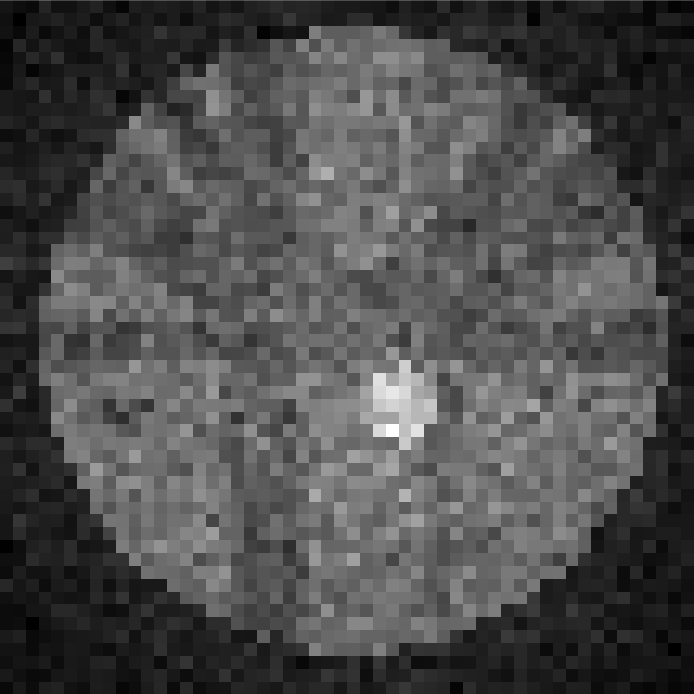}
    \end{subfigure}

    \raisebox{1.5cm}{\rotatebox[origin=t]{90}{DW image}}
    \begin{subfigure}[b]{0.19\textwidth}
            \includegraphics[width=\textwidth]{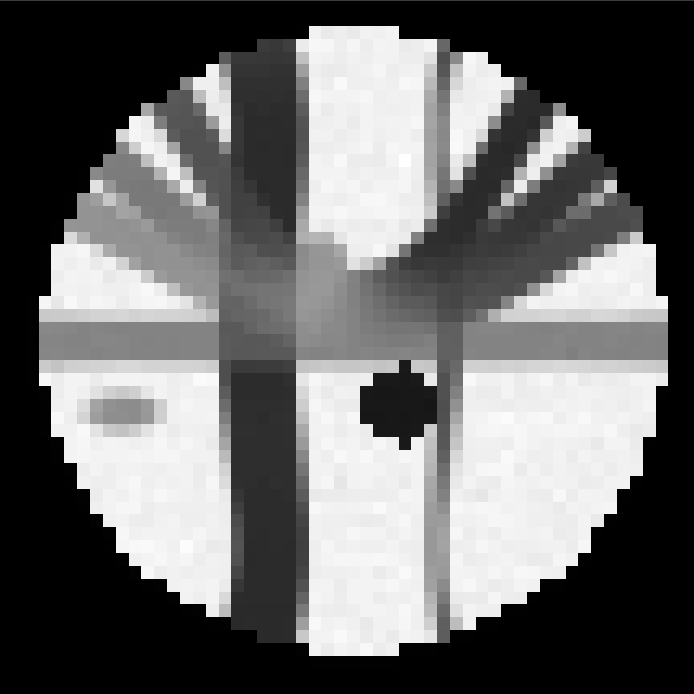}
    \end{subfigure}
    \begin{subfigure}[b]{0.19\textwidth}
            \includegraphics[width=\textwidth]{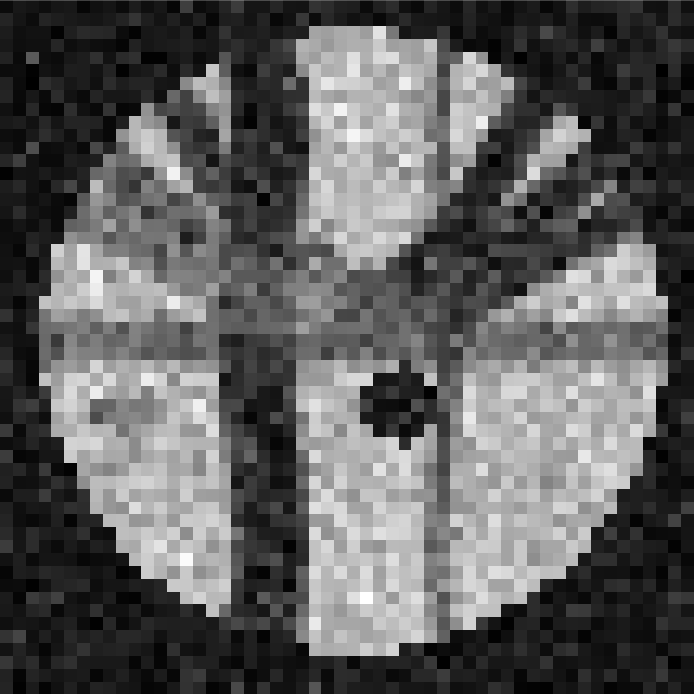}
    \end{subfigure}
    \begin{subfigure}[b]{0.19\textwidth}
            \includegraphics[width=\textwidth]{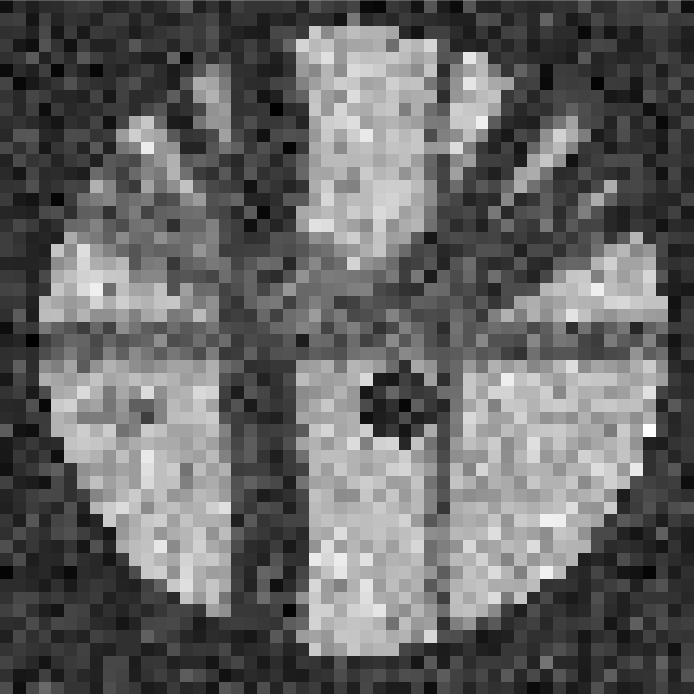}
    \end{subfigure}
    \begin{subfigure}[b]{0.19\textwidth}
            \includegraphics[width=\textwidth]{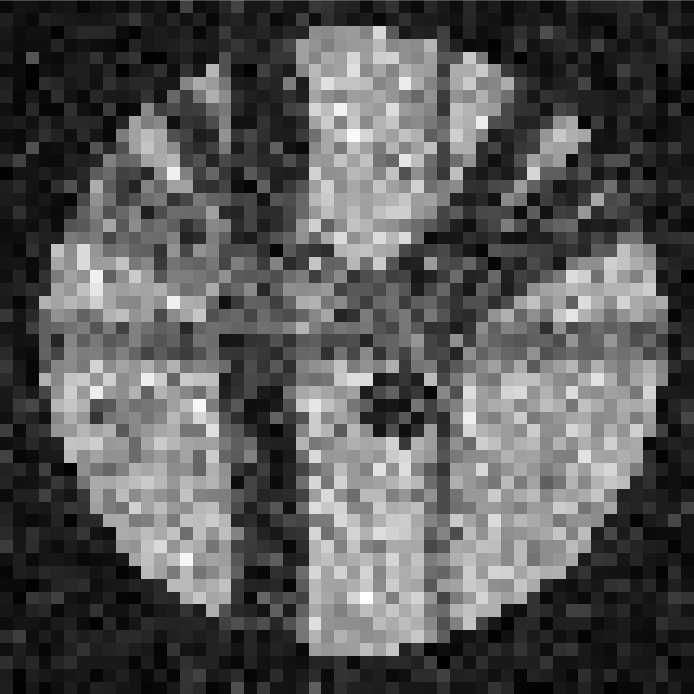}
    \end{subfigure}
    \begin{subfigure}[b]{0.19\textwidth}
            \includegraphics[width=\textwidth]{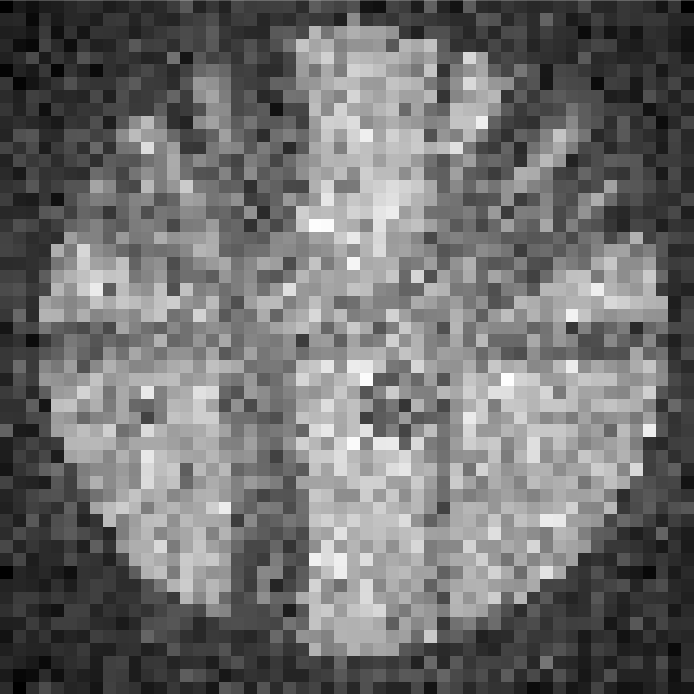}
    \end{subfigure}

    \raisebox{2cm}{\rotatebox[origin=t]{90}{Added noise \phantom{g}}}
    \begin{subfigure}[b]{0.19\textwidth}
            \includegraphics[width=\textwidth]{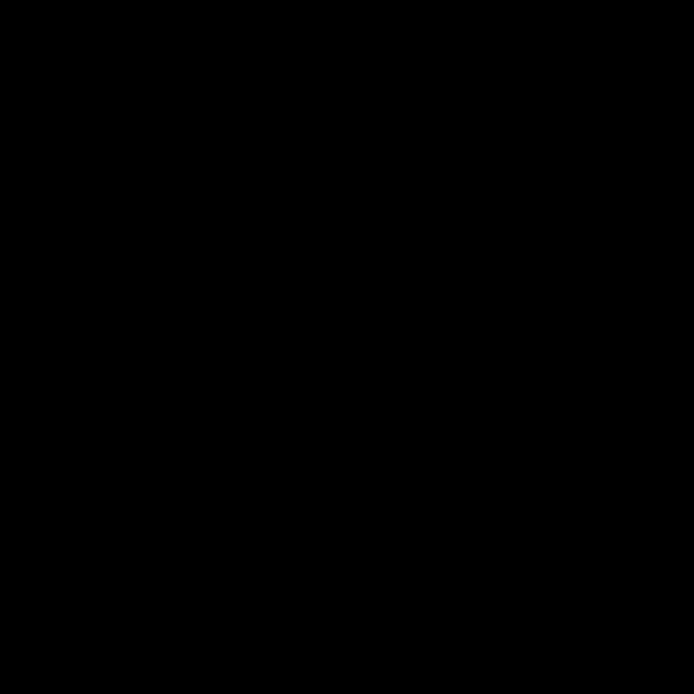}
            \caption{Noiseless}
    \end{subfigure}
    \begin{subfigure}[b]{0.19\textwidth}
            \includegraphics[width=\textwidth]{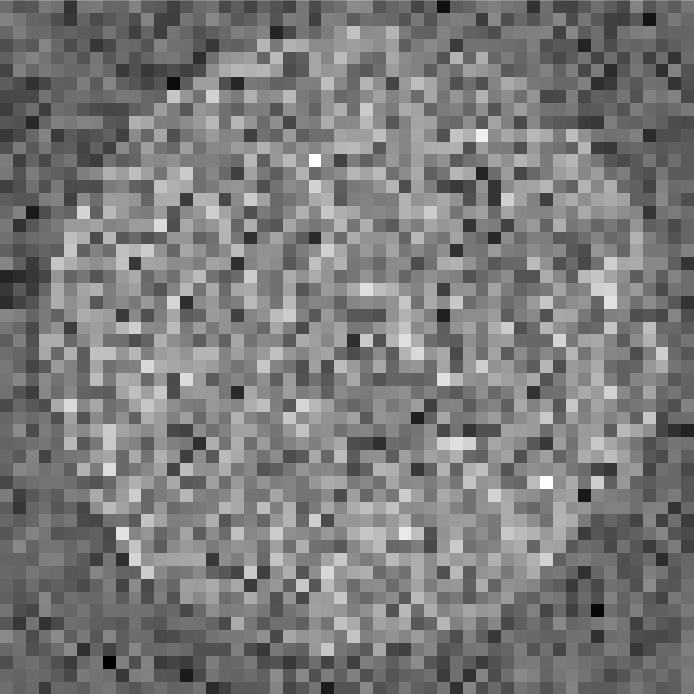}
            \caption{SNR10 stat Rician}
    \end{subfigure}
    \begin{subfigure}[b]{0.19\textwidth}
            \includegraphics[width=\textwidth]{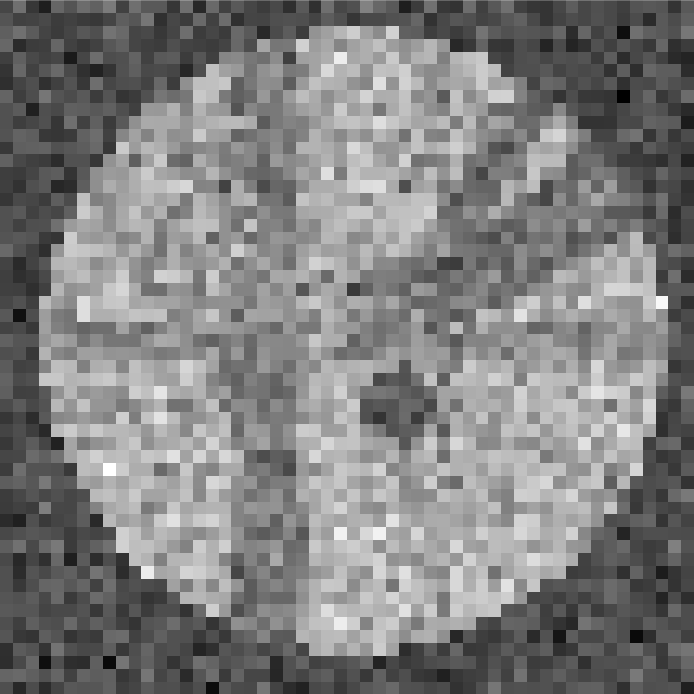}
            \caption{SNR10 stat nc-$\chi$}
    \end{subfigure}
    \begin{subfigure}[b]{0.19\textwidth}
            \includegraphics[width=\textwidth]{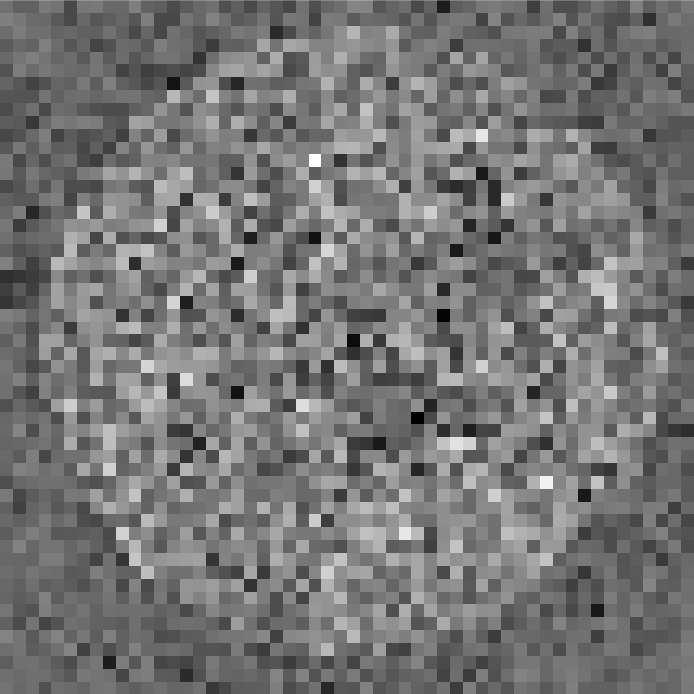}
            \caption{SNR15 var Rician}
    \end{subfigure}
    \begin{subfigure}[b]{0.19\textwidth}
            \includegraphics[width=\textwidth]{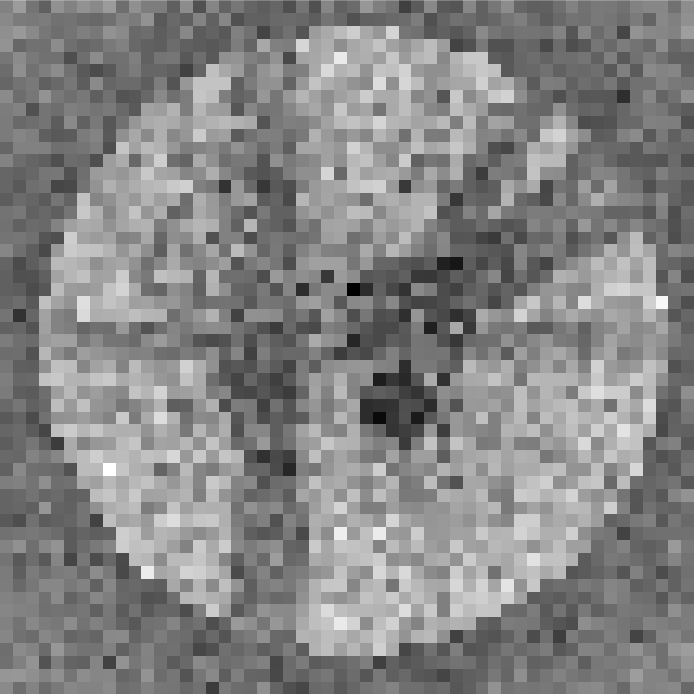}
            \caption{SNR15 var nc-$\chi$}
    \end{subfigure}

    \caption{Synthetic b1000 datasets with various noise profiles used in the experiments.
    The top row shows the b0 image, the middle row shows the same DWI across noise types
    and the bottom row shows the various noise distribution which generated the middle row.
    From left to right : the noiseless data,
    SNR 10 with stationary Rician noise, SNR 10 with stationary nc-$\chi$ noise,
    SNR 15 with spatially variable Rician noise, SNR 15 with spatially varying nc-$\chi$ noise.}
    \label{fig:spat_noise}
\end{figure}

\paragraph{Synthetic phantom datasets}
The synthetic data simulations are based on the ISBI 2013 HARDI challenge
phantom\footnote{\url{http://hardi.epfl.ch/static/events/2013_ISBI/}} and were
made with phantomas\footnote{\url{http://www.emmanuelcaruyer.com/phantomas.php}}.
We used the given 64 gradients set from the challenge at b-values of 1000 and
3000 s/mm$^2$. For simplicity, we will now refer to these datasets as the
b1000 and the b3000 datasets.
\review{The datasets were generated with} Rician and nc-$\chi$ noise profile,
both stationary and spatially varying, \sout{to both datasets} at two
different signal-to-noise ratios (SNR) for each case.
In total, we thus have 8 different noise profiles for each b-value.
The stationary noise was generated with SNR 10 and 20 and the
spatially varying noise was generated with SNR varying linearly
from 5 to 15 and from 7 to 20.
The noise distributions were generated for each SNR by setting
$N = 1$ for the Rician noise and $N = 12$ for the nc-$\chi$ noise.
The \review{noisy data} was generated according to
\begin{equation}
    \label{eq:noising}
    \hat{I} = \sqrt{\sum_{i=0, j=0}^{N} \left(\frac{I}{\sqrt{N}} + \beta\epsilon_i \right)^2 + \beta\epsilon_j^2},
    \enspace \mbox{where } \epsilon_i, \epsilon_j \sim \mathcal{N}(0,  \sigma^2),
\end{equation}
\noindent
where $\hat{I}$ is the resulting noisy volume, $\mathcal{N}(0,  \sigma^2)$
is a Gaussian distribution of mean 0 and variance $\sigma^2$ with
$\sigma$ = mean (\mbox{b0}) / \mbox{SNR} and mean(b0) is the mean signal value of
the b $= 0$ s/mm$^2$ image.
$\beta$ is a mask \sout{set to 1} to create the
\sout{stationary} noise distribution \review{set to 1 in the constant noise case}
and \review{as} a sphere \sout{with values varying
linearly from 1 at the borders to 3 at the center} for the
spatially varying noise case.
\review{For the spatially varying noise experiments,
$\beta$ has a value of 1 on the borders up to a value of 3
at the middle of the mask,}
thus generating a stronger noise profile near the middle of the phantom than for the stationary (constant) noise case.
As shown on Fig.~\ref{fig:spat_noise},
this results in a variable SNR ranging from approximately SNR 5 and SNR 7 in the middle of the phantom up to
SNR 15 and SNR 20 for the spatially varying noise case.
This noise mimics a homogeneous noise reconstruction as implemented by some scanners
while still having a spatially varying noise map.

\paragraph{Real datasets}
In order to compare our NLSAM method on a real dataset,
we acquired a full brain \textit{in-vivo} dataset consisting of 40 DWIs at
$\mbox{b} = 1000 \mbox{ s/mm}^2$ and one $\mbox{b} = 0 \mbox{ s/mm}^2$.
The acquisition spatial resolution was
$1.2 \times 1.2 \times 1.2 \mbox{ mm}^3$, TR/TE = 18.9 s / 104 ms,
gradient strength of 45 mT/m
on a 3T Philips Ingenia scanner with a 32
channels head coil for a total acquisition time of 13 minutes.
An in-plane parallel imaging factor of R = 2 was used with the SENSE reconstruction algorithm,
thus giving a fast spatially varying Rician noise distribution
(hence, the denoising algorithms will be set with $N = 1$)
even if multiple coils are used by the reconstruction algorithm
for producing the final image (see Fig.~\ref{fig:stabilization}).
No correction was applied to the dataset, as we wanted to show the effectiveness
of denoising without \review{any other preprocessing step
such as eddy current or motion correction, which could introduce blurring caused by interpolation.}
\sout{risking introducing blurring caused by interpolation
in other methods such as eddy current or motion correction.}
To obtain a comparable clinical-like baseline dataset and show the advantage of acquiring directly
high resolution DWIs, we also obtained a 64 DWIs dataset at
$\mbox{b} = 1000 \mbox{ s/mm}^2$ and one $\mbox{b} = 0 \mbox{ s/mm}^2$ of the same subject.
The spatial resolution was $1.8 \times 1.8 \times 1.8 \mbox{ mm}^3$,
TR/TE = 11.1 s / 63 ms, for a total acquisition time of 12 mins. %
The acquisition was made on the same scanner, but during another scanning session.
No further processing nor denoising was done on this dataset for the reasons mentioned above.
This can be thought of having a higher angular resolution at the cost of a lower
spatial resolution for a comparable acquisition time.
\subsection{Other denoising algorithms for comparison}
\label{sec:compared_algo}

We now present the various features and cases covered by the denoising algorithms
studied in this paper.
The Adaptive Optimized Nonlocal Means (AONLM) method~\citep{Manjon2010a} is
designed for Rician noise removal in a 3D fashion and works
separately on each DWIs volume. It also includes a Rician bias removal step
and is able to spatially adapt to a varying noise profile
automatically. We used the recommended default parameter of a 3D patch size of
3 x 3 x 3 voxels with the Rician bias correction in all cases.
The Local Principal Component Analysis (LPCA) method~\citep{Manjon2013} is also made to take
into account the Rician noise bias and is
spatially adaptive, but also uses the information from all the DWIs in the
denoising process. We used the automatic threshold set by the method with the
Rician noise correction for all experiments.
Both AONLM and LPCA can be
downloaded from the author's
website\footnote{\url{http://personales.upv.es/jmanjon/denoising/index.htm}}.
The multi-shell Position-Orientation Adaptive Smoothing (msPOAS)
algorithm~\citep{Becker2014} was designed for both Rician and nc-$\chi$
noise, while also taking into account the angular structure of the
data for adaptive smoothing.
We discussed with the authors of
msPOAS\footnote{\url{http://cran.r-project.org/web/packages/dti/index.html}}
for their recommendations and using their suggestion, we set $k^\star=12$ and
$\lambda=18$. We also supplied the correct value for $N$
and used the implemented automatic detection of the noise standard deviation
from msPOAS.
For the NLSAM algorithm, we used a patchsize of 3 x 3 x 3 voxels with 4
angular neighbors, which corresponds to the number of angular neighbors at the
same distance on the sphere for each selected DWI. The value of $N$ was
given to the algorithm and the
number of atoms was set to two times the number of voxels in a block
for the dictionary learning part, which was repeated for 150 iterations.
The other parameters were set as described in Section~\ref{sec:local_adapt}.
As shown in Table~\ref{tbl:denoising}, our method is designed to work
on both stationary and spatially
variable Rician and nc-$\chi$ noise.
The NLSAM algorithm is implemented in python and is also freely
available.\footnote{\url{https://github.com/samuelstjean/nlsam}}

Finally, we quantitatively assess the performance of each method by comparing them
against the noiseless synthetic data using

\begin{enumerate}[label=\textbf{\roman*})]

    \item The peak signal-to-noise ratio (PSNR) in dB and the structural
    similarity index (SSIM) on the raw data intensities\review{~\citep{Wang2004}}.

    \item The dispersion of the FA error, computed from a weighted least-square diffusion tensor model. %

    \item The mean angular error (AE) in degrees and the discrete number of
    compartments (DNC) error for a region of crossings~\citep{Daducci2013,Paquette2014a}.

    \item The Tractometer~\citep{Cote2013a} ranking
    platform on deterministic tractography algorithms for the synthetic
    datasets. This platform computes global connectivity metrics, giving an insight on the
    global coherence of the denoised datasets in a tractography setting.

    \item Tracking some known bundles on the high resolution \textit{in-vivo} dataset
    and qualitatively comparing them to their
    lower spatial resolution counterpart.

\end{enumerate}

\subsection{Local models reconstruction and fiber tractography}
The weighted least-square diffusion tensors were \review{reconstructed} using the default
parameters of Dipy~\citep{Garyfallidis2014} \review{to compute the FA values.}
We used the Constrained Spherical Deconvolution (CSD) \citep{Tournier2007}
with a spherical harmonics
of order 8
to reconstruct the fODFs \review{and extract the peaks subsequently
used for the deterministic tracking}. %
To compute the fiber response function (frf), we used all the
voxels in the white matter that had an FA superior to 0.7. If less than 300
voxels meeting this criterion were found, the FA threshold was lowered
by 0.05 until the criterion was met. See Sections~\ref{sec:results_bias} and
\ref{sec:discuss_bias} for more information about the bias introduced in the
FA.
For the synthetic datasets, the tracking was done inside the white matter mask
and the seeding was done from the bundles extremities to mimic seeding from the
white-gray matter interface~\citep{Girard2014}.
We used 100 seeds per voxels to allow sufficient bundle coverage, a stepsize of 0.2 mm
and a maximum angle deviation of 60 degrees.
The other parameters
used were the defaults supplied by the tractometer pipeline~\citep{Cote2013a}.
The \textit{in-vivo} datasets deterministic tracking was made with the technique
of~\citep{Girard2014} by seeding from the white matter and gray matter interface
with the particle filtering and generating approximately 1 million of streamlines.
White matter masks were created by segmenting a T1 image with
FSL FAST\footnote{\url{http://fsl.fmrib.ox.ac.uk/fsl/fslwiki/FAST}}
from the same subject and then registered with
ANTs\footnote{\url{http://picsl.upenn.edu/software/ants/}} to each \textit{in-vivo}
dataset.
The bundles were finally automatically segmented using the
White Matter Query Language (WMQL)~\citep{Wassermann2013a} Tract Querier
tool with regions obtained from a T1 white matter
and gray matter parcellation.
This atlas-based automatic dissection method extracts fiber bundles automatically using
anatomical definitions in a reproducible manner for all methods, as opposed to the
traditional way of manually defining including and excluding ROIs to define bundles.
Visualization of fODFs, peaks and tractography was made using the
fibernavigator\footnote{\url{https://github.com/scilus/fibernavigator}}~\citep{Chamberland2014b}.

\FloatBarrier

\section{Results}
\label{sec:results}
\subsection{Preserving the raw DWI data}

Fig.~\ref{fig:phantomas} (displayed in landscape to see all the compared methods on
the same figure)
shows the b1000 noiseless data, the noisy input data
at SNR 10 for nc-$\chi$ stationary noise and the
results of the denoising on the synthetic phantom for all compared methods.
This is the noise case theoretically covered by msPOAS and our NLSAM algorithm.
We also show two zoomed regions of crossings with the reconstructed peaks extracted from fODFs.
All perceptual and FA metrics were computed on the slice shown while angular
metrics were computed in the zoomed region depicted by the yellow box.
Note how the small blue bundle and its crossings are preserved on the NLSAM
denoised dataset, while other denoising methods tend to introduce blurring.

Fig.~\ref{fig:1_2mm} (also in landscape for enhanced viewing)
shows the noisy high resolution \textit{in-vivo} dataset,
the denoised version obtained for each algorithm and the low spatial resolution
acquisition of the same subject without any denoising.
Since our scanner uses a 32 channels head-coil but implements the SENSE reconstruction algorithm,
the resulting spatially varying Rician noise distribution is the case covered by
AONLM, LPCA and our NLSAM algorithm.
We show a coronal slice for the gradient direction closest to (0, 1, 0),
the colored FA map and a zoom on two regions of crossings.
The yellow region shows the junction of the corticospinal tract (CST) and
superior longitudinal fasciculus (SLF) while the white region shows
the junction of the corpus callosum (CC) and the CST.
While the high resolution dataset is noisier than its lower resolution counterpart,
the highlighted crossings regions are well recovered by the denoising algorithms
and thus offer an improvement in anatomical details over the lower spatial
resolution dataset. We also see in the yellow box  that the NLSAM denoised
dataset recovers crossings extending from the CC which are almost absent in the
compared datasets.

\begin{landscape}
\thispagestyle{empty}
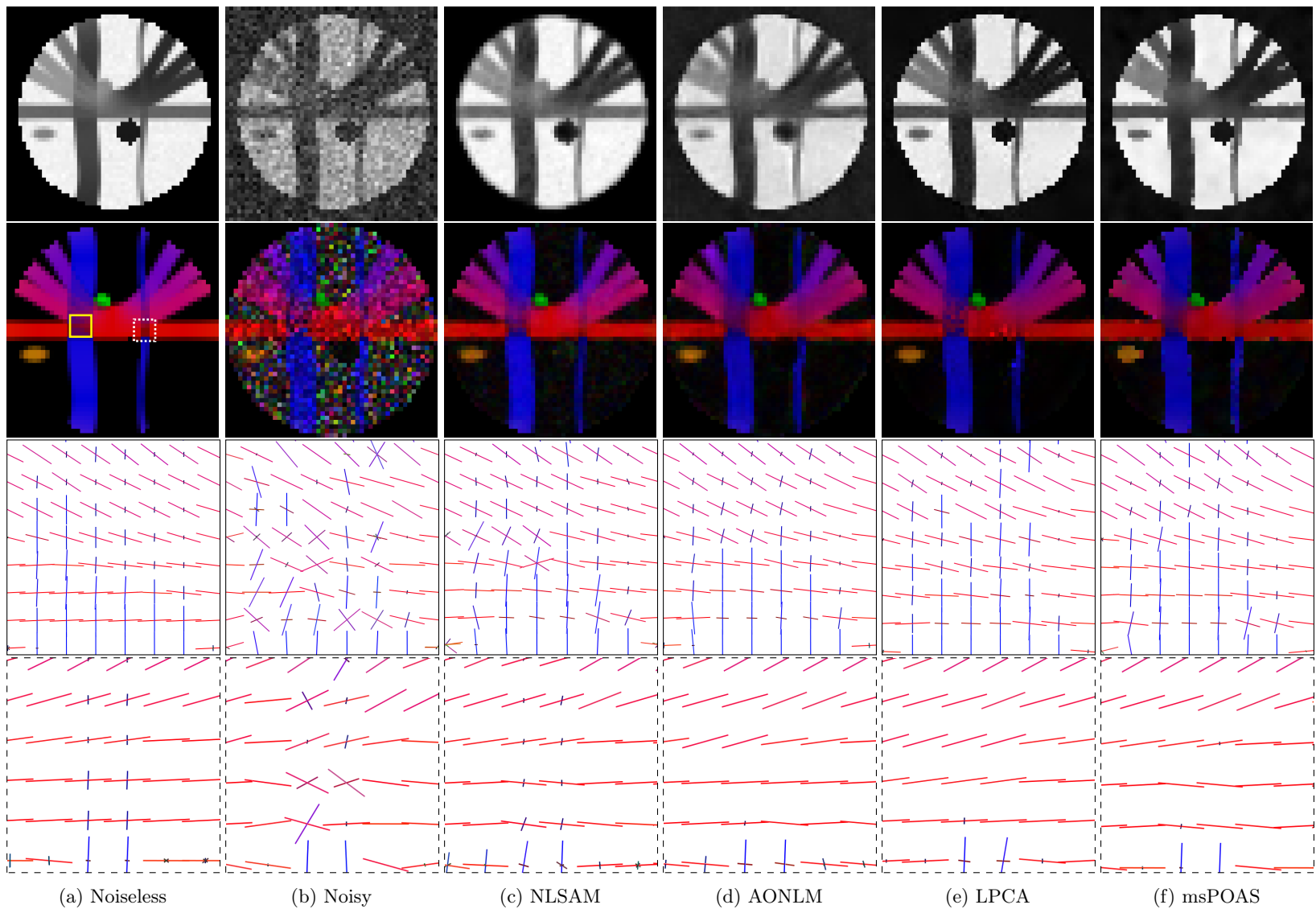
\begin{figure}[htb]
\small
\centering
            \begin{subfigure}[b]{0.16\linewidth}
                    \includegraphics[width=\textwidth]{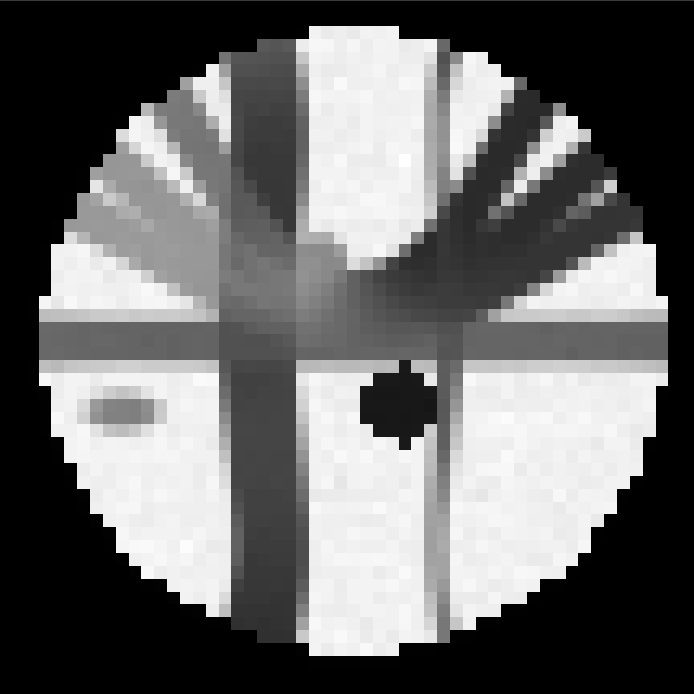}
            \end{subfigure}
            \begin{subfigure}[b]{0.16\linewidth}
                    \includegraphics[width=\textwidth]{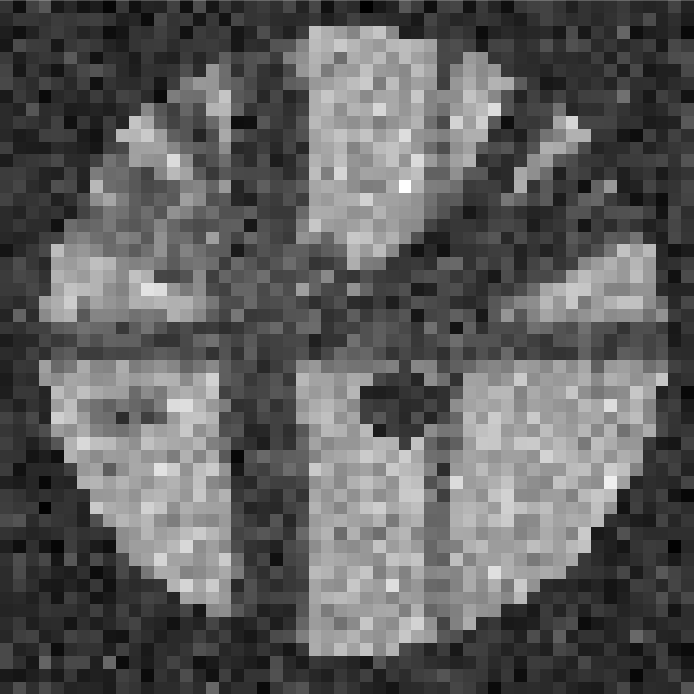}
            \end{subfigure}
            \begin{subfigure}[b]{0.16\linewidth}
                    \includegraphics[width=\textwidth]{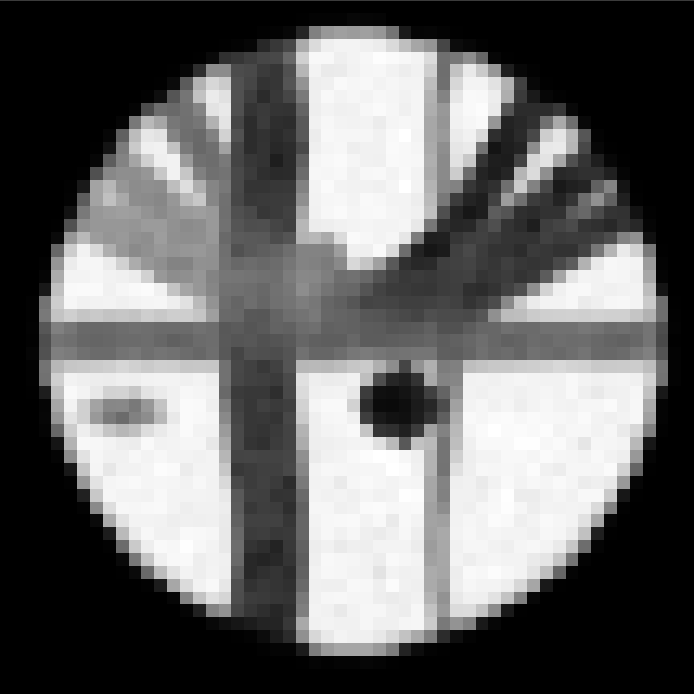}
            \end{subfigure}
            \begin{subfigure}[b]{0.16\linewidth}
                    \includegraphics[width=\textwidth]{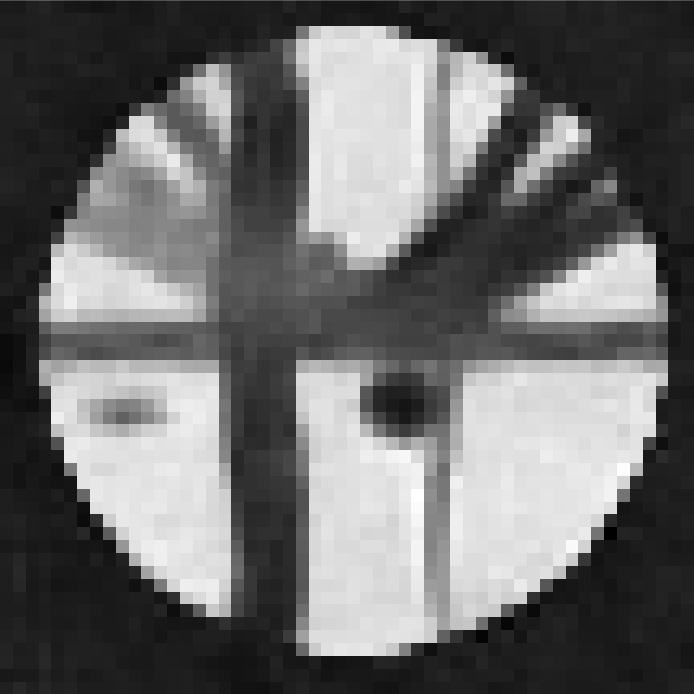}
            \end{subfigure}
            \begin{subfigure}[b]{0.16\linewidth}
                    \includegraphics[width=\textwidth]{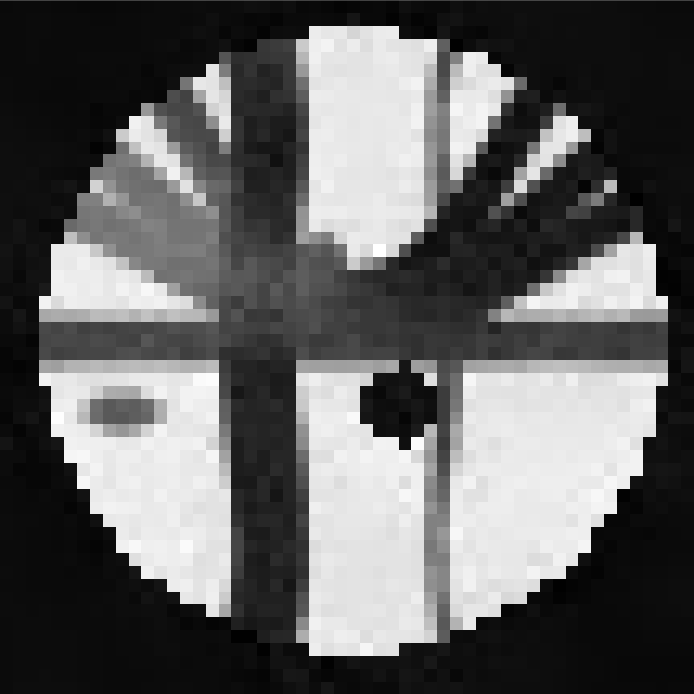}
            \end{subfigure}
            \begin{subfigure}[b]{0.16\linewidth}
                    \includegraphics[width=\textwidth]{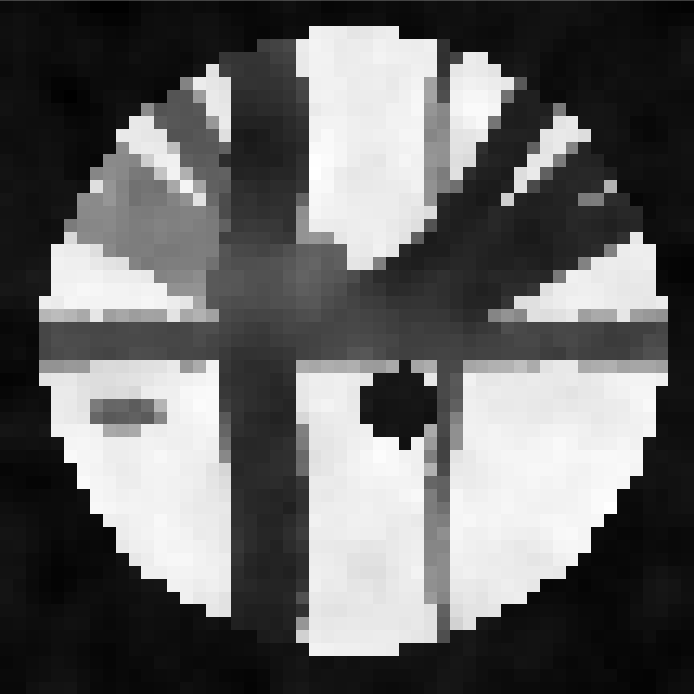}
            \end{subfigure}

            \begin{subfigure}[b]{0.16\linewidth}
                    \begin{tikzonimage}[width=\textwidth]{b1000/rgb/gt.png}%
                    \draw [yellow,line width=1pt] (0.4,0.47) rectangle (0.3,0.57) ;
                    \draw [white,densely dotted,line width=1pt] (0.7,0.55) rectangle (0.6,0.45) ;
                    \end{tikzonimage}
            \end{subfigure}
            \begin{subfigure}[b]{0.16\linewidth}
                    \includegraphics[width=\textwidth]{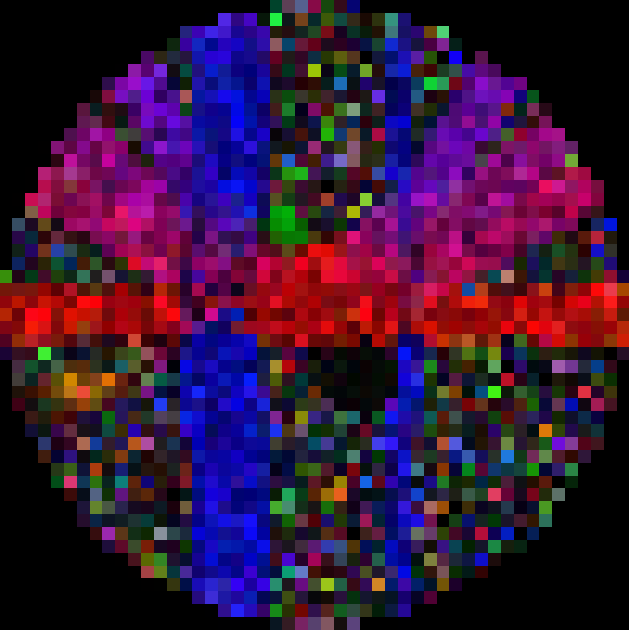}
            \end{subfigure}
            \begin{subfigure}[b]{0.16\linewidth}
                    \includegraphics[width=\textwidth]{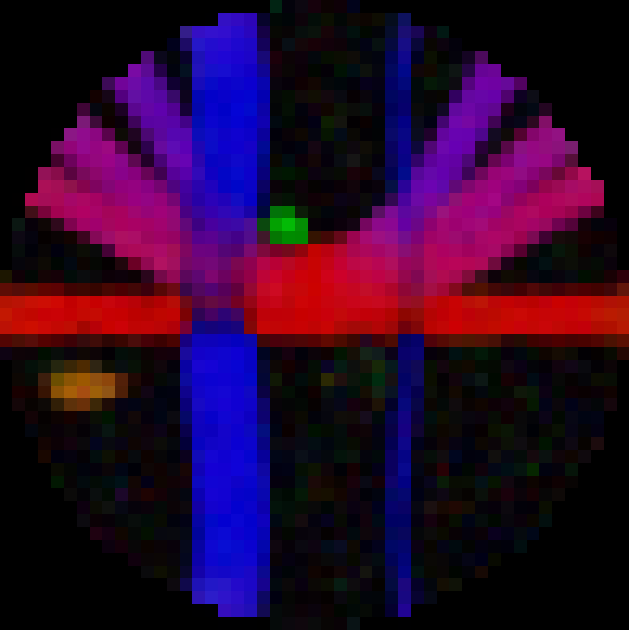}
            \end{subfigure}
            \begin{subfigure}[b]{0.16\linewidth}
                    \includegraphics[width=\textwidth]{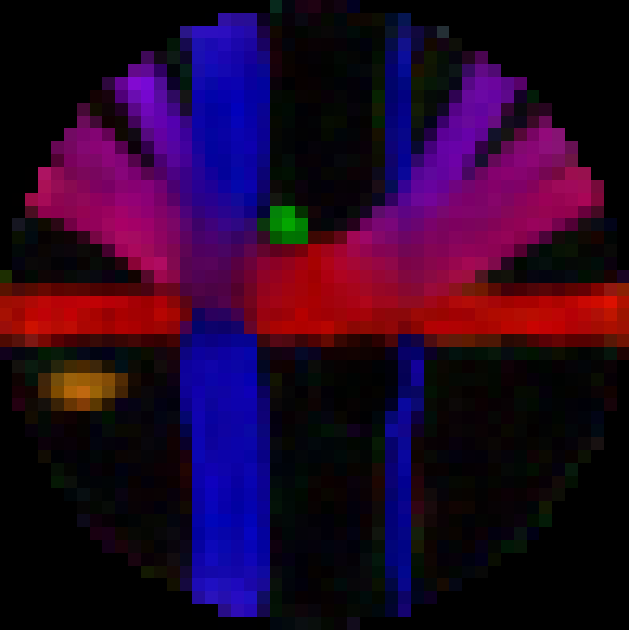}
            \end{subfigure}
            \begin{subfigure}[b]{0.16\linewidth}
                    \includegraphics[width=\textwidth]{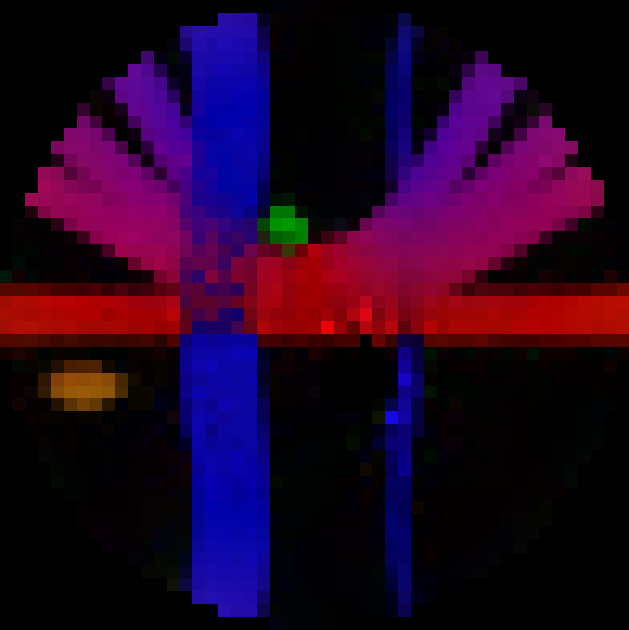}
            \end{subfigure}
            \begin{subfigure}[b]{0.16\linewidth}
                    \includegraphics[width=\textwidth]{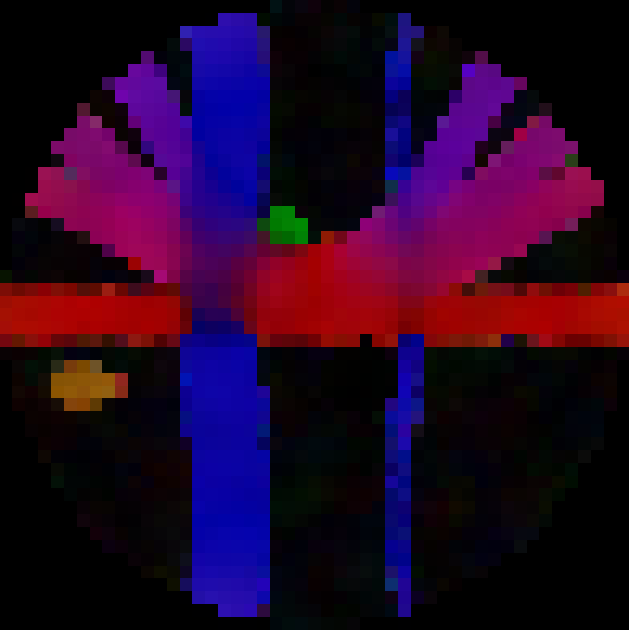}
            \end{subfigure}

            \begin{subfigure}[b]{0.16\linewidth}
                    \begin{tikzonimage}[width=\textwidth,height=3.5cm]{b1000/peaks/gt_left.png}
                    \draw [black] (0,0) rectangle (1,1) ;
                    \end{tikzonimage}
            \end{subfigure}
            \begin{subfigure}[b]{0.16\linewidth}
                    \begin{tikzonimage}[width=\textwidth,height=3.5cm]{b1000/peaks/noisy_left.png}
                    \draw [black] (0,0) rectangle (1,1) ;
                    \end{tikzonimage}
            \end{subfigure}
            \begin{subfigure}[b]{0.16\linewidth}
                    \begin{tikzonimage}[width=\textwidth,height=3.5cm]{b1000/peaks/nlsam_left.png}
                    \draw [black] (0,0) rectangle (1,1) ;
                    \end{tikzonimage}
            \end{subfigure}
            \begin{subfigure}[b]{0.16\linewidth}
                    \begin{tikzonimage}[width=\textwidth,height=3.5cm]{b1000/peaks/aonlm_left.png}
                    \draw [black] (0,0) rectangle (1,1) ;
                    \end{tikzonimage}
            \end{subfigure}
            \begin{subfigure}[b]{0.16\linewidth}
                    \begin{tikzonimage}[width=\textwidth,height=3.5cm]{b1000/peaks/lpca_left.png}
                    \draw [black] (0,0) rectangle (1,1) ;
                    \end{tikzonimage}
            \end{subfigure}
            \begin{subfigure}[b]{0.16\linewidth}
                    \begin{tikzonimage}[width=\textwidth,height=3.5cm]{b1000/peaks/mspoas_left.png}
                    \draw [black] (0,0) rectangle (1,1) ;
                    \end{tikzonimage}
            \end{subfigure}

            \begin{subfigure}[b]{0.16\linewidth}
                    \begin{tikzonimage}[width=\textwidth,height=3.5cm]{b1000/peaks/gt_right.png}
                    \draw [black,dashed] (0,0) rectangle (1,1) ;
                    \end{tikzonimage}
                    \caption{Noiseless}
            \end{subfigure}
            \begin{subfigure}[b]{0.16\linewidth}
                    \begin{tikzonimage}[width=\textwidth,height=3.5cm]{b1000/peaks/noisy_right.png}
                    \draw [black,dashed] (0.0,0.0) rectangle (1,1) ;
                    \end{tikzonimage}
                    \caption{Noisy}
            \end{subfigure}
            \begin{subfigure}[b]{0.16\linewidth}
                    \begin{tikzonimage}[width=\textwidth,height=3.5cm]{b1000/peaks/nlsam_right.png}
                    \draw [black,dashed] (0.0,0.0) rectangle (1,1) ;
                    \end{tikzonimage}
                    \caption{NLSAM}
            \end{subfigure}
            \begin{subfigure}[b]{0.16\linewidth}
                    \begin{tikzonimage}[width=\textwidth,height=3.5cm]{b1000/peaks/aonlm_right.png}
                    \draw [black,dashed] (0.0,0.0) rectangle (1,1) ;
                    \end{tikzonimage}
                    \caption{AONLM}
            \end{subfigure}
            \begin{subfigure}[b]{0.16\linewidth}
                    \begin{tikzonimage}[width=\textwidth,height=3.5cm]{b1000/peaks/lpca_right.png}
                    \draw [black,dashed] (0.0,0.0) rectangle (1,1) ;
                    \end{tikzonimage}
                    \caption{LPCA}
            \end{subfigure}
            \begin{subfigure}[b]{0.16\linewidth}
                    \begin{tikzonimage}[width=\textwidth,height=3.5cm]{b1000/peaks/mspoas_right.png}
                    \draw [black,dashed] (0.0,0.0) rectangle (1,1) ;
                    \end{tikzonimage}
                    \caption{msPOAS}
            \end{subfigure}

    \caption{Phantomas b1000 synthetic dataset at SNR 10 for stationary nc-$\chi$ noise on the y = 24 slice.
    From top to bottom~: Raw diffusion MRI, colored FA map, zoom on extracted peaks from fODF of order 8.
    Note how NLSAM restores the structure without blurring on the colored FA map
    and is the only method to restore the peaks from the noisy dataset in the zoomed white box region.}

\label{fig:phantomas}
\floatfoot{\thepage}
\end{figure}
\end{landscape}

\begin{landscape}
\thispagestyle{empty}
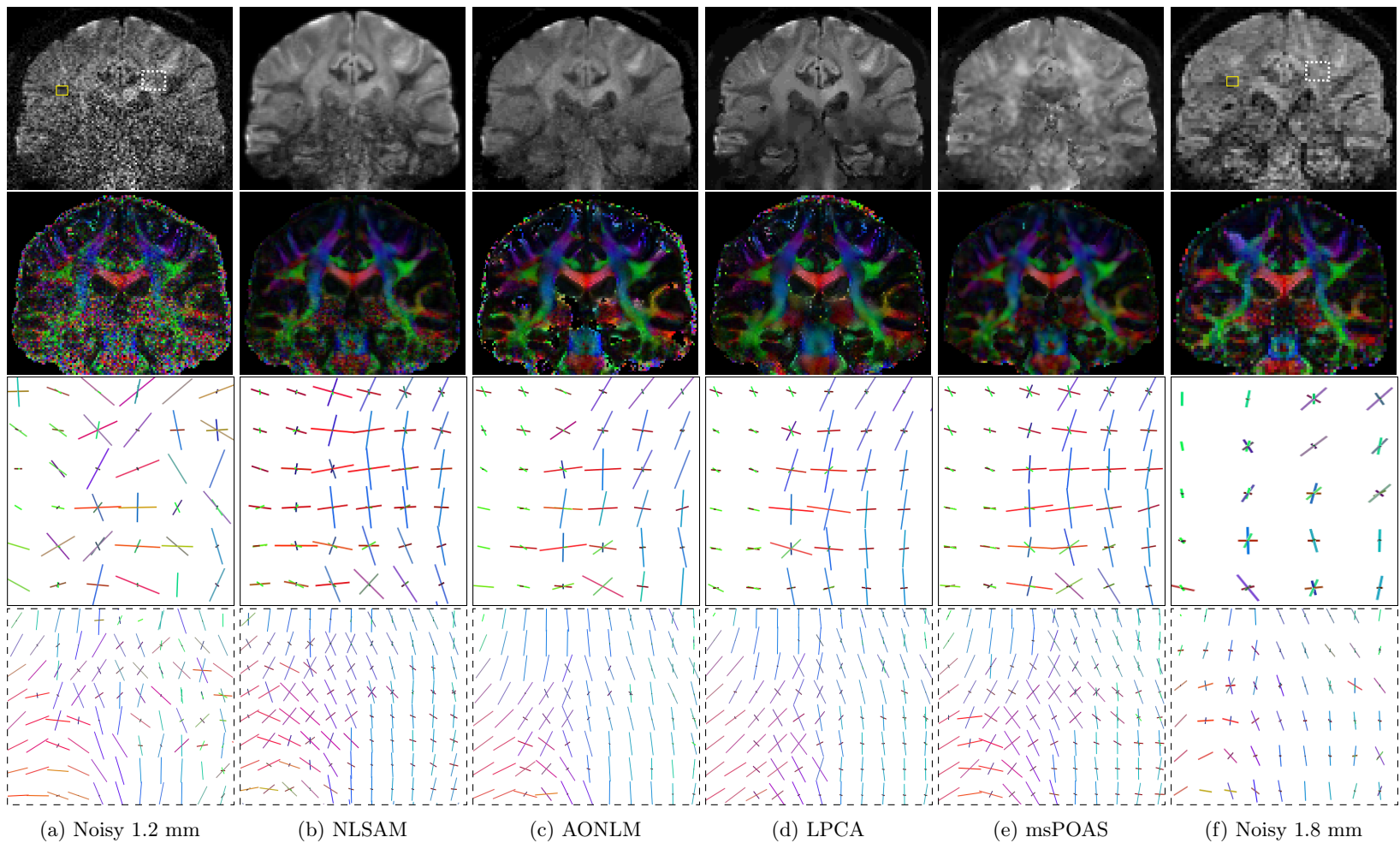
\begin{figure}[htb]
\small
\centering%
            \begin{subfigure}[b]{0.16\linewidth}
                    \begin{tikzonimage}[width=\textwidth,keepaspectratio=false,height=2.8cm]{1_2mm/raw/noisy_out.png}%
                    \draw [white,densely dotted,line width=1pt] (0.7,0.65) rectangle (0.6,0.55) ;
                    \draw [yellow] (0.27,0.57) rectangle (0.22,0.52) ;
                    \end{tikzonimage}
            \end{subfigure}
            \begin{subfigure}[b]{0.16\linewidth}
                    \includegraphics[width=\textwidth,keepaspectratio=false,height=2.8cm]{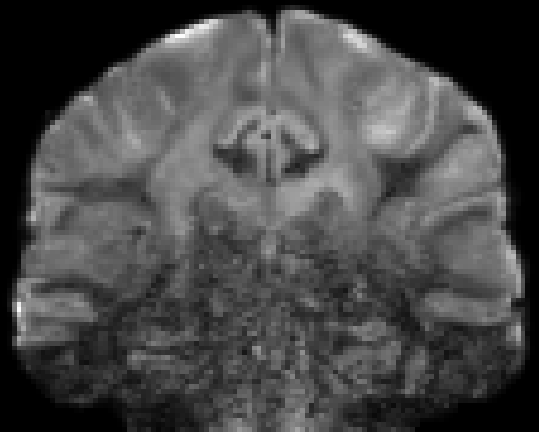}
            \end{subfigure}
            \begin{subfigure}[b]{0.16\linewidth}
                    \includegraphics[width=\textwidth,keepaspectratio=false,height=2.8cm]{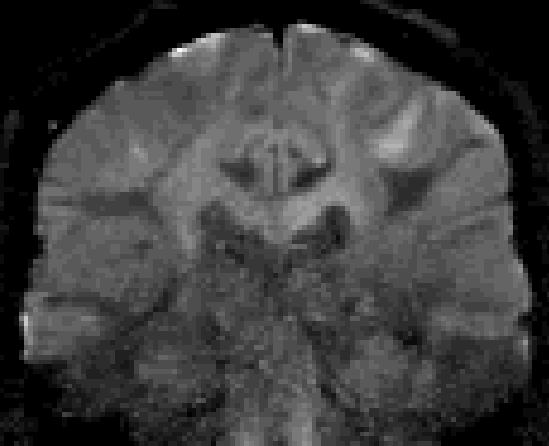}
            \end{subfigure}
            \begin{subfigure}[b]{0.16\linewidth}
                    \includegraphics[width=\textwidth,keepaspectratio=false,height=2.8cm]{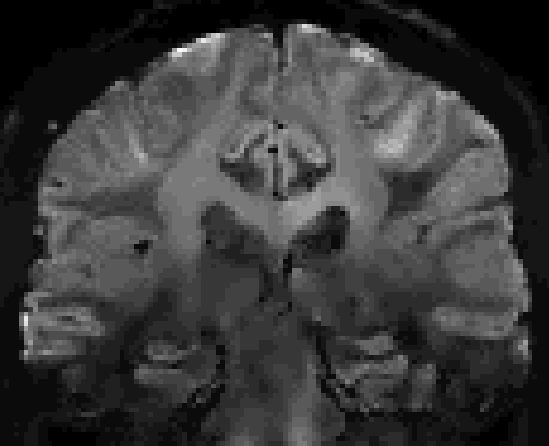}
            \end{subfigure}
            \begin{subfigure}[b]{0.16\linewidth}
                    \includegraphics[width=\textwidth,keepaspectratio=false,height=2.8cm]{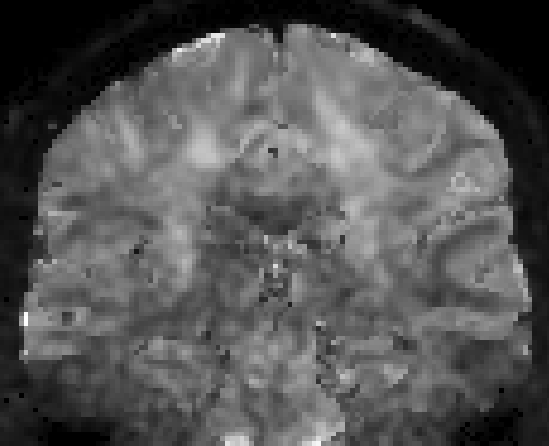}
            \end{subfigure}
            \begin{subfigure}[b]{0.16\linewidth}
                    \begin{tikzonimage}[width=\textwidth,keepaspectratio=false,height=2.8cm]{1_2mm/raw/noisy_1_8.png}%
                    \draw [white,densely dotted,line width=1pt] (0.7,0.7) rectangle (0.6,0.6) ;
                    \draw [yellow] (0.3,0.62) rectangle (0.25,0.57) ;
                    \end{tikzonimage}
            \end{subfigure}
            \begin{subfigure}[b]{0.16\linewidth}
                    \includegraphics[width=\textwidth,keepaspectratio=false,height=2.8cm]{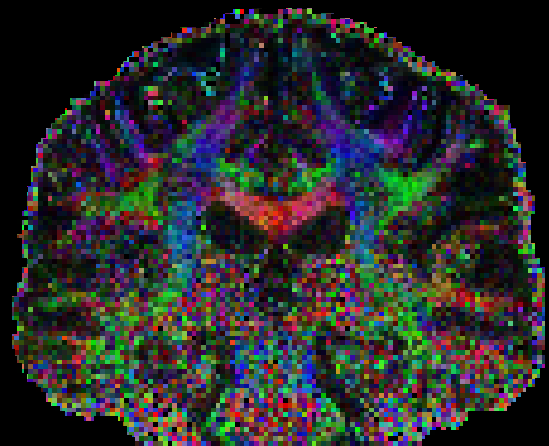}
            \end{subfigure}
            \begin{subfigure}[b]{0.16\linewidth}
                    \includegraphics[width=\textwidth,keepaspectratio=false,height=2.8cm]{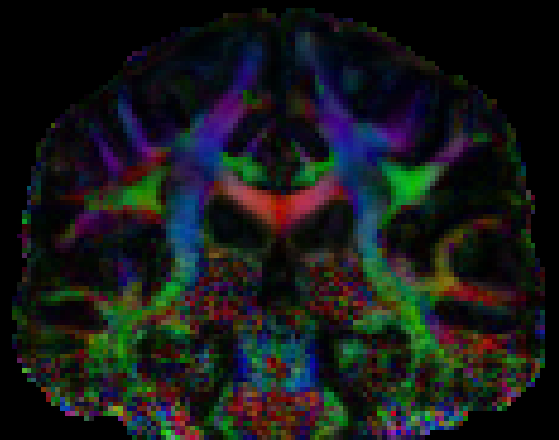}
            \end{subfigure}
            \begin{subfigure}[b]{0.16\linewidth}
                    \includegraphics[width=\textwidth,keepaspectratio=false,height=2.8cm]{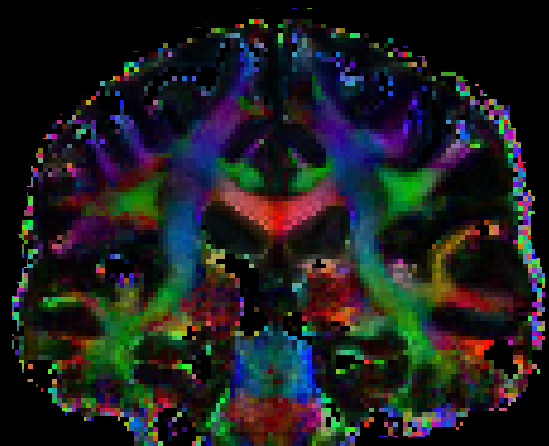}
            \end{subfigure}
            \begin{subfigure}[b]{0.16\linewidth}
                    \includegraphics[width=\textwidth,keepaspectratio=false,height=2.8cm]{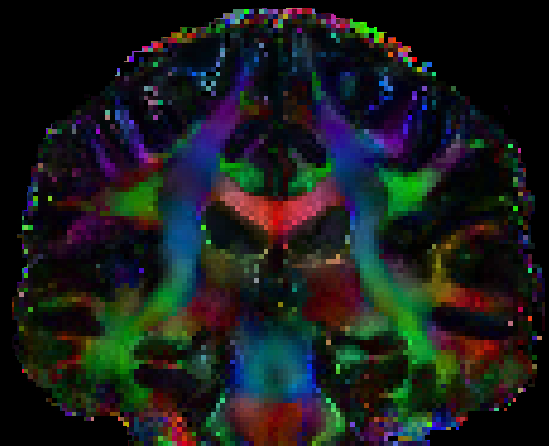}
            \end{subfigure}
            \begin{subfigure}[b]{0.16\linewidth}
                    \includegraphics[width=\textwidth,keepaspectratio=false,height=2.8cm]{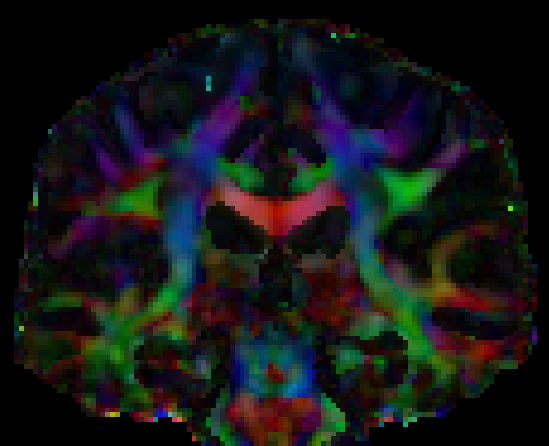}
            \end{subfigure}
            \begin{subfigure}[b]{0.16\linewidth}
                    \includegraphics[width=\textwidth,keepaspectratio=false,height=2.8cm]{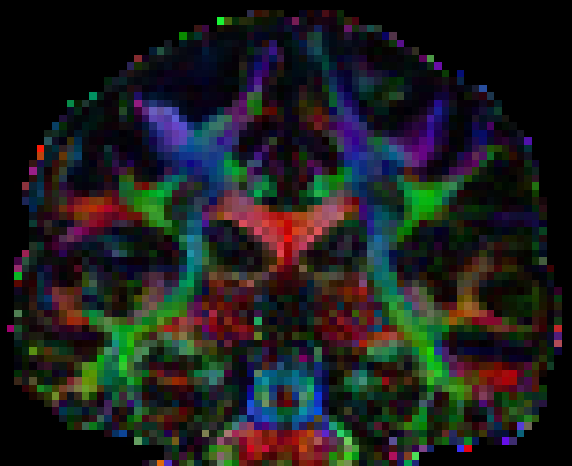}
            \end{subfigure}

            \begin{subfigure}[b]{0.16\linewidth}
                    \begin{tikzonimage}[width=\textwidth,keepaspectratio=false,height=3.5cm]{1_2mm/peaks/noisy_y.png}
                    \draw [black] (0,0) rectangle (1,1) ;
                    \end{tikzonimage}
            \end{subfigure}
            \begin{subfigure}[b]{0.16\linewidth}
                    \begin{tikzonimage}[width=\textwidth,keepaspectratio=false,height=3.5cm]{1_2mm/peaks/nlsam_y.png}
                    \draw [black] (0,0) rectangle (1,1) ;
                    \end{tikzonimage}
            \end{subfigure}
            \begin{subfigure}[b]{0.16\linewidth}
                    \begin{tikzonimage}[width=\textwidth,keepaspectratio=false,height=3.5cm]{1_2mm/peaks/aonlm_y.png}
                    \draw [black] (0,0) rectangle (1,1) ;
                    \end{tikzonimage}
            \end{subfigure}
            \begin{subfigure}[b]{0.16\linewidth}
                    \begin{tikzonimage}[width=\textwidth,keepaspectratio=false,height=3.5cm]{1_2mm/peaks/lpca_y.png}
                    \draw [black] (0,0) rectangle (1,1) ;
                    \end{tikzonimage}
            \end{subfigure}
            \begin{subfigure}[b]{0.16\linewidth}
                    \begin{tikzonimage}[width=\textwidth,keepaspectratio=false,height=3.5cm]{1_2mm/peaks/mspoas_y.png}
                    \draw [black] (0,0) rectangle (1,1) ;
                    \end{tikzonimage}
            \end{subfigure}
            \begin{subfigure}[b]{0.16\linewidth}
                    \begin{tikzonimage}[width=\textwidth,keepaspectratio=false,height=3.5cm]{1_2mm/peaks/noisy_1_8_y.png}
                    \draw [black] (0,0) rectangle (1,1) ;
                    \end{tikzonimage}
            \end{subfigure}

            \begin{subfigure}[b]{0.16\linewidth}
                    \begin{tikzonimage}[width=\textwidth,keepaspectratio=false,height=3cm]{1_2mm/peaks/noisy3_g.png}
                    \draw [black,dashed] (0,0) rectangle (1,1) ;
                    \end{tikzonimage}
                    \caption{Noisy 1.2 mm}
            \end{subfigure}
            \begin{subfigure}[b]{0.16\linewidth}
                    \begin{tikzonimage}[width=\textwidth,keepaspectratio=false,height=3cm]{1_2mm/peaks/nlsam3_g.png}
                    \draw [black,dashed] (0,0) rectangle (1,1) ;
                    \end{tikzonimage}
                    \caption{NLSAM}
            \end{subfigure}
            \begin{subfigure}[b]{0.16\linewidth}
                    \begin{tikzonimage}[width=\textwidth,keepaspectratio=false,height=3cm]{1_2mm/peaks/aonlm3_g.png}
                    \draw [black,dashed] (0,0) rectangle (1,1) ;
                    \end{tikzonimage}
                    \caption{AONLM}
            \end{subfigure}
            \begin{subfigure}[b]{0.16\linewidth}
                    \begin{tikzonimage}[width=\textwidth,keepaspectratio=false,height=3cm]{1_2mm/peaks/lpca3_g.png}
                    \draw [black,dashed] (0,0) rectangle (1,1) ;
                    \end{tikzonimage}
                    \caption{LPCA}
            \end{subfigure}
            \begin{subfigure}[b]{0.16\linewidth}
                    \begin{tikzonimage}[width=\textwidth,keepaspectratio=false,height=3cm]{1_2mm/peaks/mspoas3_g.png}
                    \draw [black,dashed] (0,0) rectangle (1,1) ;
                    \end{tikzonimage}
                    \caption{msPOAS}
            \end{subfigure}
            \begin{subfigure}[b]{0.16\linewidth}
                    \begin{tikzonimage}[width=\textwidth,keepaspectratio=false,height=3cm]{1_2mm/peaks/noisy_1_8v3_g.png}
                    \draw [black,dashed] (0,0) rectangle (1,1) ;
                    \end{tikzonimage}
                    \caption{Noisy 1.8 mm}
            \end{subfigure}

    \caption{From top to bottom, the raw high resolution \textit{in-vivo}
    data corrupted with spatially varying Rician \review{distributed intensities},
    the colored FA map and a zoom on two regions of crossings.
    All denoised methods were applied on the high spatial resolution 1.2 mm dataset.
    We also show an acquisition of the same subject at 1.8 mm for visual comparison.
    Our NLSAM algorithm is able to recover more crossings from
    the 3 way junction of the SLF, the CST and the CC as shown in the yellow and white boxes.
    While the 1.8 mm dataset is less noisy, its lower spatial resolution also means that each voxel contains
    more heterogeneous tissues and mixed diffusion orientations.
    The 1.2 mm denoised dataset shows more crossings without the averaging effect
    of the larger voxel size. For a comparable acquisition time, the denoised high
    resolution dataset has more information than its lower resolution counterpart without
    processing.}

\label{fig:1_2mm}
\floatfoot{\thepage}
\end{figure}
\end{landscape}

\begin{figure}[htb]
\small
\centering
            \begin{subfigure}[b]{0.49\textwidth}
                    \includegraphics[width=\textwidth]{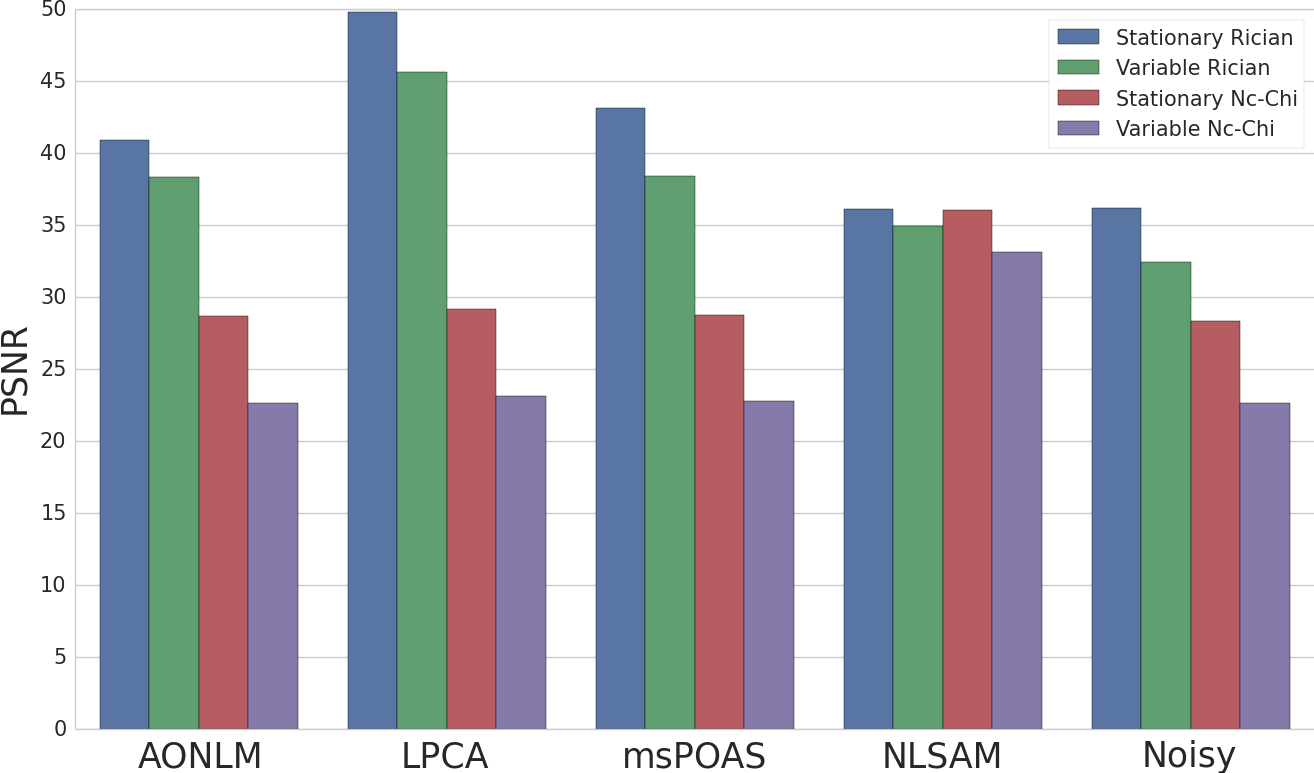}
                    \caption{PSNR b1000}
            \end{subfigure}
            \begin{subfigure}[b]{0.49\textwidth}
                    \includegraphics[width=\textwidth]{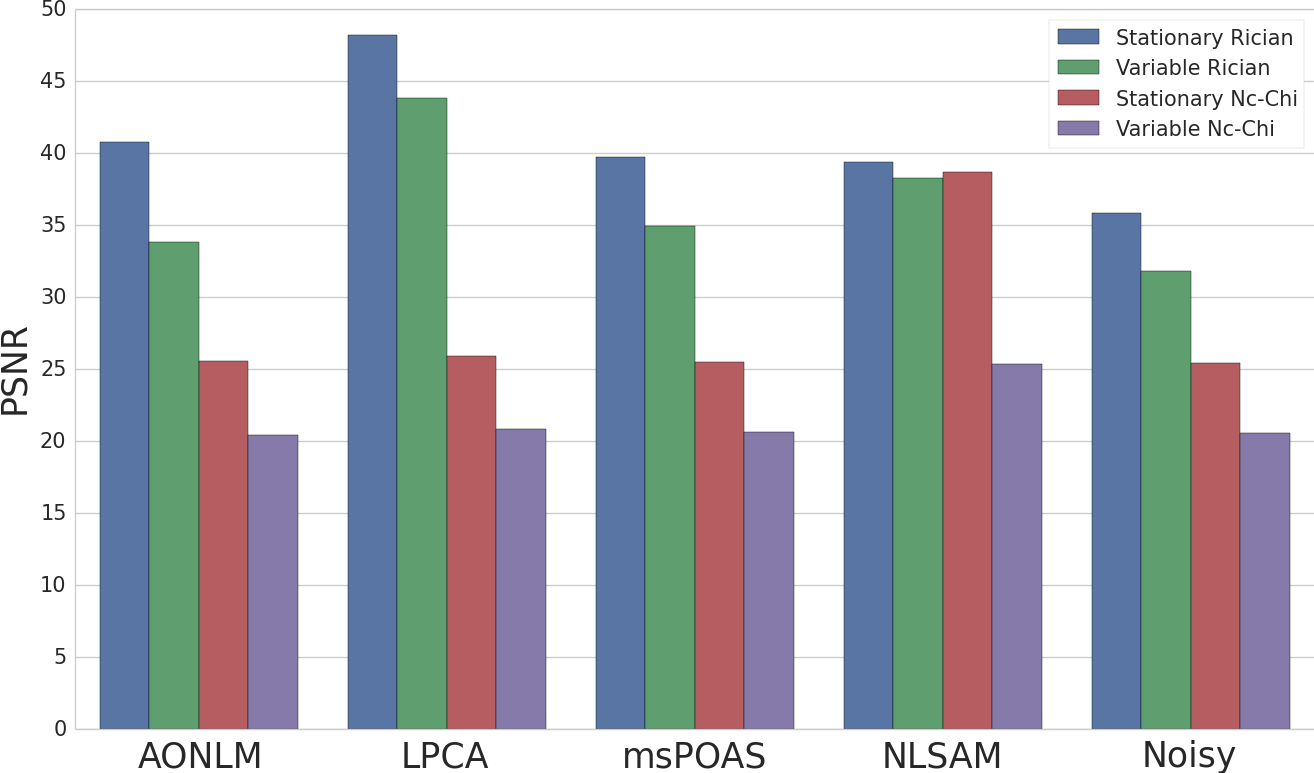}
                    \caption{PSNR b3000}
            \end{subfigure}

            \begin{subfigure}[b]{0.49\textwidth}
                    \includegraphics[width=\textwidth]{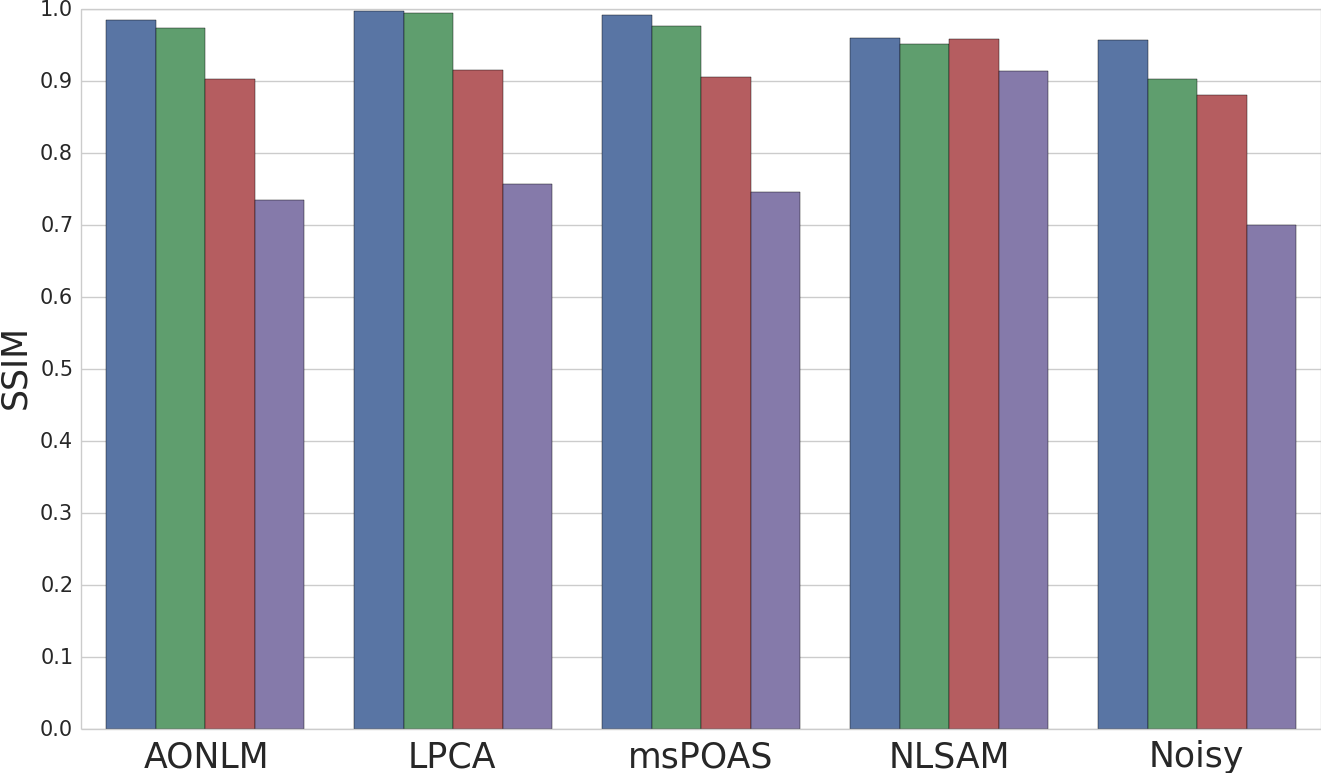}
                    \caption{SSIM b1000}
            \end{subfigure}
            \begin{subfigure}[b]{0.49\textwidth}
                    \includegraphics[width=\textwidth]{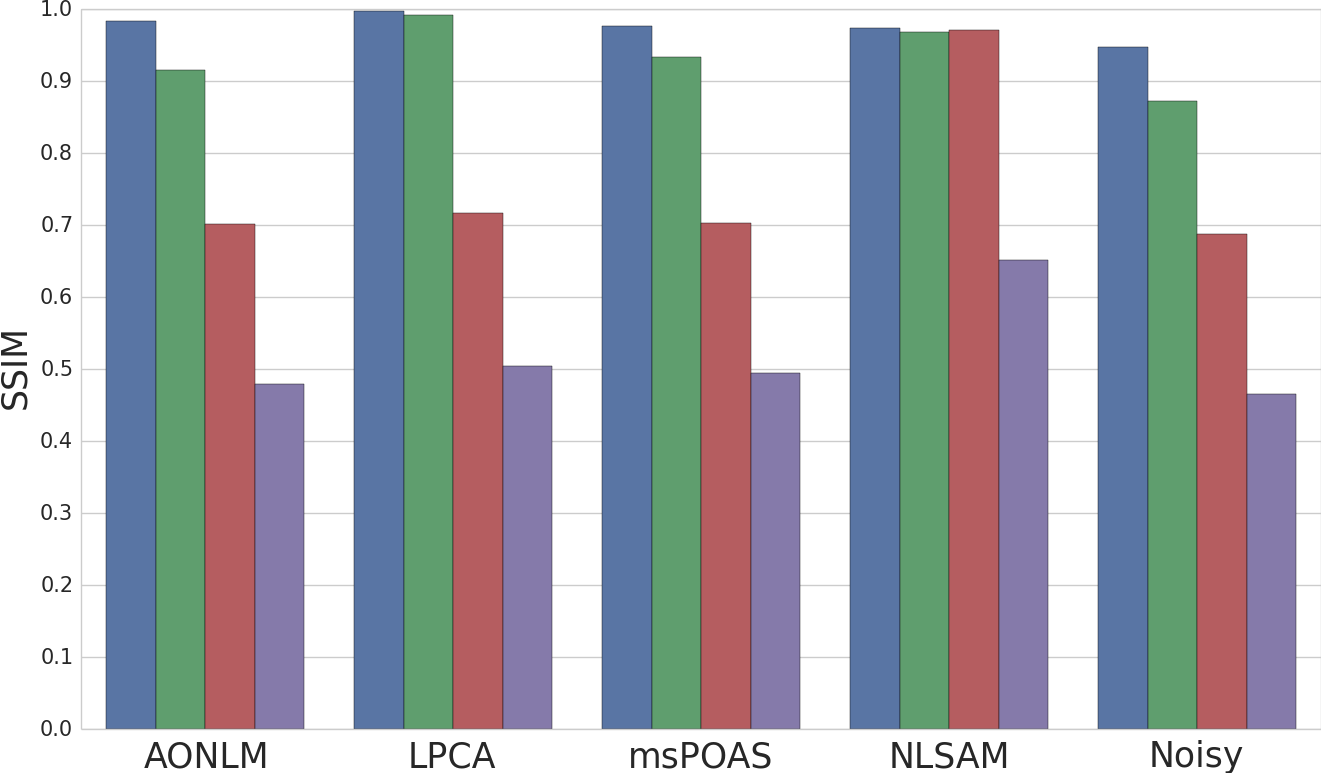}
                    \caption{SSIM b3000}
            \end{subfigure}

        \caption{PSNR and SSIM metrics for the SNR 10 stationary
        and SNR 15 spatially variable noise cases datasets.
        All methods can correct
        the stationary and spatially varying Rician noise case to some extent while only
        our NLSAM algorithm has the best performance for the
        nc-$\chi$ noise case, especially for the spatially varying noise case.}

    \label{fig:graph_percep}

\end{figure}

Fig.~\ref{fig:graph_percep} shows the PSNR and SSIM for the SNR 10 (stationary noise) and
SNR 15 (spatially varying noise) synthetic datasets.
The LPCA algorithm performs best
in term of PSNR on the Rician noise
case, but attains a lower score for nc-$\chi$ noisy datasets.
The same trend is seen for AONLM and msPOAS algorithms, where the SNR 15 nc-$\chi$
case is the hardest test case.
In contrast, our NLSAM technique is above 30 dB
for the PSNR and 0.9 for the SSIM in most cases, with a relatively stable performance
amongst the majority of tested cases.
We also note that even though msPOAS is made to adjust itself to nc-$\chi$ noise,
the fact that the algorithm does not account for the intensity bias makes
the perceptual metrics drop for the nc-$\chi$ noise cases.

% \FloatBarrier
\clearpage
\subsection{Bias introduced in the FA}
\label{sec:results_bias}

\begin{figure}[htpb]
\small
\centering

\begin{subfigure}[b]{\textwidth}
    \begin{subfigure}[b]{0.97\textwidth}
          \rotatebox{90}{\kern5pt Stationary nc-$\chi$}
            \begin{subfigure}[b]{0.19\textwidth}
                    \includegraphics[width=\textwidth]{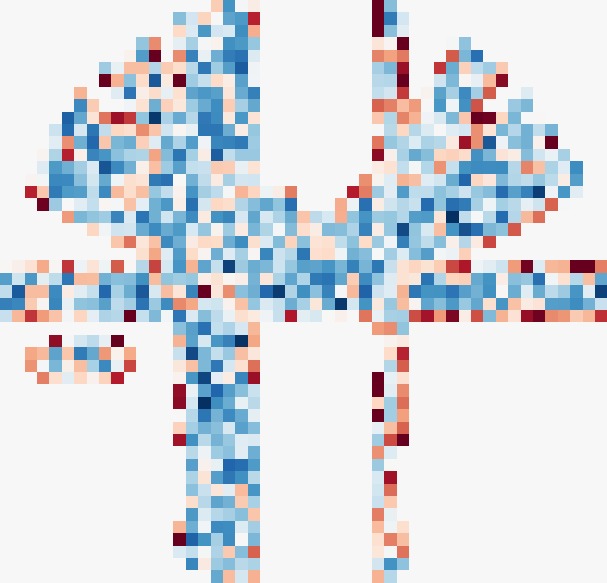}
            \end{subfigure}
            \begin{subfigure}[b]{0.19\textwidth}
                    \includegraphics[width=\textwidth]{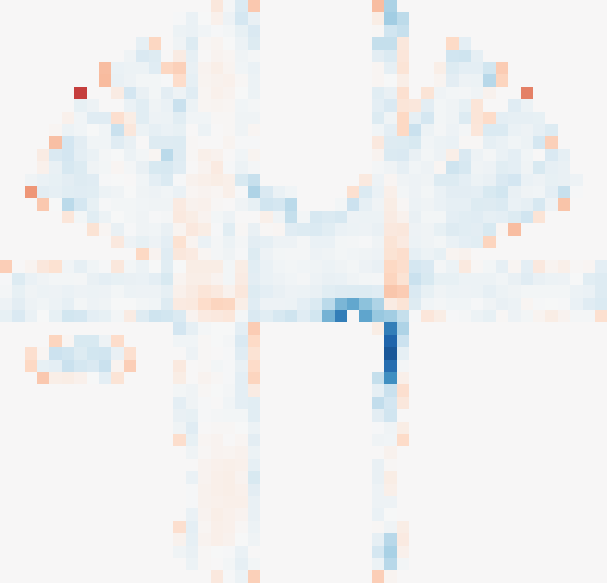}
            \end{subfigure}
            \begin{subfigure}[b]{0.19\textwidth}
                    \includegraphics[width=\textwidth]{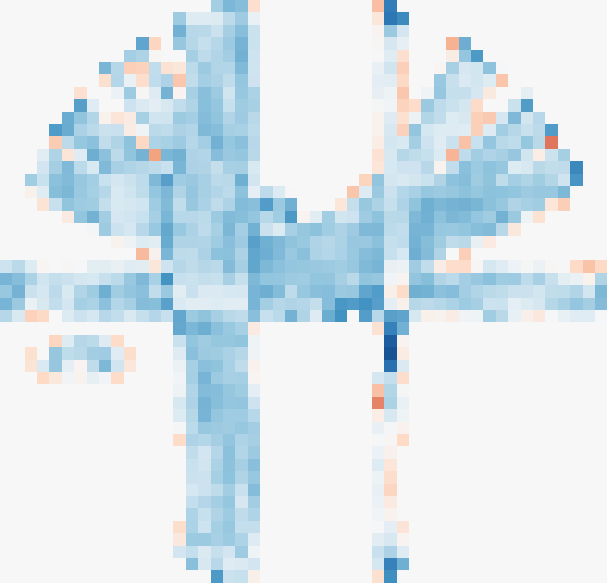}
            \end{subfigure}
            \begin{subfigure}[b]{0.19\textwidth}
                    \includegraphics[width=\textwidth]{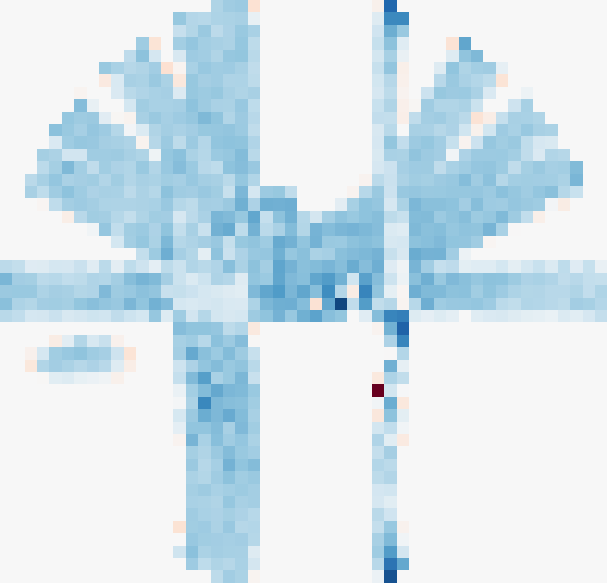}
            \end{subfigure}
            \begin{subfigure}[b]{0.19\textwidth}
                    \includegraphics[width=\textwidth]{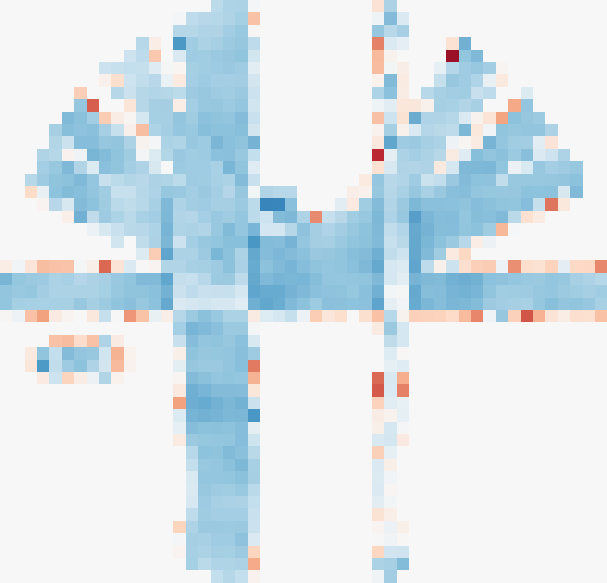}
            \end{subfigure}

            \rotatebox{90}{\kern15pt Spatially var. Rician}
            \begin{subfigure}[b]{0.19\textwidth}
                    \includegraphics[width=\textwidth]{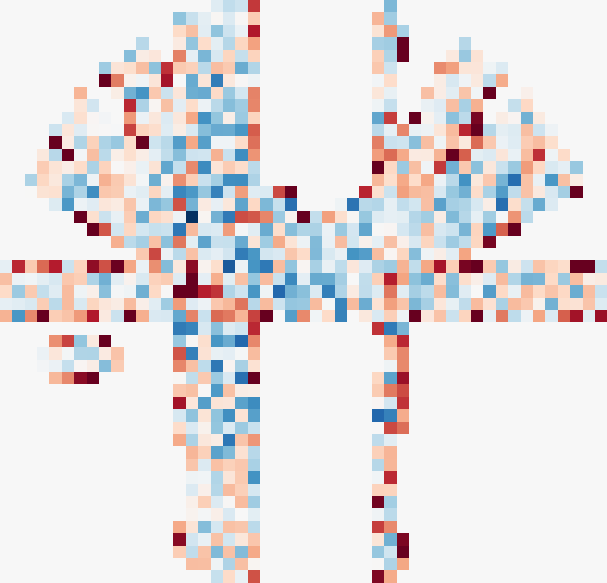}
                    \caption{Noisy}
            \end{subfigure}
            \begin{subfigure}[b]{0.19\textwidth}
                    \includegraphics[width=\textwidth]{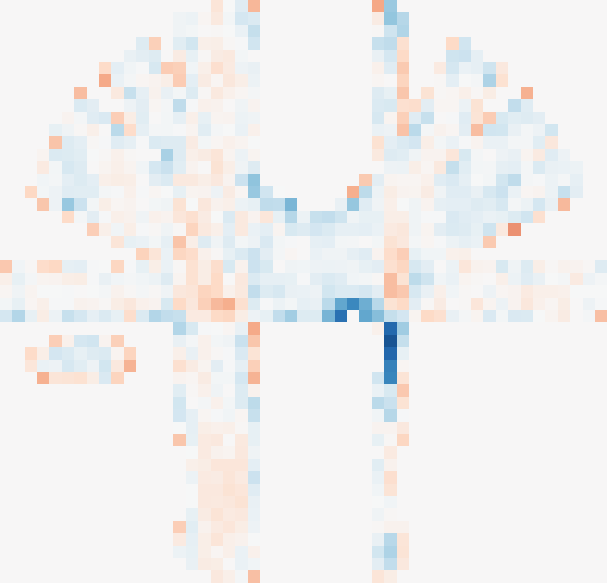}
                    \caption{NLSAM}
            \end{subfigure}
            \begin{subfigure}[b]{0.19\textwidth}
                    \includegraphics[width=\textwidth]{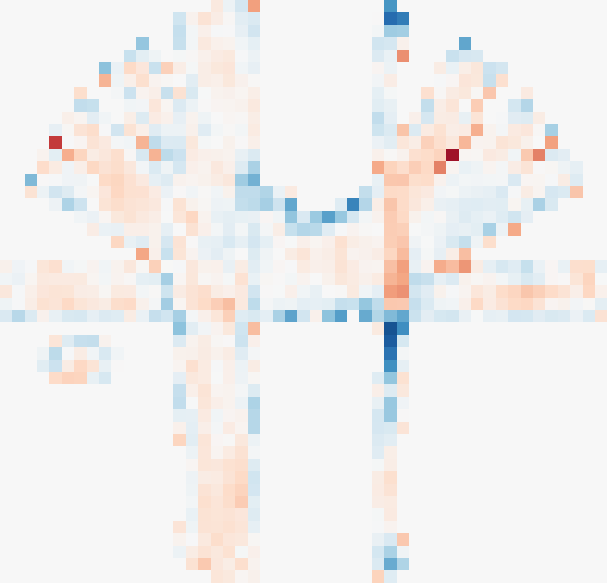}
                    \caption{AONLM}
            \end{subfigure}
            \begin{subfigure}[b]{0.19\textwidth}
                    \includegraphics[width=\textwidth]{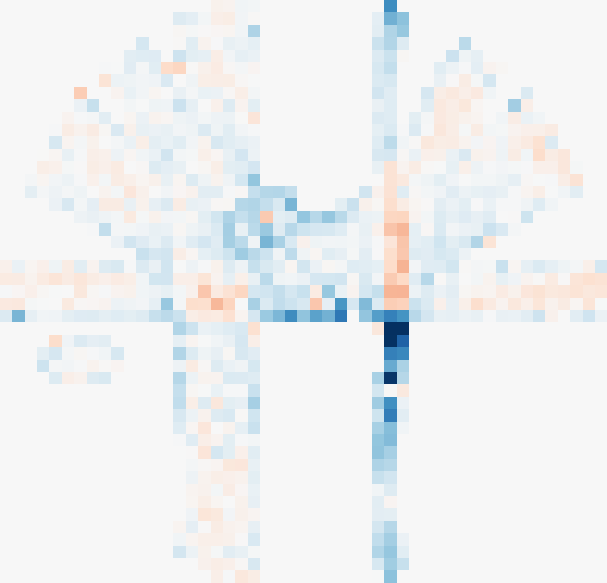}
                    \caption{LPCA}
            \end{subfigure}
            \begin{subfigure}[b]{0.19\textwidth}
                    \includegraphics[width=\textwidth]{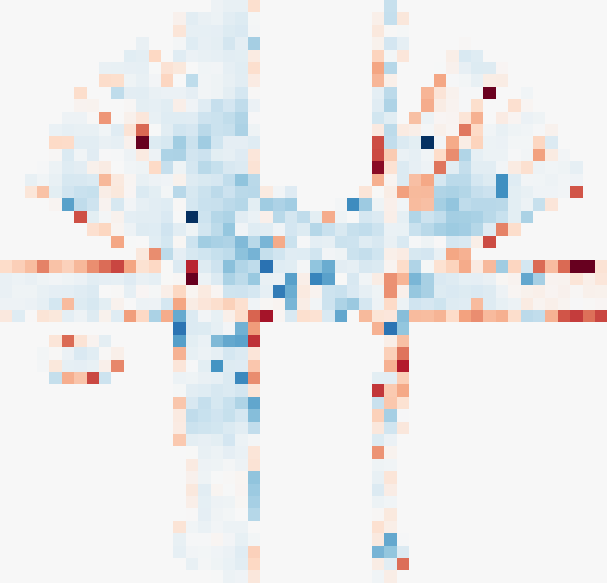}
                    \caption{msPOAS}
            \end{subfigure}
    \end{subfigure}
    \begin{subfigure}[b]{0.01\textwidth}
        \centering
        \includegraphics[height=7cm]{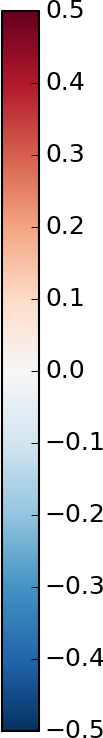}
    \end{subfigure}

    \caption*{FA difference for the phantomas stationary nc-$\chi$ SNR 10
    and spatially varying Rician SNR 15 b1000 datasets.
    Blue values denote underestimation while red values show overestimation of the FA.
    \textbf{Top}~: Stationary nc-$\chi$ noise.
    NLSAM is less biased
    than the other methods in large, homogeneous regions, while the compared methods
    produces more underestimation for the nc-$\chi$ case.
    \textbf{Bottom}~: Spatially variable Rician noise. While being a harder case
    than the SNR 10 dataset since it varies from SNR 5 to 15,
    all methods adapt themselves to some extent to the varying noise profile.}
\end{subfigure}

    \begin{subfigure}[b]{0.49\textwidth}
            \includegraphics[width=\textwidth]{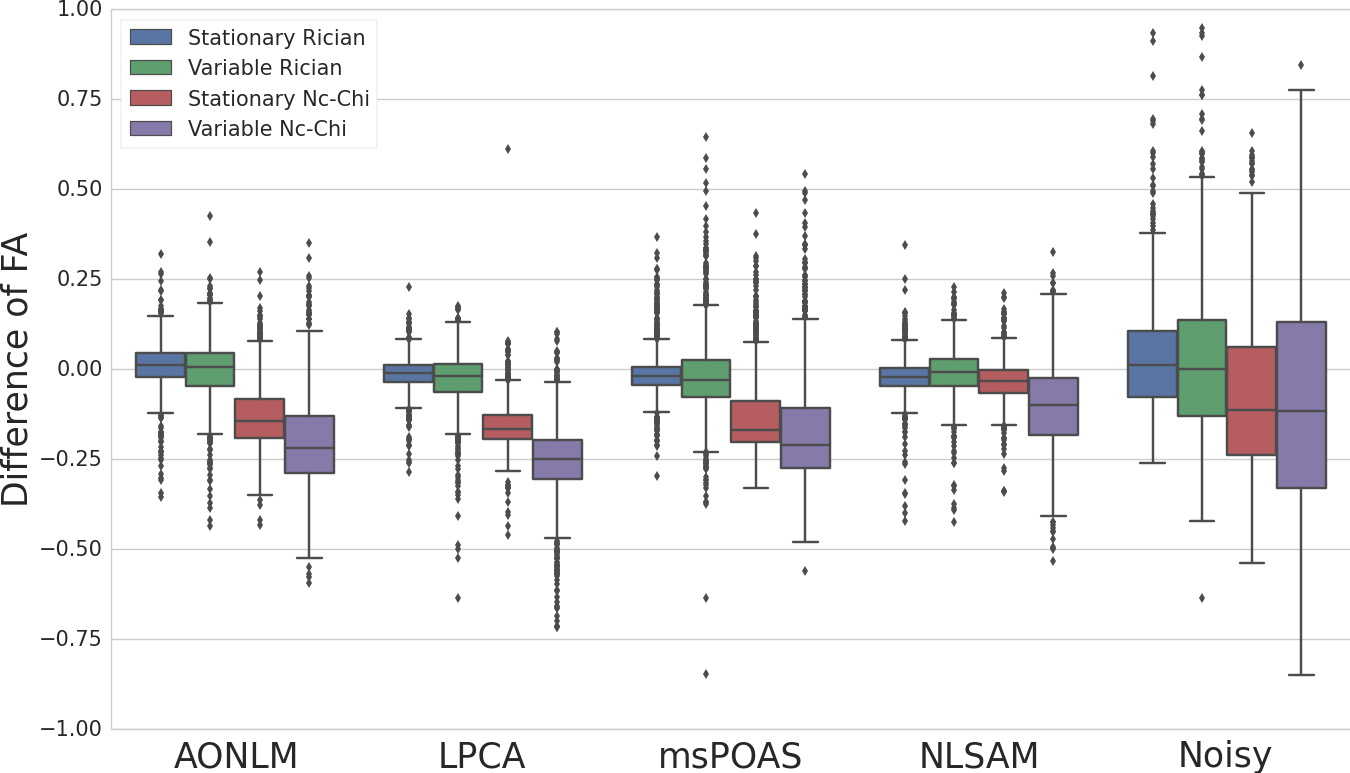}
            \caption*{b1000}
    \end{subfigure}
    \begin{subfigure}[b]{0.49\textwidth}
            \includegraphics[width=\textwidth]{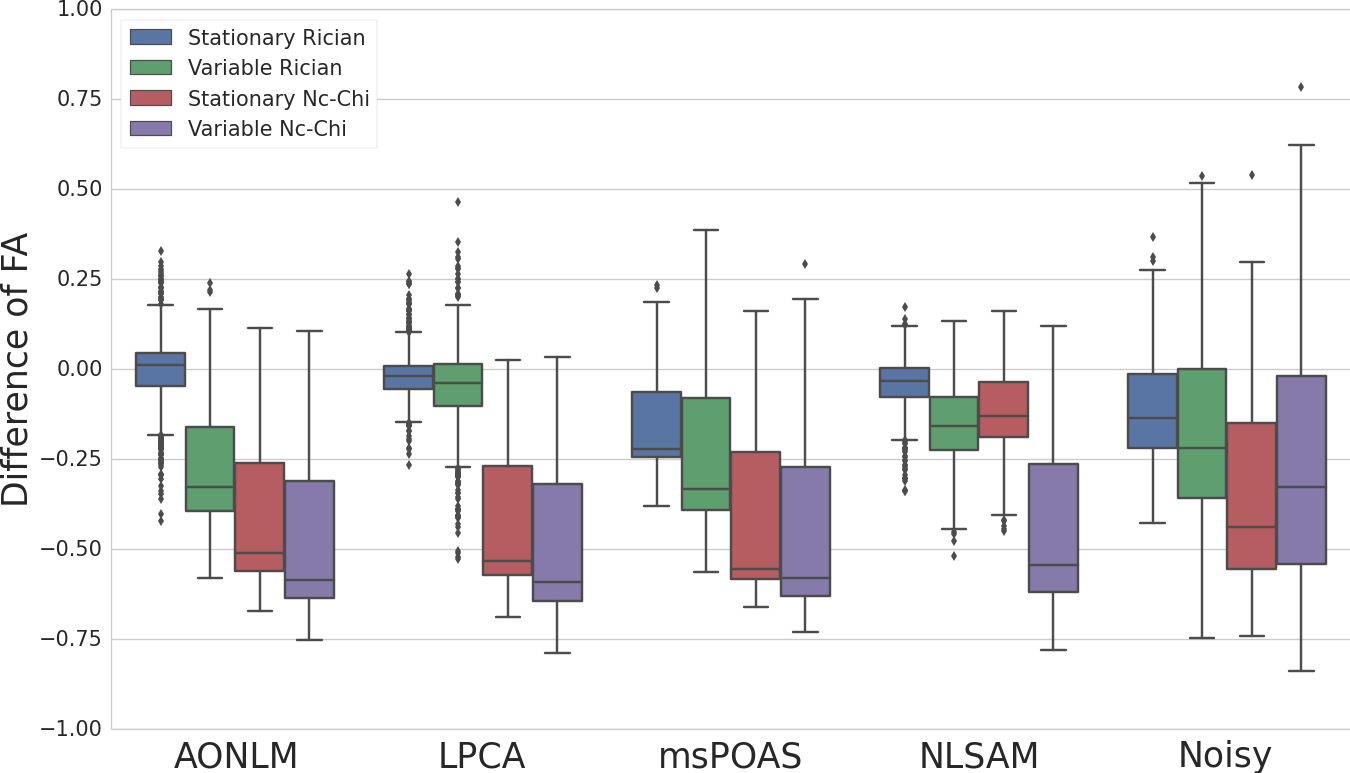}
            \caption*{b3000}
    \end{subfigure}

    \caption{\small Boxplot of the difference in FA for the synthetic datasets at b1000
    (left) and b3000 (right).
    The whiskers show 1.5 times the interquartile range (1.5 x IQR),
    where outliers are plotted individually. The bars represent the first quantile,
    the median and the third quartile.
    No method performs well on the
    nc-$\chi$ b3000 spatially varying noise case, which is the hardest test case.
    NLSAM overall produces less error or is equal to the other
    methods, but has a lower bias in the FA error along noise type.}
    \label{fig:graph_fa}

\end{figure}

As shown by the FA difference map on Fig.~\ref{fig:graph_fa},
our NLSAM method commits a small FA error locally with a smaller
maximum error than the compared methods.
Voxelwise underestimation is denoted in blue and
overestimation in red, where white means the computed value is close to the
expected value.
The noisy data largely overestimates the FA values
for the synthetic datasets, while other denoising methods underestimate the
real FA value most of the time.
On the b1000 datasets, NLSAM has the smallest spread of FA error. The effect of
stabilizing the data prior to denoising can also be seen by the stable FA median error
committed by NLSAM across all noise types.
For the b3000 datasets, the need to correct the intensity bias caused by the noise
becomes more important, as seen by the increased error in underestimating the
correct FA value for most methods.
For the spatially varying Rician noise case, our method commits the lowest
overestimation, as opposed to AONLM and LPCA, which are developed for this
particular noise case.
\review{It is also important to note that in contrast to the other methods,
msPOAS does not explicitly correct for the intensity bias by design,
but rather leaves this correction to subsequent processing steps.}
The SNR 15 nc-$\chi$ noise case is where all the methods
make the biggest error, as they reduce the variance but still suffer from a
large bias in FA.
Overall, our method restores the value of the FA for large bundles
more accurately. We also see that most methods make their largest error near
the partial volume ball mimicking cerebrospinal fluid (CSF).

\subsection{Impact on angular and discrete number of compartments (DNC) error}
\label{sec:results_models}

We now study the angular error and the mean relative error in the discrete
number of compartments (DNC)~\citep{Paquette2014a,Daducci2013}.
The mean relative discrete number of compartments error is defined as
$DNC_i = 100 \times \abs{P_{i_{true}} - P_{i_{est}}} / P_{i_{true}}$ for voxel $i$,
$P_{i_{true}}$ and $P_{i_{est}}$ is the number of crossings respectively
found on the noiseless dataset and on the compared dataset.
All metrics were computed on the voxels containing at least two crossings fibers
on the noiseless dataset shown previously in Fig.~\ref{fig:phantomas}.

\begin{figure}[htb]
\small
\centering
\begin{subfigure}[b]{\textwidth}
            \begin{subfigure}[b]{0.49\textwidth}
                    \includegraphics[width=\textwidth]{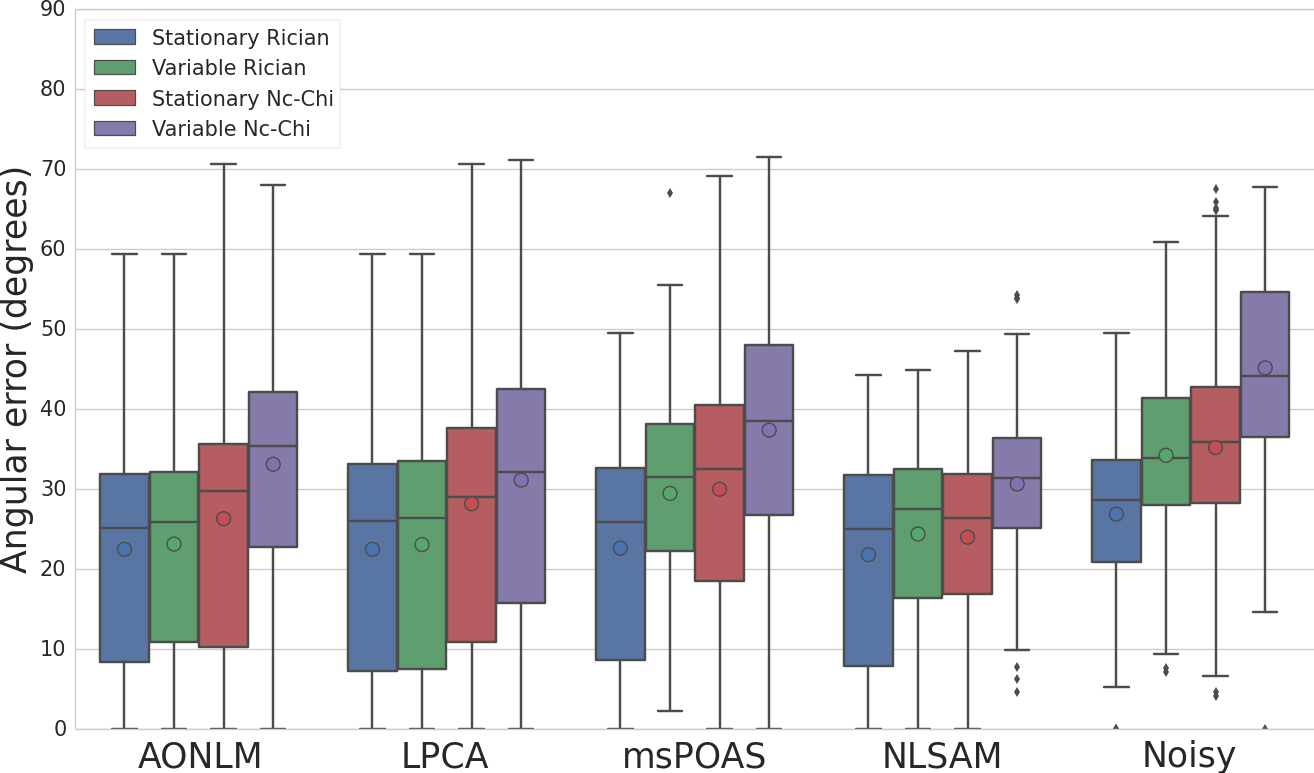}
                    \caption*{b1000}
            \end{subfigure}
            \begin{subfigure}[b]{0.49\textwidth}
                    \includegraphics[width=\textwidth]{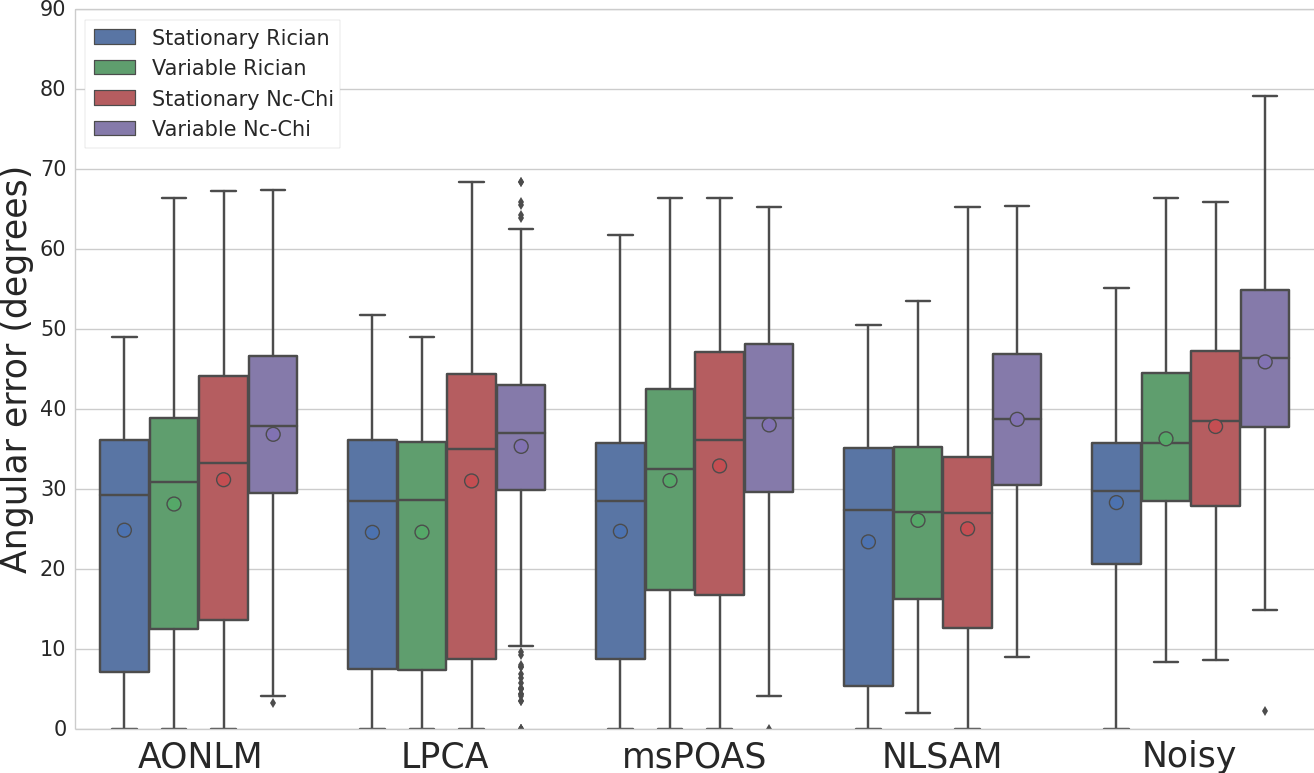}
                    \caption*{b3000}
            \end{subfigure}
    \caption*{Boxplot of the angular error in degrees on the synthetic datasets,
    where the dot represents the mean angular error.
    A low angular error means that the extracted fODFs peaks are aligned with
    the noiseless dataset extracted peaks.}
\end{subfigure}

\begin{subfigure}[b]{\textwidth}

            \begin{subfigure}[b]{0.49\textwidth}
                    \includegraphics[width=\textwidth]{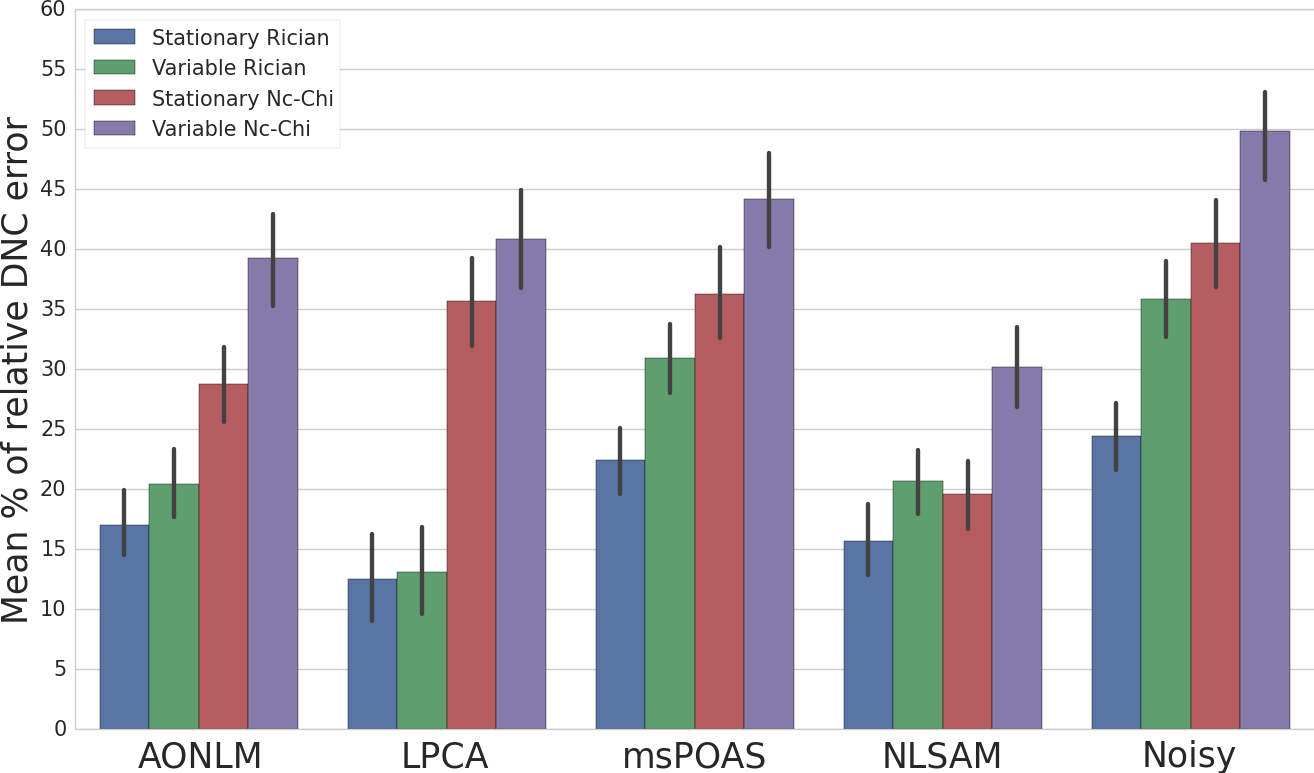}
                     \caption*{b1000}
            \end{subfigure}
            \begin{subfigure}[b]{0.49\textwidth}
                    \includegraphics[width=\textwidth]{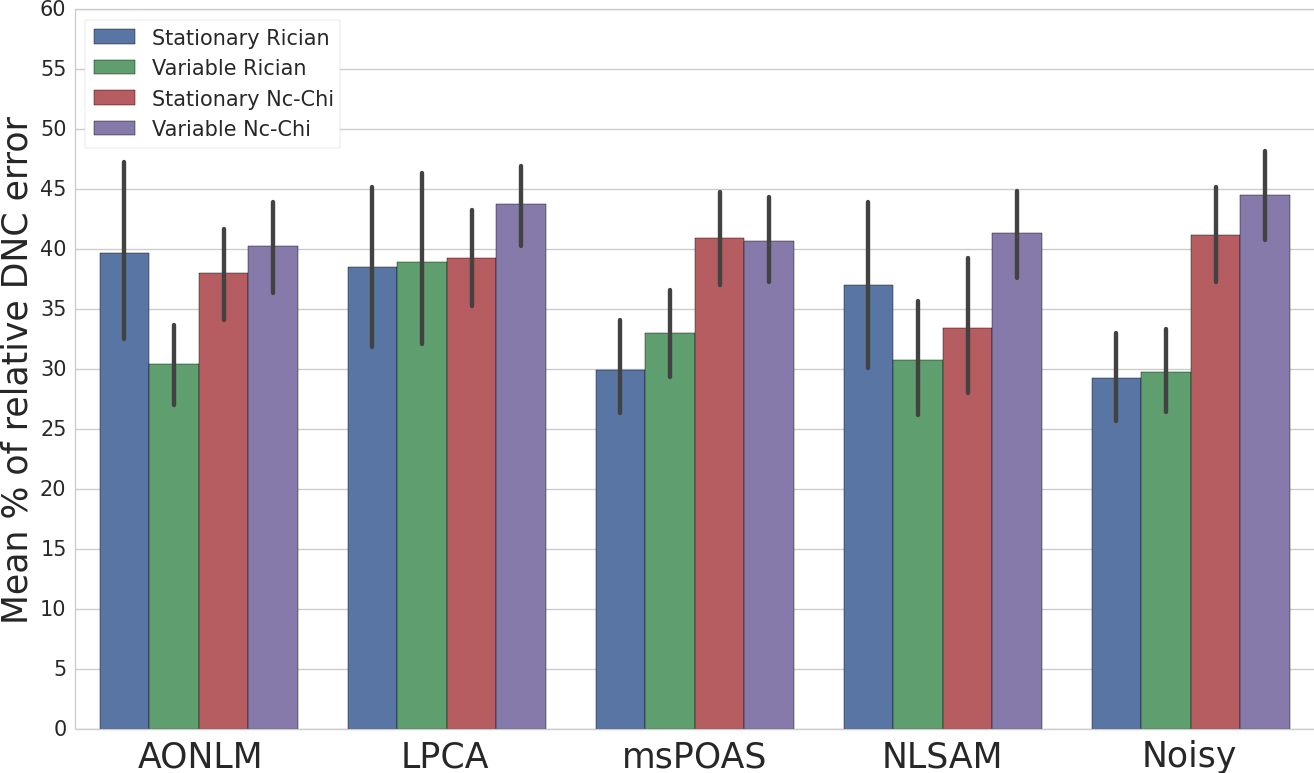}
                    \caption*{b3000}
            \end{subfigure}
\end{subfigure}
    \caption{\small The mean relative percentage of DNC error for the synthetic datasets.
    The bar represents the 95\% confidence interval on the mean as computed by bootstrapping.
    The DNC error is the number of peaks found in excess or missing in each voxel with
    respect to the noiseless dataset.}
    \label{fig:graph_dnc}
\end{figure}

Fig.~\ref{fig:graph_dnc} shows the distribution of the angular error
and of the DNC error found in the
region studied in addition to the mean angular error.
All of the denoising algorithms have a lower median and mean angular error than the
noisy datasets. In addition, the NLSAM denoised datasets have an almost equal or lower
angular error than the other denoising methods, but with a lower maximum
error most of the time as shown by the smaller whiskers.
For the b1000 dataset DNC error, all three of AONLM, LPCA and NLSAM improve on the noisy dataset
for the Rician noise case as they are devised for this kind of data. LPCA also has
a better performance than the other two for the spatially varying Rician noise case,
while NLSAM has a lower mean DNC error for both of the nc-$\chi$ noise case.
The effect of the intensity bias is also seen on msPOAS, where the DNC error is always
lower than the noisy dataset, but also higher than all the other methods which
take into account the intensity bias. The b3000 dataset is much harder, where
no method seems to have a clear advantage in all cases over the others. One
interesting thing to note is that the noisy dataset has a low DNC error for both
of the Rician noise case, but the confidence interval indicates it is in the
same range as the denoised datasets.

\subsection{Impact on tractography}
\label{sec:tracto}

We now show how denoising techniques impact tractography by evaluating the
number of valid bundles (VB), invalid bundles (IB)~\citep{Cote2013a} and
the valid connection to connection ratio (VCCR)~\citep{Girard2014}
found by the tracking
algorithm.
\review{A valid bundle is defined as connecting two ROIs in the ground truth data
while an invalid bundle is a connection made between two ROIs which is not
supported by the ground truth data.}
The valid connection to connection ratio is the
total of valid connections (VC) over the sum of valid and invalid connections
(IC), i.e. $VCCR = \frac{VC}{VC + IC}$.
A good denoising algorithm should find a high number of valid bundles,
a low number of invalid bundles and a high percentage of valid connection
to connection ratio.

\paragraph{Deterministic tractography on the synthetic phantom}

Table~\ref{tbl:tractometer} %
shows the results of deterministic tractography
on the SNR 10, 15 and 20 synthetic datasets for both b1000 and b3000.
The noiseless b1000 dataset had 25/27 valid bundles, 55 invalid bundles and a
valid connection to connection ratio of 65\% and the noiseless b3000 dataset
had 27/27 valid bundles, 40 invalid bundles and a
valid connection to connection ratio of 68\%.
One of the first thing to note is that even though the noisy dataset always has
a high number of valid bundles, it is at the price of a huge number of
invalid bundles.
Moreover, the valid connection to connection ratio is
systematically lower for the SNR 10 datasets than any of the denoising methods.
This indicates that only looking at the number of valid and invalid bundles
does not show how many streamlines reached each region since only at least
one streamline is required to make a connection, thus counting as a valid bundle.
Another observation is that denoising
helps controlling the number of invalid bundles and gives a better
valid connection to connection ratio in most cases over the noisy data.
For the SNR 15 cases, NLSAM has the highest
number of valid bundles in almost all cases, but at the price of a larger
number of invalid bundles at lower SNR. Another interesting trend is the tradeoff
between valid bundles and invalid bundles~: AONLM and LPCA both manage to get
a lower number of invalid bundles, but also tend to have a lower number of valid bundles
than msPOAS or NLSAM overall.

For the SNR 20 stationary noise cases, all methods are close in valid bundles
with some difference in the number of
invalid bundles.
This shows that tractography could
benefit from variable tracking parameters instead of fixed
values depending on the preferred trade-off for the task at hand~\citep{Chamberland2014b}.

\begin{table}[htbp]
\centering
\small
\resizebox{\textwidth}{!}{
\begin{tabular}{ccccccccc@{\hskip 6\tabcolsep}cccccc}

\toprule
\multicolumn{3}{c}{} & \multicolumn{6}{c}{\textbf{Stationary noise}\hspace*{1cm}} & \multicolumn{6}{c}{\textbf{Spatially variable noise}}\\
\cmidrule(lr{\dimexpr 6\tabcolsep+0.5em}){4-9}
\cmidrule(lr){10-15}
\multicolumn{3}{c}{} & \multicolumn{3}{c}{\textbf{SNR 10}} & \multicolumn{3}{c}{\textbf{SNR 20}\hspace*{1cm}} & \multicolumn{3}{c}{\textbf{SNR 15}} & \multicolumn{3}{c}{\textbf{SNR 20}}\\
\cmidrule(lr){4-6}
\cmidrule(lr{\dimexpr 6\tabcolsep+0.5em}){7-9}
\cmidrule(lr){10-12}
\cmidrule(lr){13-15}
\multicolumn{3}{c}{Method / Noise} & VB & IB & VCCR & VB & IB & VCCR  & VB & IB & VCCR & VB & IB & VCCR\\
\midrule
\multirow{4}{*}{\textbf{AONLM}}
& \multirow{2}{*}{b1000}
   & Rician & 25 & 78 & 49\% & 25 & 75 & 51\% & 23 &  91 & 45\% & 25 & 89 & 50\% \\
& & nc-$\chi$ & 25 & 88 & 50\% & 26 & 88 & 52\% & 21 & 111 & 44\% & 23 & 93 & 47\% \\

& \multirow{2}{*}{b3000}
   & Rician & 25 & 69 & 52\% & 25 & 60 & 56\% & 24 & 85 & 50\% & 26 & 72 & 52\% \\
& & nc-$\chi$ & 25 & 78 & 55\% & 26 & 67 & 55\% & 20 & 95 & 48\% & 22 & 78 & 54\% \\

\addlinespace
\multirow{4}{*}{\textbf{LPCA}}
& \multirow{2}{*}{b1000}
  & Rician & 23 & 61 & 49\% & 25 & 64 & 54\% & 16 & 36 & 42\% & 18 & 38 & 45\% \\
& & nc-$\chi$ & 22 & 66 & 50\% & 24 & 70 & 54\% & 16 & 46 & 51\% & 20 & 56 & 52\% \\

& \multirow{2}{*}{b3000}
  & Rician & 23 & 44 & 47\% & 26 & 46 & 53\% & 17 & 37 & 42\% & 19 & 41 & 45\% \\
& & nc-$\chi$ & 20 & 42 & 58\% & 25 & 57 & 53\% & 18 & 40 & 55\% & 20 & 56 & 55\% \\
\addlinespace

\multirow{4}{*}{\textbf{msPOAS}}
& \multirow{2}{*}{b1000}
  & Rician & 25 & 101 & 49\% & 25 & 89 & 52\% & 23 & 129 & 44\% & 25 & 118 & 46\% \\
& & nc-$\chi$ & 23 & 121 & 40\% & 25 & 95 & 54\% & 20 & 131 & 35\% & 25 & 141 & 41\% \\

& \multirow{2}{*}{b3000}
  & Rician & 26 & 108 & 53\% & 26 & 74 & 58\% & 25 & 88 & 52\% & 25 & 93 & 49\% \\
& & nc-$\chi$ & 17 & 84 & 37\% & 25 & 84 & 57\% & 22 & 96 & 33\% & 23 & 94 & 47\% \\
\addlinespace

\multirow{4}{*}{\textbf{NLSAM}}
& \multirow{2}{*}{b1000}
  & Rician & 25 & 90 & 49\% & 26 & 96 & 54\% & 25 & 127 & 42\% & 25 &  114 & 45\% \\
& & nc-$\chi$ & 25 & 120 & 48\% & 25 & 90 & 54\% &  25 & 170 & 28\% & 26 & 144 & 43\% \\

& \multirow{2}{*}{b3000}
  & Rician & 25 & 92 & 50\% & 26 & 67 & 54\% & 25 & 108 & 43\% & 25 &  97 & 47\% \\
& & nc-$\chi$ & 23 & 100 & 45\% & 24 & 82 & 53\% &  23 & 173 & 29\% & 25 & 131 & 37\% \\
\addlinespace

\multirow{4}{*}{\textbf{Noisy}}
& \multirow{2}{*}{b1000}
  & Rician & 25 & 138 & 41\% & 25 & 107 & 53\% & 25 & 159 & 36\% & 25 & 134 & 42\% \\
& & nc-$\chi$ & 25 & 166 & 34\% & 26 & 119 & 49\% &  17 & 120 &  9\% & 25 & 209 & 24\% \\

& \multirow{2}{*}{b3000}
  & Rician & 25 & 116 & 46\% & 27 & 87 & 54\% & 25 & 160 & 36\% & 25 & 149 & 42\% \\
& & nc-$\chi$ & 25 & 182 & 36\% & 26 & 103 & 53\% & 18 & 124 &  9\% & 25 & 210 & 24\% \\

\cmidrule(lr{55pt}){1-9}
\multirow{3}{*}{\textbf{Noiseless}}

&       & & \multicolumn{2}{r}{VB} & \multicolumn{2}{c}{IC} & \multicolumn{2}{l}{VCCR} & & & & & & \\
& b1000 & & \multicolumn{2}{r}{25} & \multicolumn{2}{c}{55} & \multicolumn{2}{l}{65\%} & & & & & & \\
& b3000 & & \multicolumn{2}{r}{27} & \multicolumn{2}{c}{40} & \multicolumn{2}{l}{68\%} & & & & & & \\
\bottomrule
\end{tabular}}

\caption{\small Tractometer results for the deterministic tracking.}

\label{tbl:tractometer}
\end{table}

% \FloatBarrier

%
\paragraph{Tracking the real data}

We now look at tractography on the \textit{in-vivo} high spatial resolution
dataset and its clinical spatial resolution counterpart of the same subject
previously shown on Fig.~\ref{fig:1_2mm}.
The high spatial resolution dataset at 1.2 mm isotropic has 40 unique
gradient directions while the lower spatial
resolution dataset at 1.8 mm isotropic
has 64 unique gradient directions for a comparable acquisition time.
The background is masked by the scanner and has a spatially varying Rician noise
profile due to the SENSE reconstruction, which is the specific noise case
covered by the AONLM and LPCA denoising algorithm. We use the deterministic
tractography algorithm from~\citep{Girard2014}, which considers anatomical
constraints for more anatomically plausible tractography.
Fig.~\ref{fig:bundles} (which is displayed in landscape to present
all the bundles at once) shows from top to bottom the left arcuate fasciculus (AF),
the inferior fronto-occipital fasciculus (IFOF)
and the corticospinal tract (CST) as dissected automatically by the Tract
querier~\citep{Wassermann2013a}.
The noisy 1.2 mm AF stops prematurely in the frontal part of the bundle,
while the 1.8 mm noisy AF misses the temporal lobe.
In contrast, the streamlines from the
NLSAM denoised bundle go further into the temporal lobe.
Also note how the right IFOF has a better
coverage for all the 1.2 mm datasets and more fanning near the
front of the brain than the noisy 1.8 mm dataset.
We also see that the left IFOF is thinner than its right counterpart,
but most of the bundles tracked from the denoised datasets
produces less spurious tracks while keeping the anatomical details.
The LPCA denoised IFOF stops prematurely for \review{the left posterior} part of the bundle,
possibly because of a lost crossing region along
the fibers during the denoising process.
The CST does show some commissural fibers through the pons in the noisy 1.8 mm
dataset, while they are present but look like spurious fibers on the noisy
1.2 mm dataset. AONLM can recover some of those commissural fibers,
while NLSAM is the only algorithm which recovers
them in addition to richer fanning near both sides of the motor cortex.

\begin{landscape}
\thispagestyle{empty}
\begin{figure}[htb]
\small
\centering
            \begin{subfigure}[t]{0.16\linewidth}
                    \includegraphics[width=\textwidth,height=2.1cm,keepaspectratio=false]{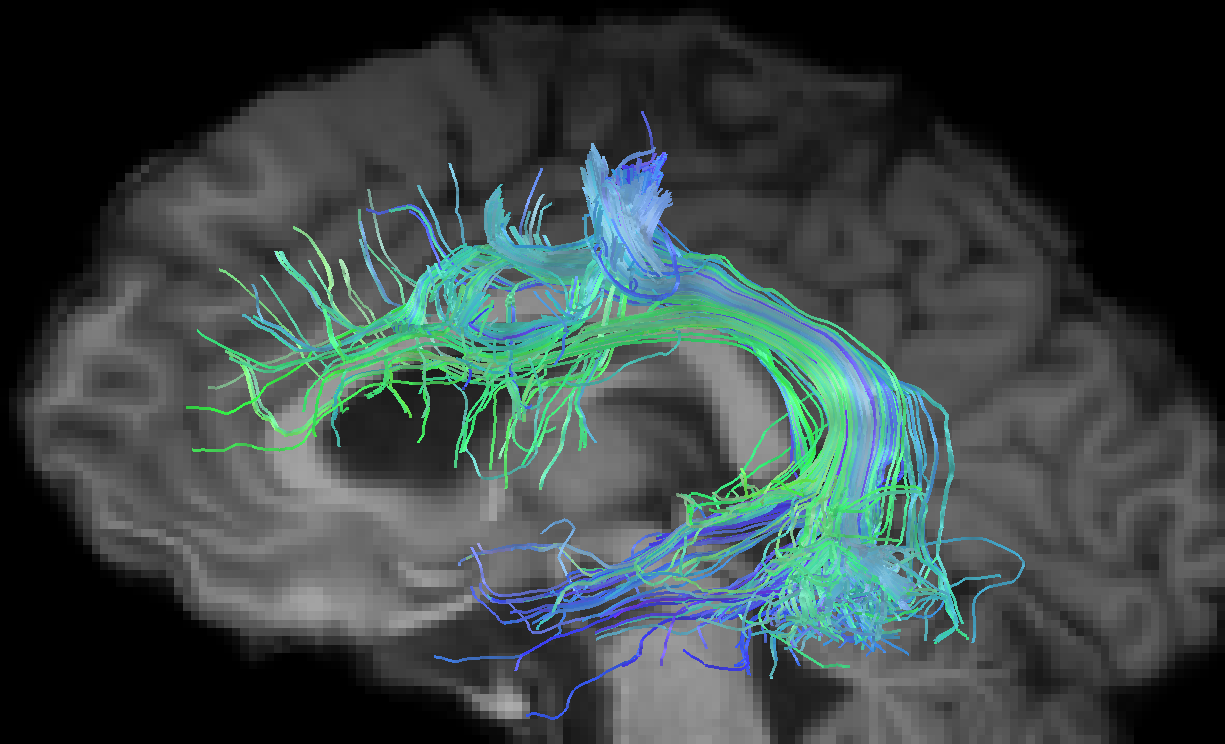}
            \end{subfigure}
            \begin{subfigure}[t]{0.16\linewidth}
                    \includegraphics[width=\textwidth,height=2.1cm,keepaspectratio=false]{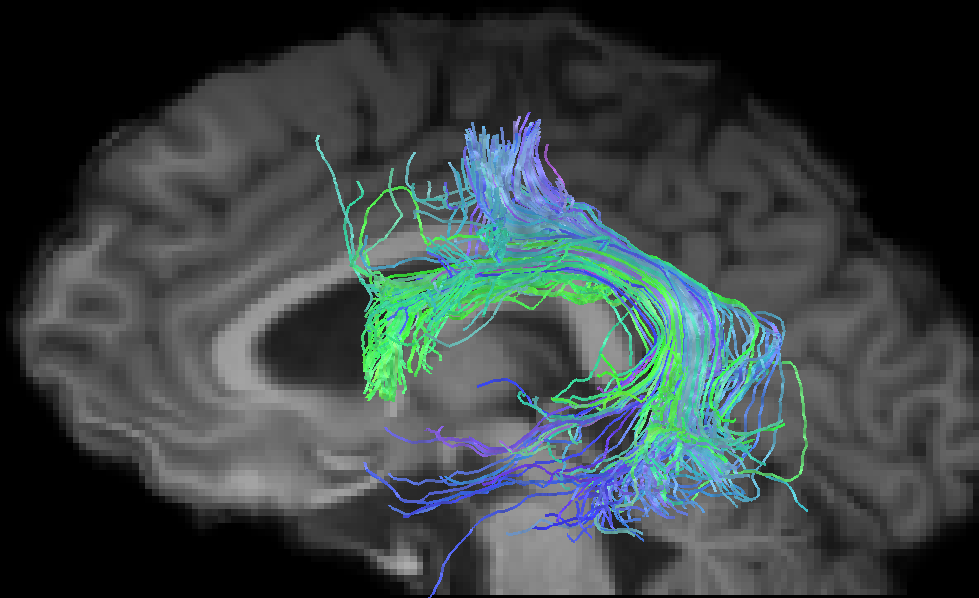}
            \end{subfigure}
            \begin{subfigure}[t]{0.16\linewidth}
                    \includegraphics[width=\textwidth,height=2.1cm,keepaspectratio=false]{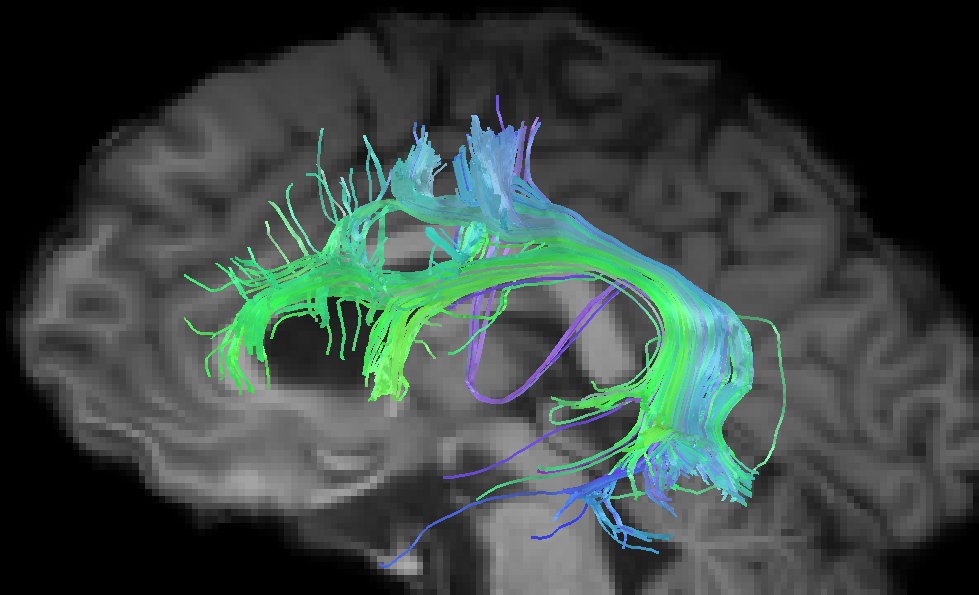}
            \end{subfigure}
            \begin{subfigure}[t]{0.16\linewidth}
                    \includegraphics[width=\textwidth,height=2.1cm,keepaspectratio=false]{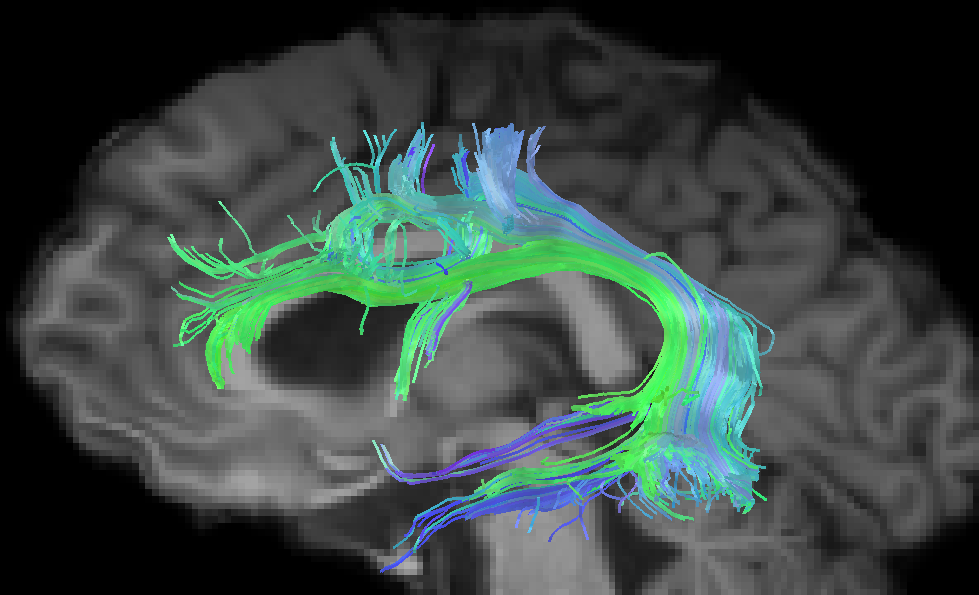}
            \end{subfigure}
            \begin{subfigure}[t]{0.16\linewidth}
                    \includegraphics[width=\textwidth,height=2.1cm,keepaspectratio=false]{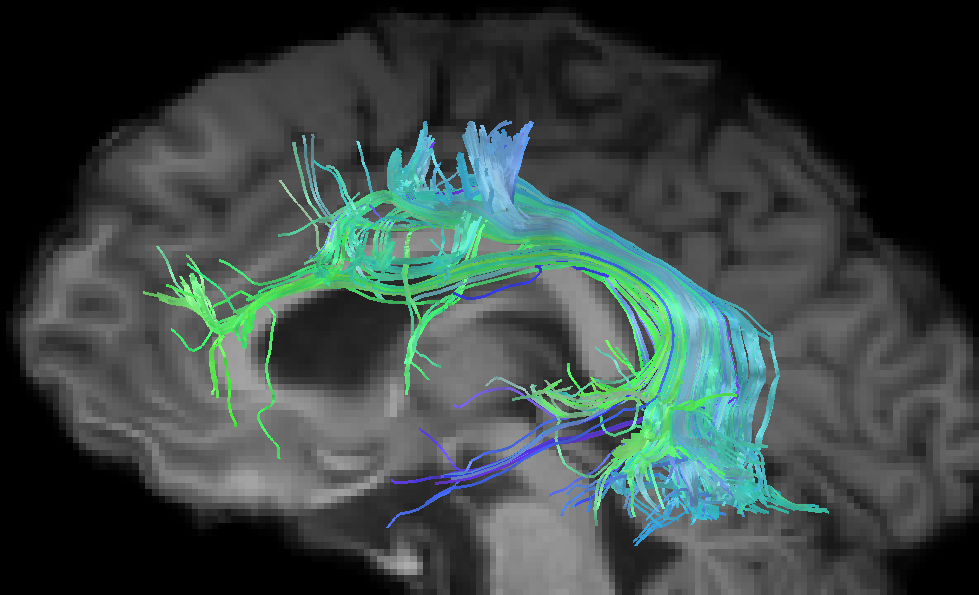}
            \end{subfigure}
            \begin{subfigure}[t]{0.16\linewidth}
                    \includegraphics[width=\textwidth,height=2.1cm,keepaspectratio=false]{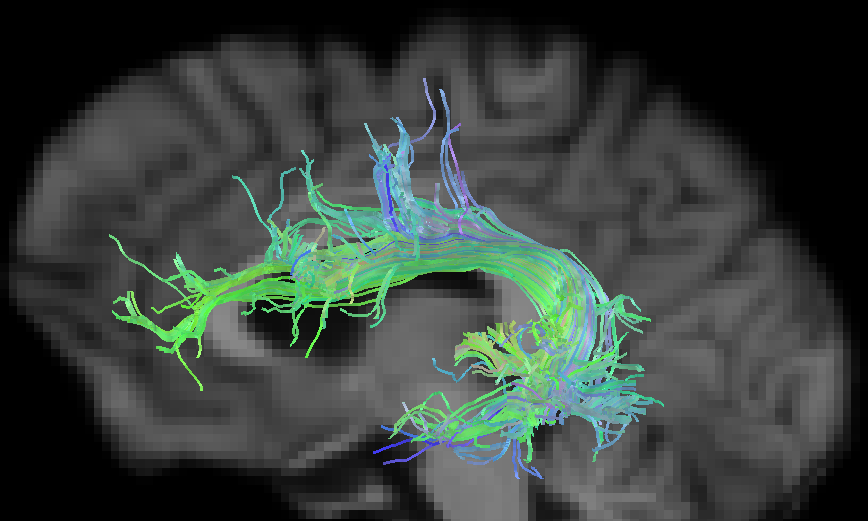}
            \end{subfigure}

            \begin{subfigure}[c]{0.16\linewidth}
                    \includegraphics[width=\textwidth,height=4cm,keepaspectratio=false]{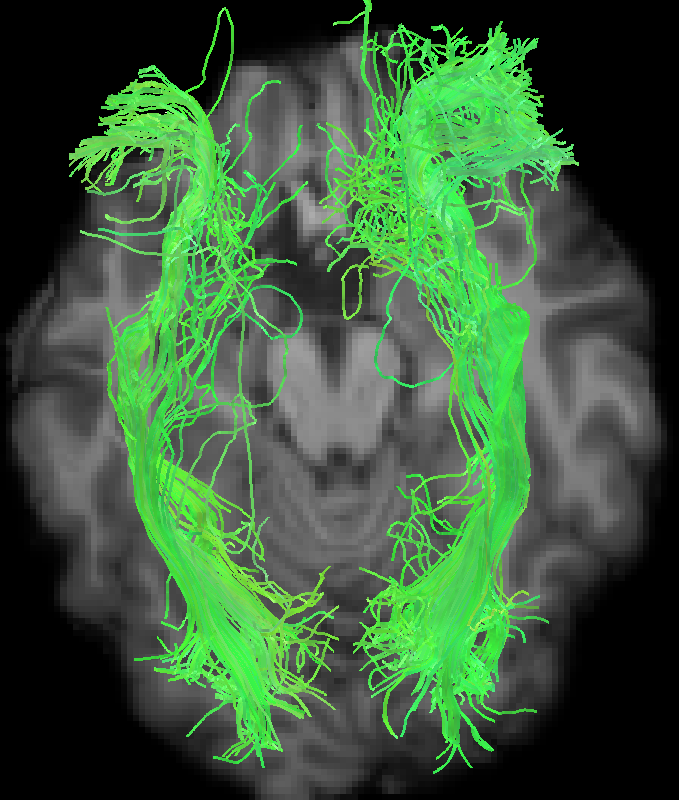}
            \end{subfigure}
            \begin{subfigure}[c]{0.16\linewidth}
                    \includegraphics[width=\textwidth,height=4cm,keepaspectratio=false]{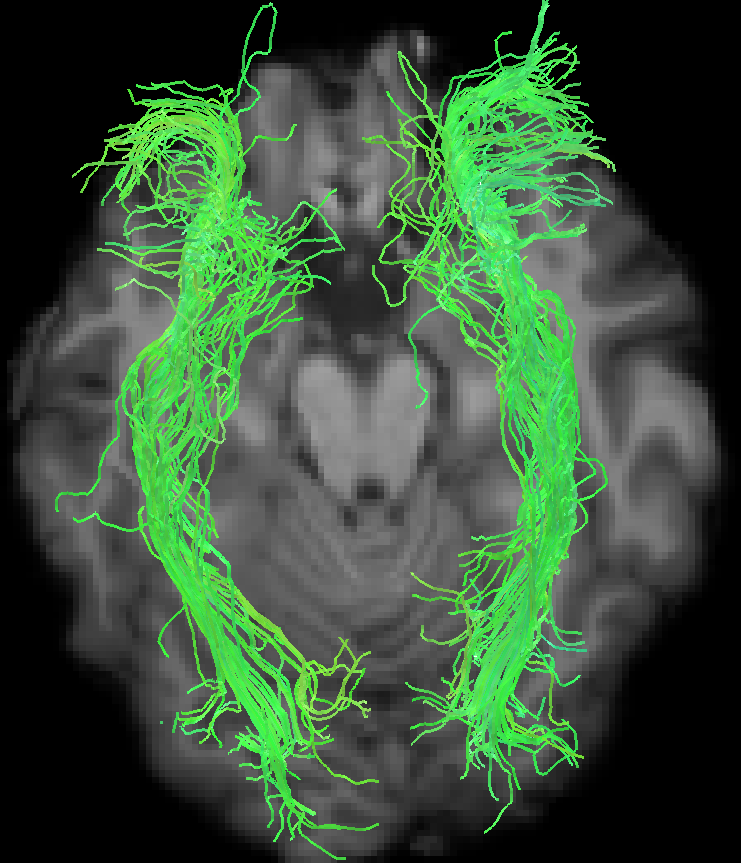}
            \end{subfigure}
            \begin{subfigure}[c]{0.16\linewidth}
                    \includegraphics[width=\textwidth,height=4cm,keepaspectratio=false]{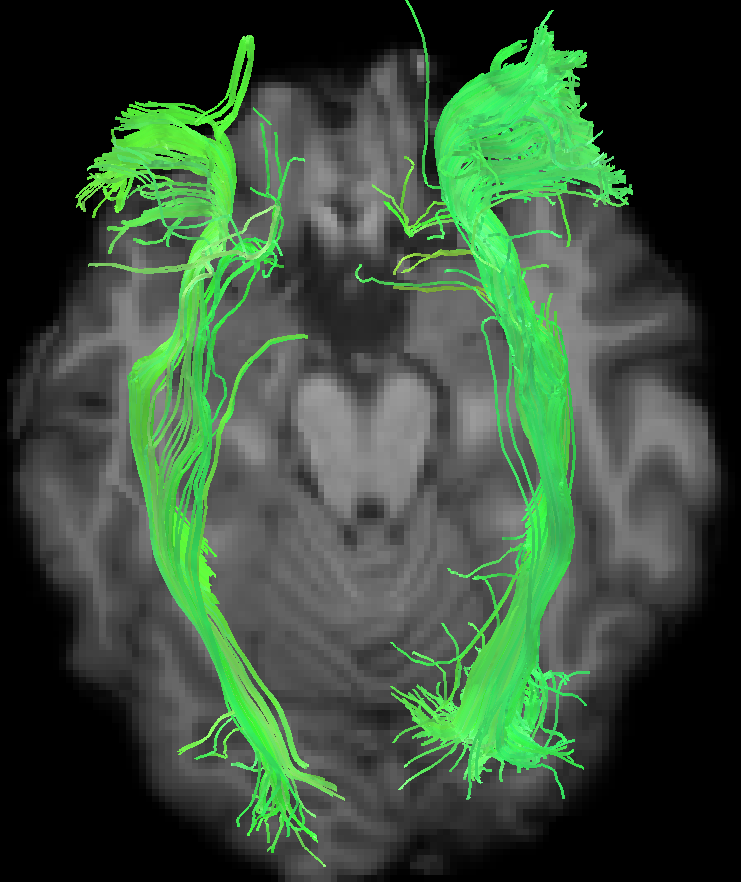}
            \end{subfigure}
            \begin{subfigure}[c]{0.16\linewidth}
                    \includegraphics[width=\textwidth,height=4cm,keepaspectratio=false]{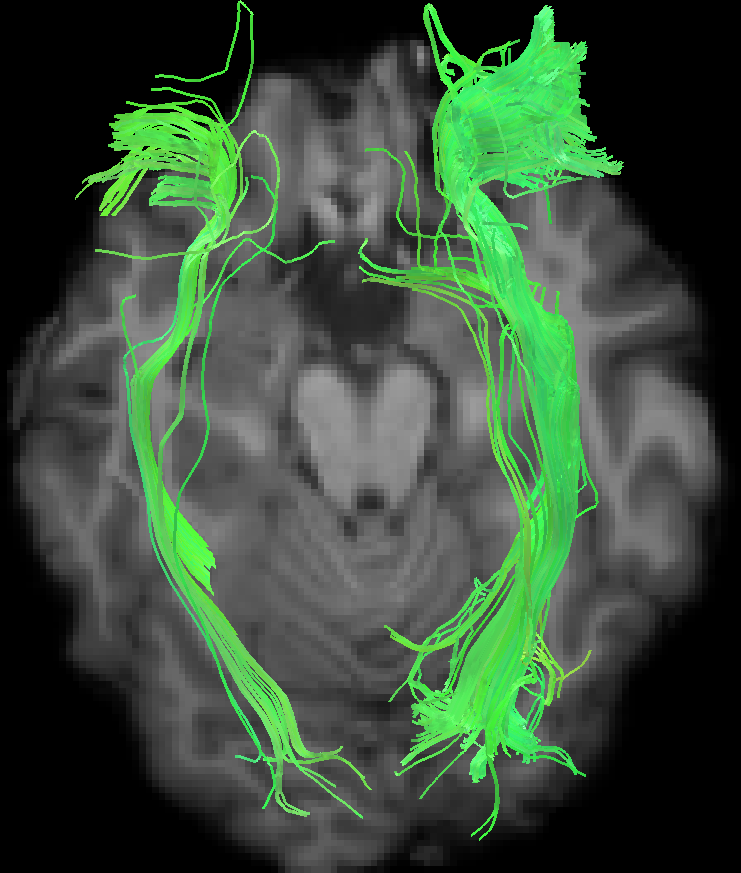}
            \end{subfigure}
            \begin{subfigure}[c]{0.16\linewidth}
                    \includegraphics[width=\textwidth,height=4cm,keepaspectratio=false]{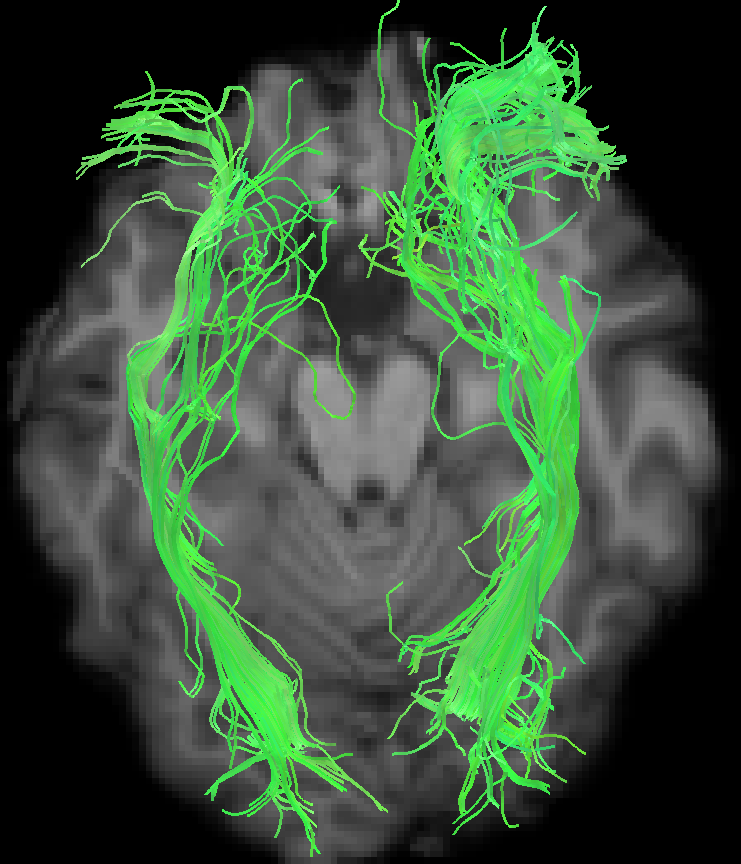}
            \end{subfigure}
            \begin{subfigure}[c]{0.16\linewidth}
                    \includegraphics[width=\textwidth,height=4cm,keepaspectratio=false]{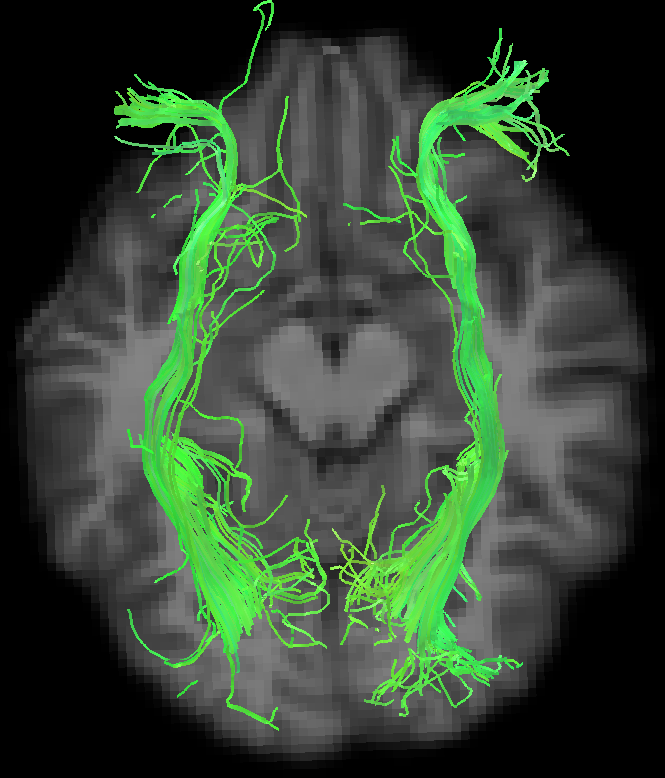}
            \end{subfigure}

            \begin{subfigure}[b]{0.16\linewidth}
                    \includegraphics[width=\textwidth,height=2.85cm,keepaspectratio=false]{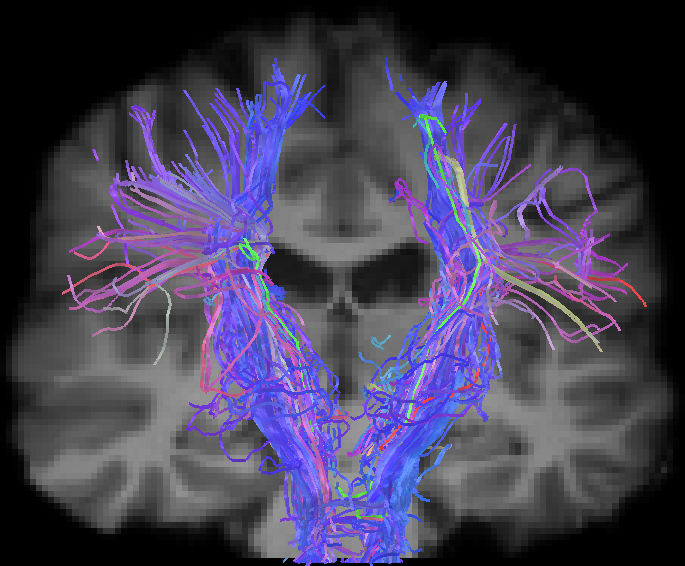}
                    \caption{NLSAM}
            \end{subfigure}
            \begin{subfigure}[b]{0.16\linewidth}
                    \includegraphics[width=\textwidth,height=2.85cm,keepaspectratio=false]{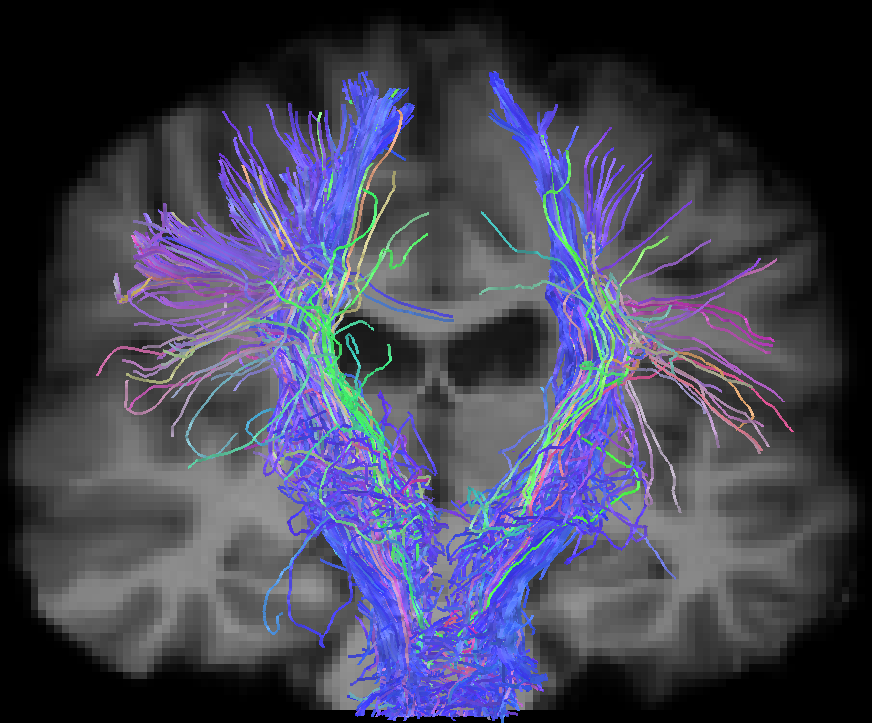}
                    \caption{Noisy 1.2 mm}
            \end{subfigure}
            \begin{subfigure}[b]{0.16\linewidth}
                    \includegraphics[width=\textwidth,height=2.85cm,keepaspectratio=false]{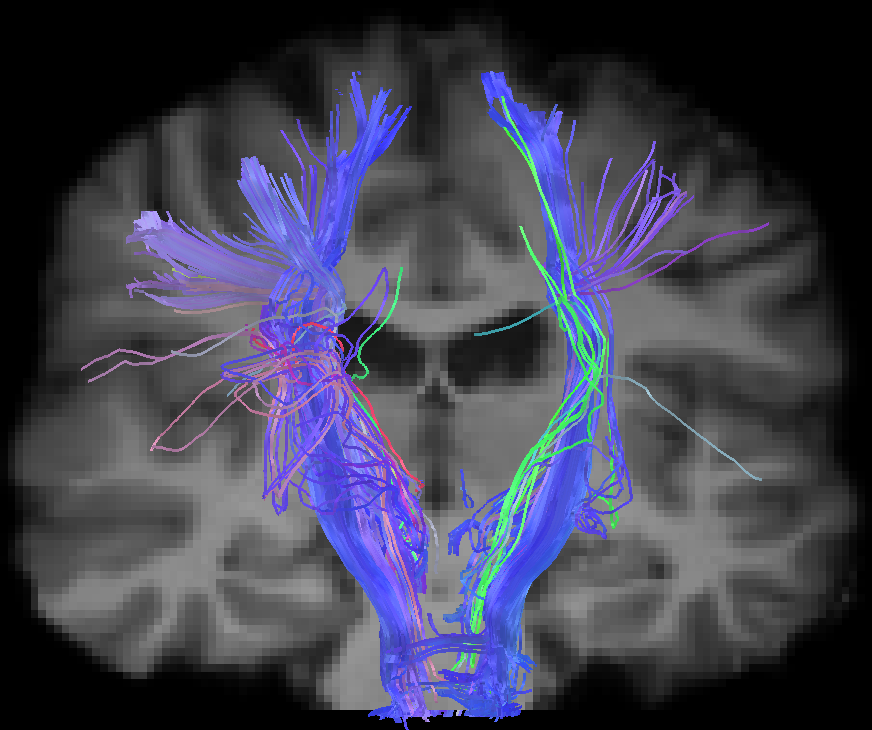}
                    \caption{AONLM}
            \end{subfigure}
            \begin{subfigure}[b]{0.16\linewidth}
                    \includegraphics[width=\textwidth,height=2.85cm,keepaspectratio=false]{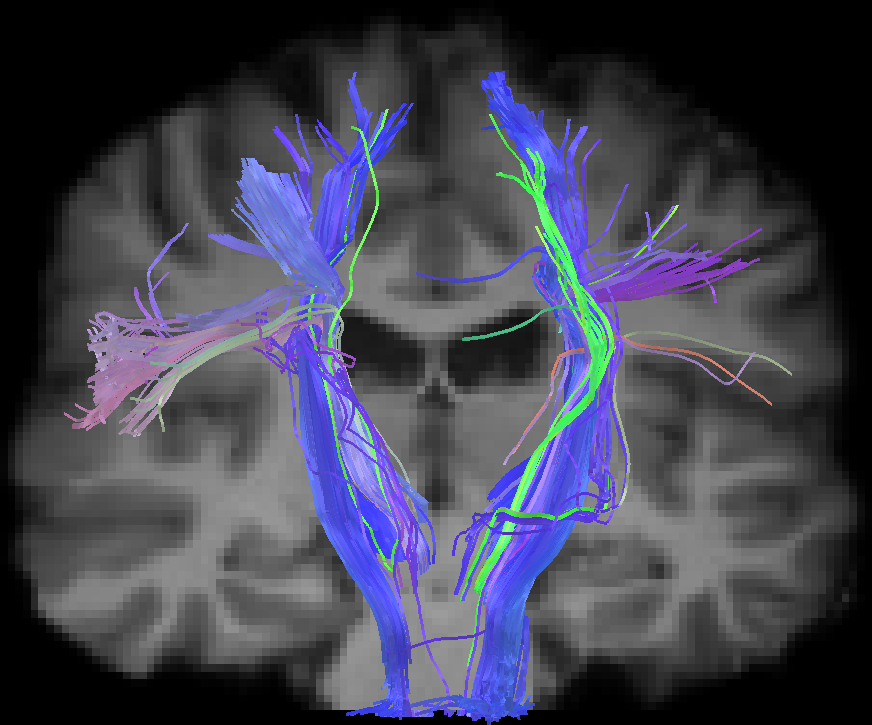}
                    \caption{LPCA}
            \end{subfigure}
            \begin{subfigure}[b]{0.16\linewidth}
                    \includegraphics[width=\textwidth,height=2.85cm,keepaspectratio=false]{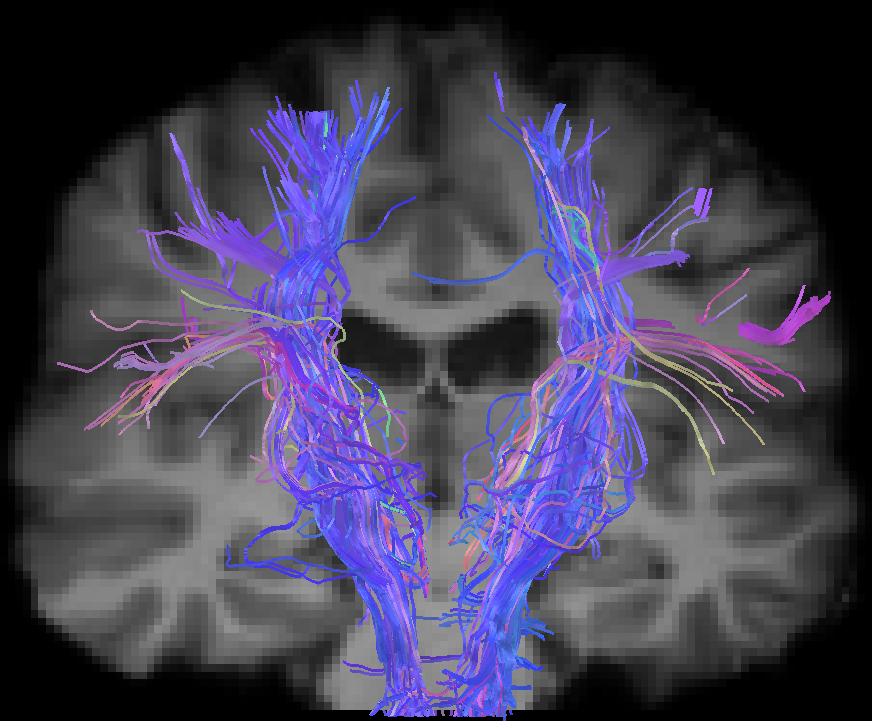}
                    \caption{msPOAS}
            \end{subfigure}
            \begin{subfigure}[b]{0.16\linewidth}
                    \includegraphics[width=\textwidth,height=2.85cm,keepaspectratio=false]{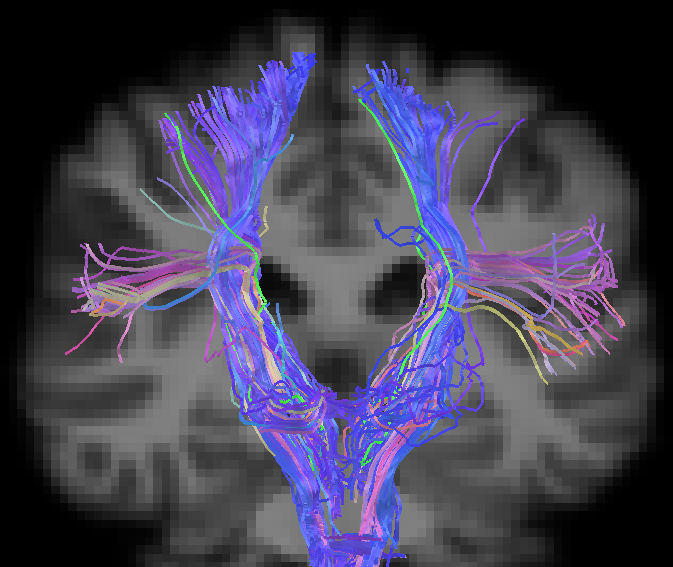}
                    \caption{Noisy 1.8 mm}
            \end{subfigure}

    \caption{Deterministic tractography for selected bundles on the \textit{in-vivo} dataset.
    We also show a T1 weighted image aligned in the diffusion space for anatomical reference.
    \textbf{Top}~: The left arcuate fasciculus. Note how the denoised NLSAM arcuate fasciculus
    goes further into the frontal and temporal region
    than both of its noisy 1.2 mm and 1.8 mm counterparts.
    \textbf{Middle}~: The inferior fronto-occipital fasciculus. The AONLM denoised bundle
    has a denser part for the
    right IFOF while the LPCA bundle stops prematurely for the left IFOF, possibly due to a
    missing crossing along the bundle.
    \textbf{Bottom}~: The corticospinal tract. We see that NLSAM recovers the commissural fibers
    in the pons
    from the noisy 1.2 mm dataset, which are not even present on the noisy 1.8 mm dataset nor on the other
    denoising algorithm's bundles. NLSAM also recovers more fanning to both sides of the brain
    than all the compared methods.}

\label{fig:bundles}
\floatfoot{\thepage}
\end{figure}
\end{landscape}

\FloatBarrier

\section{Discussion}
\label{sec:discuss}
\subsection{Enhancing the raw data}

We quantitatively showed in Fig.~\ref{fig:graph_percep} that denoising restores perceptual
information when compared to the unprocessed noisy data.
Taking the spatially varying aspect and the particular
nature of the noise into account is also important since
modern scanners implement parallel imaging which changes the nature of the
noise~\citep{Dietrich2008}, leading to a lower performance for denoising methods
not fully taking into account the introduced bias.
Fig.~\ref{fig:1_2mm}
shows that this is also qualitatively true for \textit{in-vivo} data, where denoising
visually restores information in regions heavily corrupted by noise.
While perceptual metrics might indicate the performance of an algorithm, one must
remember that the relative signal difference is of interest in diffusion MRI,
which is not fully captured by perceptual metrics like the PSNR or the SSIM.
One is also usually interested in diffusion MRI metrics as opposed to
perceptual information brought by the raw diffusion MRI datasets.
\review{AONLM is able to remove most of the noise, but still shows some residuals near
the inferior part of the brain, possibly due to only considering the 3D volumes separately,
which means that the algorithm can not benefit from the additional angular information
brought by diffusion MRI.
LPCA can restore visual information and sharp edges from the noisy dataset,
but the region in the pons, where the noise level is higher and crossing fibers are more complex,
also seems to be piecewise constant.
This might arise from the fact that the algorithm uses all of the DWIs at once for its PCA decomposition step
and treats all intensities at the same in the noise removal step.
As for NLSAM, the algorithm only works in the local angular domain, thus exploiting similar
contrast and redundant edge structure under different noise realization.
msPOAS also uses a similar idea, where the angular similarity is weighted according to the Kullback-Leibler divergence
to control the importance of dissimilar intensities in the denoising process.}
Nevertheless, these \review{perceptual} metrics show that denoising improves upon the noisy data,
but one should also look at metrics derived from the studied object
of interest i.e. tensor or fODF derived metrics, since high perceptual metrics
might also reflect blurring of diffusion features, which is the main interest
in this type of acquisition rather than the perceived quality.

\subsection{Impact of the stabilization algorithm on the compared denoising methods}

\begin{figure}[htb]
\small
\centering

            \begin{subfigure}[b]{0.19\textwidth}
                    \includegraphics[width=\textwidth]{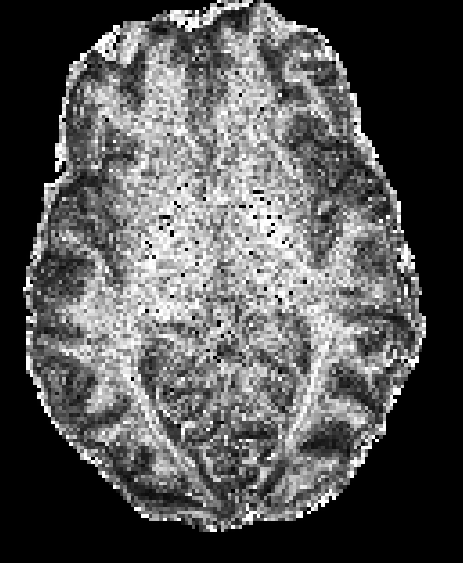}
            \end{subfigure}
            \begin{subfigure}[b]{0.19\textwidth}
                    \center
                    \includegraphics[width=\textwidth]{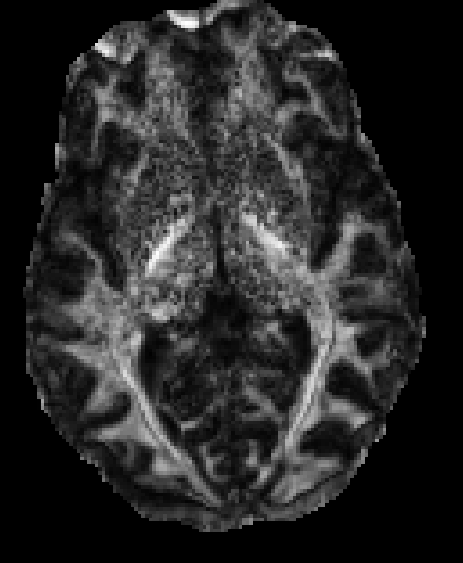}
            \end{subfigure}
            \begin{subfigure}[b]{0.19\textwidth}
                    \center
                    \includegraphics[width=\textwidth]{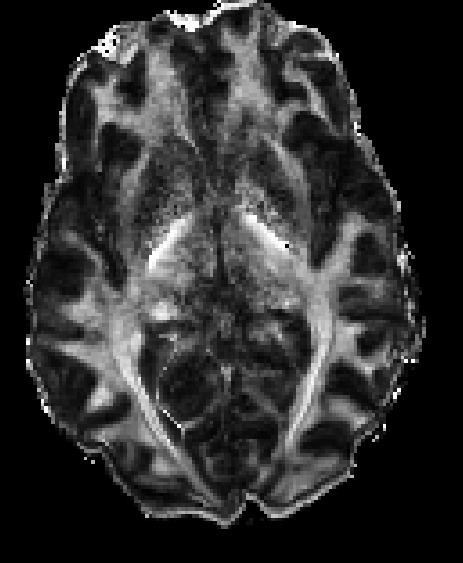}
            \end{subfigure}
            \begin{subfigure}[b]{0.19\textwidth}
                    \includegraphics[width=\textwidth]{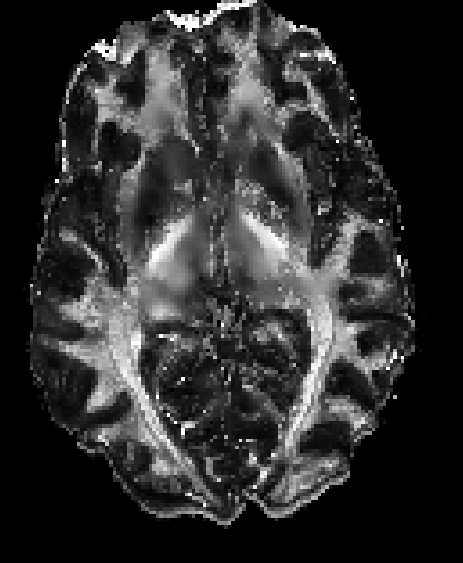}
            \end{subfigure}
            \begin{subfigure}[b]{0.19\textwidth}
                    \center
                    \includegraphics[width=\textwidth]{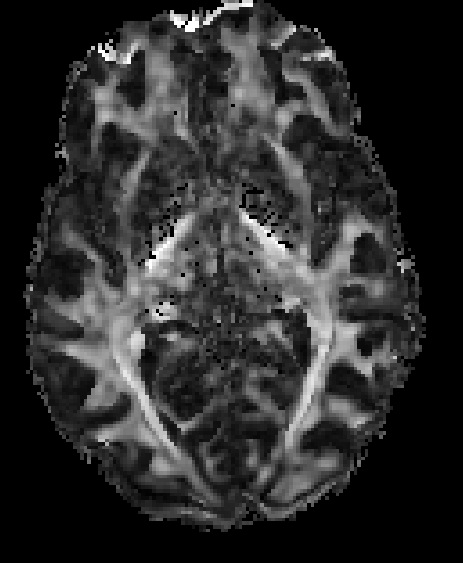}
            \end{subfigure}

            \begin{subfigure}[b]{0.19\textwidth}
                    \includegraphics[width=\textwidth]{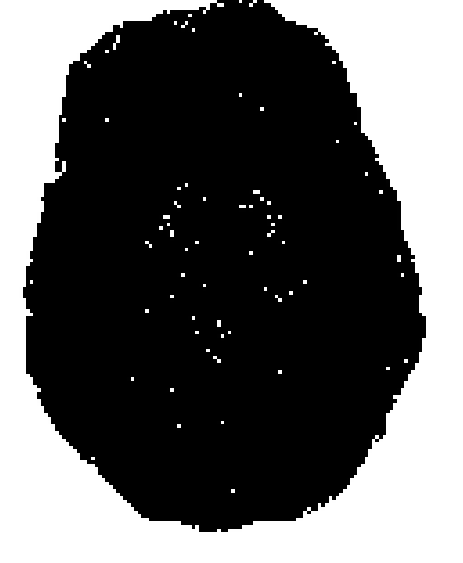}
                    \caption{Noisy}
            \end{subfigure}
            \begin{subfigure}[b]{0.19\textwidth}
                    \includegraphics[width=\textwidth]{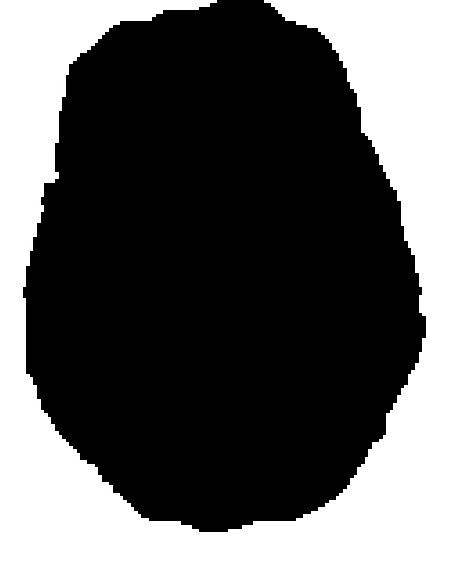}
                    \caption{NLSAM}
            \end{subfigure}
            \begin{subfigure}[b]{0.19\textwidth}
                    \center
                    \includegraphics[width=\textwidth]{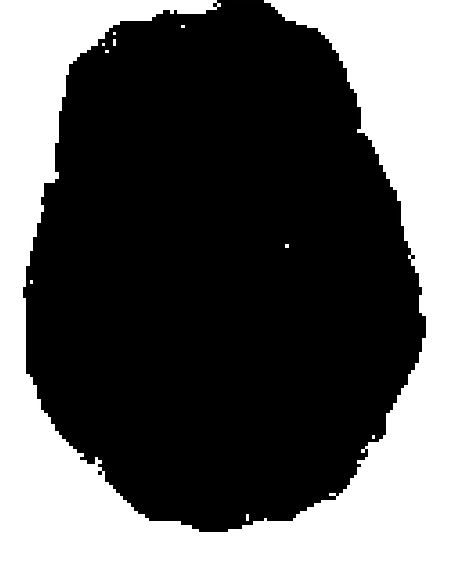}
                    \caption{AONLM}
            \end{subfigure}
            \begin{subfigure}[b]{0.19\textwidth}
                    \includegraphics[width=\textwidth]{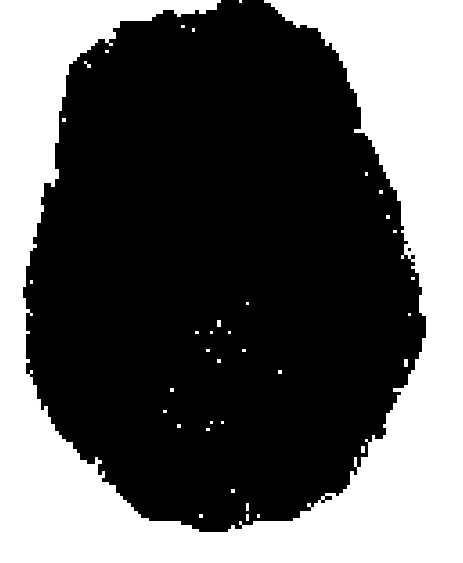}
                    \caption{LPCA}
            \end{subfigure}
            \begin{subfigure}[b]{0.19\textwidth}
                    \center
                    \includegraphics[width=\textwidth]{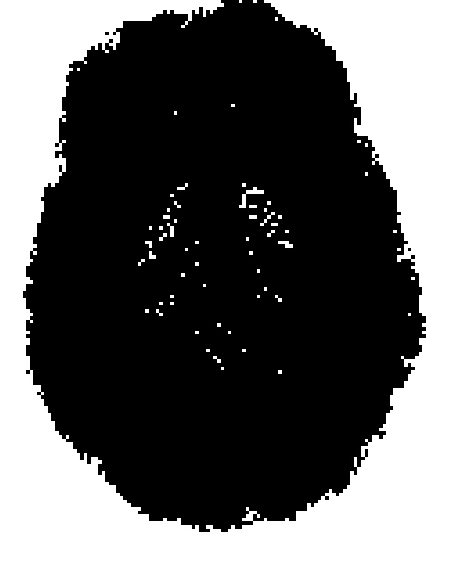}
                    \caption{msPOAS}
            \end{subfigure}

    \caption{Effect of the stabilization algorithm on the compared methods.
    The top row shows an axial slice of the \textit{in-vivo} FA map computed
    on the stabilized dataset, where some voxels are degenerated.
    The bottom row shows their location on a binary brain mask.
    As shown in Table~\ref{tbl:degen_fa}, all methods produce degenerated FA
    voxels on both the regular and stabilized data, with the sole exception of NLSAM.}

    \label{fig:effect_stab}

\end{figure}

Fig.~\ref{fig:effect_stab} shows the FA map when the
compared denoising algorithm are applied on the stabilized data with the algorithm
of~\citep{Koay2009a}. For this experiment, we consider a voxel as being degenerated
if its FA is exactly 0.
The first thing to note is that the algorithm only reprojects
the noisy data on plausible Gaussian distributed values and does not do any
denoising. %
\review{While we used here the algorithm of~\citep{Koay2009a} to correct the noise bias,
another interesting approach consist of producing real-valued datasets as shown in~\citep{Eichner2015}.
This approach does not require estimation of $\sigma^2$ or $N$, but instead use
information contained in the phase of complex-valued acquisitions.}
Secondly, all of the other compared denoising algorithms produce some invalid voxels
on the raw dataset, while having less degenerated voxels on the stabilized dataset
as shown in Table~\ref{tbl:degen_fa}. Nevertheless, only our NLSAM algorithm does
not produce any degenerated FA voxel on the \textit{in-vivo} dataset. As tractography
might rely on a thresholded FA mask~\citep{Chamberland2014b}, any missing white
matter voxel will end the tractography early and produce anatomically invalid
tractography. In the same way, computing FA based statistics in search of group differences
inside a white matter mask might lead to erroneous conclusions when degenerated voxels
are present. This undesirable side effect should be avoided when possible by
choosing a method producing a low number of invalid voxels, such as NLSAM.

\begin{table}[htb]
\centering
\resizebox{\textwidth}{!}{
\begin{tabular}{@{}llrrrrrr@{}}

\toprule
 &  & AONLM & LPCA & msPOAS & NLSAM & Noisy & Mask \\ \midrule
\multirow{2}{*}{Built-in}
& Brain mask & 83 314 (10.1\%) & 10 526 (1.3\%) & 84 319 (10.2\%) & $\oslash$ & 5 994 (0.7\%) & 823 068 (100\%)\\
& WM mask & 29 664 (5.1\%) & 1 298 (0.2\%) & 16 665 (2.9\%) & $\oslash$ & 1 769 (0.3\%) & 578 418 (100\%)\\
\addlinespace
\multirow{2}{*}{Stabilization}
& Brain mask & 10 052 (1.2\%) & 15 750 (1.9\%) & 29 377 (3.6\%) & 0 (0\%) & 9 395 (3.6\%) & 823 068 (100\%)\\
& WM mask & 404 (0.1\%) & 1 468 (0.3\%) & 4 267 (0.7\%) & 0 (0\%) & 2 850 (0.5\%) & 578 418 (100\%)\\
\bottomrule
\end{tabular}}

\caption{Number of degenerated FA voxels inside a brain mask and a white matter
mask for the \textit{in-vivo} dataset.
All methods were compared with their built-in noise estimation on the stabilized
version, but without any additional noise correction factor.
The percentage of degenerated voxels is indicated in parenthesis for
each mask, where a voxel is considered degenerated if its FA value is exactly 0.}
\label{tbl:degen_fa}
\end{table}

\subsection{Reducing the diffusion metrics bias}
\label{sec:discuss_bias}

Fig.~\ref{fig:graph_fa} shows that knowing where errors are committed
gives a better view of how denoising improves upon the noisy data.
We see that our NLSAM algorithm actually has a smaller maximum error
in underestimating the FA most of the time while other methods both over and underestimate the
real FA value and make larger errors near CSF or at borders with the
background. This could indicate that they are subject to problems with partial volume
effect, which seems less important
for NLSAM.

While stabilizing the data alleviates the FA underestimation problem in most cases, it also helps
to reduce the number of degenerated voxels in the \textit{in-vivo} data as shown in
Fig.~\ref{fig:effect_stab}. Both AONLM and msPOAS produce less degenerated FA voxels
on the stabilized dataset as shown in Table~\ref{tbl:degen_fa}, while NLSAM does not produce any
degenerated voxel. In contrast, the noisy data and LPCA have an increased
number of degenerated FA voxels, which might be caused by the diffusion signal
being near the noise floor, thus producing a flat profile which is not
properly recovered in this case.
\review{Reducing the FA bias and avoiding degenerated voxels
by including denoising in the processing pipeline
could improve the statistical analysis in along-tract metrics~\citep{Colby2012}
when looking for group differences.}

\sout{If one is interested by FA difference for group studies
in a particular brain region for example, reducing the FA bias with denoising should be
\review{considered} in the processing pipeline to produce reliable estimates and lower
the number of degenerated FA voxels.}

\subsection{Restoring the coherence of local models}

The CSD algorithm relies on the estimation of the fiber response function
(frf), which in turn relies on the diffusion tensor. To estimate the frf,
one must select voxels containing only a single fiber population. One way to do
this is to estimate it from voxels with a high
FA, usually with FA > 0.7~\citep{Tournier2007,Descoteaux2009e}. We
observed that for the SNR 10 dataset with nc-$\chi$ noise, the noisy dataset, AONLM
and LPCA could not find as much single fiber voxels based on the FA threshold
method as msPOAS or NLSAM since their reconstructed tensors have an inherently lower FA.
This in turns impacts deconvolution since the estimates used for the deconvolution
kernel are less stable,
and lowering the FA threshold too much might violates the single fiber assumption,
which is crucial for the CSD algorithm.
One way to circumvent this could be by using the method
of~\citep{Tax2014}, which is based on a peak amplitude
criterion instead of an FA threshold to identify single fiber voxels.

Fig.~\ref{fig:graph_dnc} shows that msPOAS and NLSAM have larger angular error than
AONLM or LPCA, but this does not seem to impact much the number of valid bundles found by
deterministic tractography. Indeed, the noisy data has the largest angular
error in all cases, but still has a high number of valid bundles in most cases.
This also suggests that a large overestimation or underestimation
of fiber crossings (as reflected by the DNC error) has a higher impact
on tractography. Both LPCA and msPOAS have a lower number of valid bundles than AONLM or NLSAM,
which both have a rather symmetric under and over estimation of the number of peaks.
This means that an overall estimation
or underestimation of the number of crossings bias tractography, as it tends to follow
false crossings or stops prematurely due to a lack of crossings, while a distributed error
is not skewed toward these effects.

In Fig.~\ref{fig:1_2mm}, we see that denoising restores coherence in regions
of crossing fibers that were lost on the noisy dataset or not even present in the lower
spatial resolution 1.8 mm dataset due to a smaller voxel size.
We also see that NLSAM
restores more coherent crossings than the other denoising methods in the junction
of the SLF and the CST, whereas the noisy dataset only shows incoherent crossings.
This actually
enables tractography algorithms to reconstructs tracts which are in agreement
with the expected anatomy.
In the same amount of acquisition time, one can thus acquire higher spatial resolution
DWIs and get better angular information
by post-processing the acquired data with denoising.

\subsection{Limiting spurious fibers from tractography}

We studied the impact of denoising techniques on deterministic
tractography on a synthetic dataset in Section~\ref{sec:tracto}.
One often has to choose between having a high
number of valid bundles and invalid bundles, or having less valid bundles and
at the same time reducing the amount of invalid bundles. The noisy dataset always reaches
a high number of valid bundles,
but also at the price of having the highest number of invalid bundles most of the time.
Our NLSAM algorithm shows a good balance between the number of valid and invalid bundles
at low SNR, especially for the
spatially varying noise case. This is always a tradeoff that one has to choose as seen in the ISMRM 2015
tractography challenge\footnote{\url{http://www.tractometer.org/ismrm_2015_challenge/}}.

For example, the LPCA algorithm
has always a low number of invalid bundles, but also the lowest number of valid bundles for
the spatially varying noise case. In opposition, NLSAM has a high number of valid bundles,
but also a high number of invalid bundles most of the time.

Regarding the deterministic tractography,
changing the stepsize or the maximum curving angle would give different
results in terms of connectivity metrics, indicating that the tractography algorithm
and chosen tractography parameters have a non negligible influence on the end
results~\citep{Chamberland2014b,Girard2014}.
We also used a seeding strategy of 100 seeds per voxel from the ROIs
at each bundles endpoints to ensure a maximal number of valid bundles\review{,}
which promotes a high number of valid bundles
for each dataset.
This shows that the missed bundles are hard to recover or not supported by the data itself,
as opposed to being missed because of inadequate seeding~\citep{Cote2013a}.
On the other hand, this can artificially
increase the number of invalid bundles, which could be reduced by reducing
the number of seeds per voxel.
Since automatic tractography pruning techniques such as~\citep{Cote2015}
might help reduce the number of spurious tracks,
this would indicate that having a higher number of valid bundles would be preferable
since invalid bundles could be potentially removed afterward. In contrast,
a low number of valid bundles cannot be circumvented with further
postprocessing. %
Nevertheless, denoising increases the valid connection to connection ratio
and reduces the number of invalid bundles,
thus bringing confidence in the validity of the tractography results when compared to
the noisy datasets.

For the \textit{in-vivo} dataset tracking shown in Fig.~\ref{fig:bundles}, we see that
tractography benefits from higher spatial resolution acquisitions, but that the produced
tracts are slightly more noisy. Combining the high spatial resolution, highly noisy dataset with a
denoising algorithm at the beginning of the processing pipeline gives more
anatomically plausible tracts in the end. The AF and CST produced by the NLSAM
denoised dataset are both more anatomically plausible than their noisy
or lower spatial resolution counterpart, which have less fanning fibers in
the case of the CST. This shows that high resolution DWIs exhibits more anatomical
information thanks to the smaller voxel size which might not be easily discernible
at a lower spatial resolution~\citep{Sotiropoulos}.
Acquiring at higher spatial resolution could also help resolve
complicated fiber configurations such as crossings fibers from fanning fibers
\review{or disentangle small structures like the optic chiasm~\citep{Roebroeck2008}},
which is not possible at lower spatial resolution~\citep{Jones2012,Calabrese2014}.

\subsection{\review{Other methods for high spatial resolution acquisitions}}

\review{
We have shown in Fig.~\ref{fig:1_2mm} that high spatial resolution acquisitions
which are noisy at first can reveal improved anatomical details when they are
subsequently denoised.
This indeed suggests that high resolution acquisition can now be acquired on clinical
scanners. Recently, other algorithms enabling a high spatial resolution
at the acquisition level have been suggested~\citep{Scherrer2015,Ning2015c}.
These methods both rely on smartly fusing the (complementary) data of
multiple acquisitions acquired at a lower spatial resolution to obtain a single
high resolution volume. While the approach we suggest here is to acquire a
single volume using a standard sequence, both techniques are fundamentally exploiting
different aspects to increase the available spatial resolution.
As such, it would be possible to combine our denoising algorithm with the
reconstruction algorithms presented in~\citep{Scherrer2015,Ning2015c}.
}

\subsection{Current limitations and possible improvements}

Although most models assume a Rician or  nc-$\chi$ noise distribution,
this does not take into account the noise correlation between each
coils \review{or the effect of acceleration techniques that subsample the k-space}~\citep{Aja-Fernandez2014}.
The development of
correction factors for existing algorithms relies on computing the
\textit{effective} values for the noise standard deviation $\sigma$ and the
number of \review{degrees of freedom of the nc-$\chi$ distribution,
which is expected to be smaller than 2N}.
These values can be used to take into account the
correlation introduced between the coils in parallel imaging
acquisitions~\citep{Brion2013b}.
To consider the fact that
the noise distribution nature might vary spatially in addition to the noise
variance, one can use
\textit{a priori} information obtained from the scanner through
the SENSE sensitivity maps or the GRAPPA weights and need to estimate the
correlation between each of the receiver coils.
We could explicitly add such a correction
to our algorithm since we work locally with the
stabilization algorithm, Eq.~(\ref{eq:dl}) and Eq.~(\ref{eq:dl_constrained}).
Using multiband acceleration also modifies the noise properties due to the introduced aliasing,
which further strengthen the idea that spatially adaptive denoising algorithms should be used
on modern scanners and sequences~\citep{Ugurbil2013}.
Nevertheless, obtaining the needed map for a SENSE reconstruction
or the required GRAPPA weights might not be easily feasible in a clinical setting.
We also intend to revisit the order in which preprocessing algorithms (motion correction,
eddy currents correction, distortions correction) should be applied since these steps
require interpolation, which could also introduce spatial correlation in the noise profile.
\review{This also makes the noise distribution deviate slightly from its theoretical distribution,
where parameters vary spatially instead of being fixed constants
for the whole volume~\citep{Aja-Fernandez2014}.}
\review{Nevertheless, we have observed experimentally that our NLSAM algorithm is robust to
small subject motion thanks to the local neighborhood processing.
In case where artifacts (such as epi distortions) might undermine the denoising process,
one can first correct for these artifacts using a nearest neighbor interpolation, which should not
modify the noise distribution. Subsequent corrections can then be performed after denoising using other
kinds of interpolation as needed.}

While developing the NLSAM algorithm, we found that using a bigger 3D
patchsize did not significantly improve the denoising quality, while
augmenting both computing time and memory requirements.
Our implementation also allows one create the smallest subset of angular
neighbors covering all DWIs through a greedy set cover algorithm.
This option (named "NLSAM fast" in Table~\ref{tbl:runtime}) leads to a speedup of 3 to 4 times,
but at the cost of slightly
reducing the denoising performance since some DWIs might be denoised only
once instead of multiple times. We used the fully covered version for our experiments,
which were run on a machine running Ubuntu Linux 12.04 with a quad core
Intel i7 930 at 2.8 GHz and 18 \review{GB} of RAM. Table~\ref{tbl:runtime} reports the runtime of the various
algorithms in minutes and their RAM usage. While the computing time required by NLSAM
is larger than the other methods, our Python implementation is fairly unoptimized and could
be sped up to competitive runtimes by various code optimizations or lowering
the maximum number of iterations in Eq.~(\ref{eq:dl_constrained}).

\begin{table}[bht]
\centering
\begin{tabular}{@{}lccccc@{}}
\toprule
                       & AONLM & LPCA & msPOAS & NLSAM & NLSAM fast\\
    \midrule
    Time (mins)        & 22.2     & 3.7    & 4.0     & 37.1  & 9.8 \\
    RAM usage  \review{(MB)}     & 552      & 640    & 1543    & 606   & 412 \\
\bottomrule
\end{tabular}
\caption{Required time and RAM usage for the compared denoising algorithms on the b1000
SNR 10 dataset with stationary Rician noise.}
\label{tbl:runtime}
\end{table}

\FloatBarrier

\section{Conclusion}
\label{sec:conclusion}

In this paper, we introduced a new denoising method, the Non Local
Spatial and Angular Matching (\mbox{NLSAM}), which is specifically designed to take
advantage of diffusion MRI data.
Our method is based on
1) Correcting the spatially varying Rician and nc-$\chi$ noise bias
2) Finding similar DWIs through angular neighbors to promote sparsity
3) Iteratively denoise similar patches and their neighbors locally with dictionary learning,
where the local variance is used as an upper bound on the $\ell_2$ reconstruction error.
We extensively compared quantitatively our new method against three other state-of-the-art
denoising methods on a synthetic
phantom and qualitatively on an \textit{in-vivo} high resolution dataset.
We also showed that taking into account both the effect of spatially varying noise and
\review{non-Gaussian} distributed noise is crucial in order to denoise effectively the DWIs.
Our NLSAM algorithm is freely available\footnote{\url{https://github.com/samuelstjean/nlsam}},
restores perceptual information, removes the noise bias in
common diffusion metrics
and produces more
anatomically plausible tractography on a high spatial resolution
\textit{in-vivo} dataset when compared to a lower spatial resolution acquisition
of the same subject.

Since our NLSAM algorithm can be used on any already acquired dataset and does not add
any acquisition time, this shows that denoising the data should be a pre-processing
part of every pipeline, just like any other correction method that is commonly
applied for artifacts removal.
With that in mind, the diffusion MRI community could aim for higher spatial resolution DWIs,
without requiring the use of
costly new hardware or complicated acquisition schemes.
This could in turn reveal new anatomical details,
which are not achievable at the spatial resolution currently used in diffusion MRI.

\section*{Acknowledgments}

We would like to thank Emmanuel Caruyer for adding our requested features and
general help with the phantomas library. We also thank the Tractometer team
(Jean-Christophe Houde and Marc-Alexandre C\^{o}t\'{e}, \url{www.tractometer.org}) for
the help in using the tractography evaluation system. We also extend our
thanks to José V. Manjón for making the LPCA publicly available and Karsten Tabelow
for the publicly available msPOAS code, helping us in the usage and in the choosing
of good parameters for the
\textit{in-vivo} dataset denoising.
Another special thanks to Guillaume Gilbert from Philips Healthcare
for the help with the \textit{in-vivo} datasets acquisition.
Samuel St-Jean was funded by the
NSERC CREATE Program in Medical Image Analysis.
\review{Pierrick Coup\'{e} was supported by the French State,
managed by the French National Research Agency (ANR) in the frame
of the Investments for the future Programme IdEx Bordeaux (HL-MRI ANR-10-IDEX-03-02),
Cluster of excellence CPU, TRAIL (HR-DTI ANR-10-LABX-57) and the CNRS multidisciplinary project \textit{D\'{e}fi ImagIn}.}

\FloatBarrier

\appendix
\section{The NLSAM Algorithm}
\label{sec:appendix}

This appendix outlines the NLSAM algorithm as pseudo code.
The original implementation in Python is freely
available at \url{https://github.com/samuelstjean/nlsam}.

\begin{algorithm}[htb]

 \SetAlgoLined
 \KwData{4D dMRI data, $an=$ number of angular neighbors, $ps=$ spatial patch size, $N=$ Number of coils, $max\_iter = 40$}
 \KwResult{Denoised data with NLSAM}

    \textbf{Step 1}\;
        \hspace*{30pt} Find $\sigma_G$ with either PIESNO or Eq.~(\ref{eq:noise_field})\;
        \hspace*{30pt} Apply noise stabilization with $\sigma_G$ and $N$ coils\;

    \vspace{10pt}

   \ForEach{DWI in dMRI data}
    {
        \textbf{Step 2}\;
        \hspace*{30pt} Find the closest $an$ angular neighbors\;
        \hspace*{30pt} Create 4D block with b0, DWI and its $an$ neighbors\;
        \hspace*{30pt} Extract all overlapping patches of size $(ps, ps, ps)$\;

        \textbf{Step 3}\;
        \hspace*{30pt} Apply Eq.~(\ref{eq:dl}) to find \textbf{D}\;
        \hspace*{30pt} Iterate Eq.~(\ref{eq:dl_constrained}) to find $\boldsymbol{\alpha}$ until convergence or $max\_iter$ is reached\;
        \hspace*{30pt} Average overlapping patches based on sparsity with Eq.~(\ref{eq:w_avg})\;
        \hspace*{30pt} Return to original shape\;
    }

    \vspace{10pt}

   \ForEach{Denoised DWI in dMRI data}
    {
        Average all Denoised DWI representations\;
    }

\caption{The proposed NLSAM denoising algorithm.}
\label{alg:main}
\end{algorithm}

\FloatBarrier

\bibliographystyle{elsarticle-harv}
\bibliography{nlsam_denoising.bbl}

\end{document}